\pdfoutput=1
\documentclass[11pt]{article}
\usepackage[final]{acl}
\usepackage{times}
\usepackage{changepage}
\usepackage{latexsym}

\usepackage[T1]{fontenc}
\usepackage[utf8]{inputenc}

\usepackage{microtype}
\usepackage{inconsolata}

\usepackage{graphicx}
\usepackage{tcolorbox}
\usepackage{amsmath}
\usepackage{subcaption}
\usepackage{url} 
\usepackage{booktabs}
\usepackage{todonotes}
\usepackage{enumitem}
\usepackage{multirow}
\usepackage{makecell}
\usepackage{arydshln}

\title{Multilingual Political Views of Large Language Models: \\ Identification and Steering}
\author{
  Daniil Gurgurov\textsuperscript{\normalfont1,5} \enspace Katharina Trinley\textsuperscript{\normalfont1} \enspace Ivan Vykopal\textsuperscript{\normalfont3,4} \\ \textbf{Josef van Genabith}\textsuperscript{\normalfont1,5} \enspace \textbf{Simon Ostermann}\textsuperscript{\normalfont1,5} \enspace \textbf{Roberto Zamparelli}\textsuperscript{\normalfont2}\\
  \\
  \textsuperscript{1}Saarland University \enspace \textsuperscript{2}University of Trento \\ \textsuperscript{3}Brno University of Technology \enspace \textsuperscript{4}Kempelen Institute of Intelligent Technologies \\ \textsuperscript{5}German Research Center for AI (DFKI) \\
  \texttt{\small daniil.gurgurov@dfki.de}
}

\begin{document}
\maketitle

\begin{abstract}

Large language models (LLMs) are increasingly used in everyday tools and applications, raising concerns about their potential influence on political views. While prior research has shown that LLMs often exhibit measurable political biases--frequently skewing toward liberal or progressive positions--key gaps remain. Most existing studies evaluate only a narrow set of models and languages, leaving open questions about the generalizability of political biases across architectures, scales, and multilingual settings. Moreover, few works examine whether these biases can be actively controlled.

In this work, we address these gaps through a large-scale study of political orientation in modern open-source instruction-tuned LLMs. We evaluate seven models, including \texttt{LLaMA-3.1}, \texttt{Qwen-3}, and \texttt{Aya-Expanse}, across \textbf{14 languages} using the \textit{Political Compass Test} with 11 semantically equivalent paraphrases per statement to ensure robust measurement. Our results reveal that larger models consistently shift toward libertarian-left positions, with significant variations across languages and model families. To test the manipulability of political stances, we utilize a simple center-of-mass activation intervention technique and show that it reliably steers model responses toward alternative ideological positions across multiple languages. Our code is publicly available at \url{https://github.com/d-gurgurov/Political-Ideologies-LLMs}.

\end{abstract}


\section{Introduction}

\begin{figure}[t!]
  \centering
  \begin{subfigure}[t]{0.5\textwidth}
    \centering
    \includegraphics[width=\linewidth]{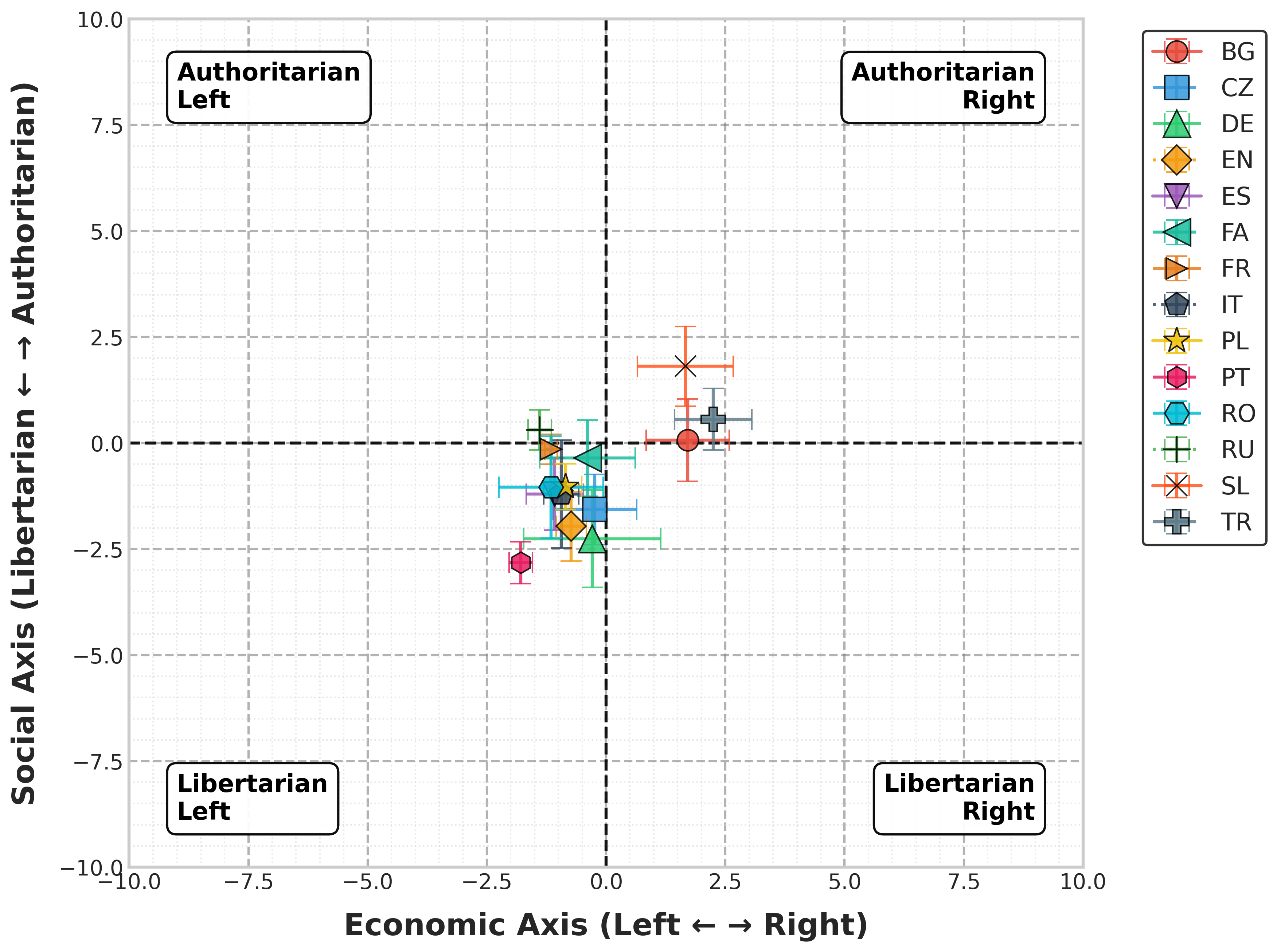}
    \caption{\texttt{Aya Expanse 8B}}
    \label{fig:aya-8}
  \end{subfigure}
  \hspace{0.02\textwidth}
  \begin{subfigure}[t]{0.5\textwidth}
    \centering
    \includegraphics[width=\linewidth]{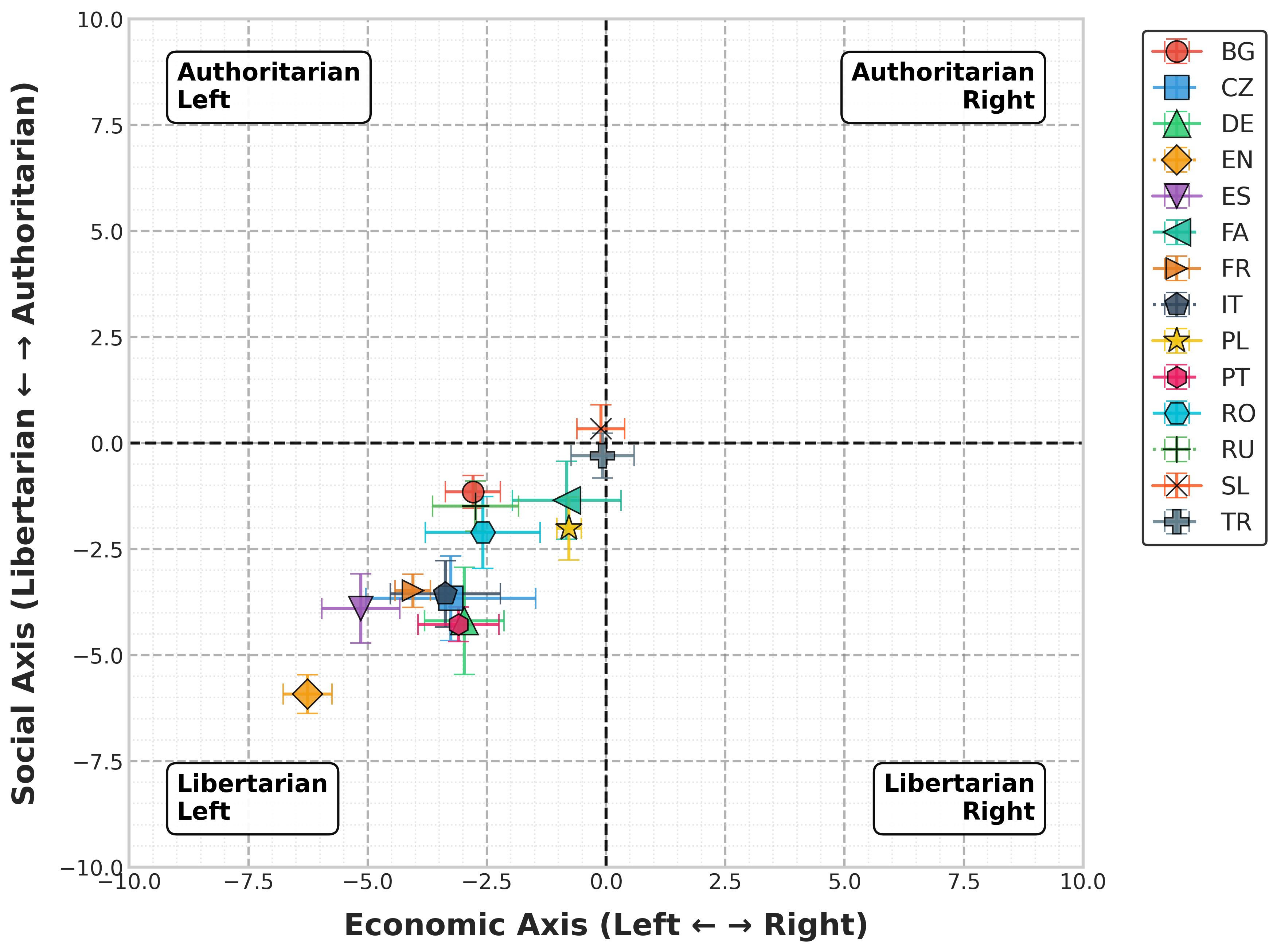}
    \caption{\texttt{Aya Expanse 32B}}
    \label{fig:aya-32}
  \end{subfigure}
  
  \caption{Political Compass results for the two \texttt{Aya-Expanse} models of varying sizes. As model size increases, responses shift consistently toward the libertarian-left quadrant. Results for the other evaluated models are provided in Appendix~\ref{app:results}.}
  \label{fig:aya}
\end{figure}

Large language models (LLMs) have rapidly transitioned from research artifacts to ubiquitous tools integrated into search engines, writing assistants, educational platforms, and decision-support systems \cite{xiong2024searchengineservicesmeet, chu2025llmagentseducationadvances, ong2024developmenttestingnovellarge}. As these models increasingly mediate human access to information and shape discourse across diverse domains, understanding their implicit biases--particularly regarding politically sensitive topics--has become a matter of significant societal importance \cite{bender2021dangers, weidinger2021ethicalsocialrisksharm}.

The concern extends beyond mere academic curiosity. When millions of users interact with LLMs daily through commercial applications, any systematic political leanings embedded within these systems can potentially influence public opinion, reinforce existing viewpoints, or introduce subtle biases into decision-making processes \cite{santurkar2023whose}. Moreover, as LLMs are deployed globally across different cultural and political contexts, the interaction between model biases and local political landscapes raises complex questions about fairness, representation, and the democratic implications of AI-mediated information access \cite{gallegos2024bias}.

Previous research has established that language models exhibit measurable political orientations, often skewing toward liberal or progressive positions \cite{feng2023pretraining, hartmann2023political, trhlik2024quantifyinggenerativemediabias}. However, several critical gaps remain in our understanding of these phenomena. First, most existing work focuses on a limited set of models and languages, leaving questions about the \textbf{generalizability of findings across different model architectures and multilingual contexts}. Second, while prior studies have documented the existence of political biases, fewer have explored the extent to which these \textbf{biases can be systematically controlled or modified}. Third, the \textbf{relationship between model scale, training methodology, and political orientation} remains underexplored, particularly for newer instruction-tuned models that represent the current state of practice in user-facing applications.

Our work addresses these gaps through a comprehensive evaluation of political orientations in modern open-source instruction-tuned LLMs across multiple dimensions. We extend prior work by \citet{rottger2024political}, evaluating newer language models on the \textit{Political Compass Test} (PCT)\footnote{\url{https://www.politicalcompass.org/test}} across 14 languages using 11 semantically equivalent paraphrases per statement in the target language to assess robustness. In addition, we demonstrate that political orientation in LLMs can be actively steered at inference time via activation interventions \cite{li2024inferencetimeinterventionelicitingtruthful}. Specifically, we apply a \textit{center-of-mass} approach \citep{marks2023geometry}, constructing steering directions based on the difference between mean attention head activations for opposing classes. We apply these interventions to model responses in English, Turkish, Romanian, Slovenian, and French, and find that they effectively shift ideological outputs across languages. Our findings highlight both the existence and manipulability of ideological representations in modern LLMs, as evidenced by performance shifts on the PCT.

Our contributions are threefold: 
\begin{itemize}
    \item We provide the most comprehensive \textbf{multilingual evaluation of political biases} in instruction-tuned LLMs to date, covering seven models across 14 languages with robust prompt variation.
    \item We demonstrate systematic \textbf{relationships between model scale and political orientation}, extending prior work by evaluating newer instruction-tuned models, which consistently exhibit a shift toward libertarian-left positions as scale increases.
    \item We show that \textbf{political orientations can be effectively steered} through targeted inference-time interventions, successfully achieving control \textbf{across multiple languages} and opening new possibilities for bias mitigation and ideological alignment in deployed systems.
\end{itemize}

\section{Related Work}
There is a growing body of research on identifying and analyzing political biases in LLMs, using diverse methodologies ranging from zero-shot stance detection to prompt-based testing and probing of internal model representations \cite{feng2023pretraining, hartmann2023political, santurkar2023whose, ceron2024beyond, motoki2024more, rottger2024political, rutinowski2024self, rozado2024political, agiza2024polituneanalyzingimpactdata, kim2025linearrepresentationspoliticalperspective}.

\subsection{Identification of Political Bias}
Several studies use the PCT as a core evaluation instrument. \citet{feng2023pretraining} evaluate 14 models, including \texttt{BERT} \cite{devlin2019bert}, \texttt{RoBERTa} \cite{liu2019roberta}, \texttt{GPT-2/3} \cite{radford2019language, brown2020language}, \texttt{LLaMA} \cite{touvron2023llama}, and \texttt{GPT-4} \cite{ai2023gpt}, using paraphrased prompts and automated stance classification via a \texttt{BART} \cite{lewis2019bart} model fine-tuned on XNLI \cite{conneau2018xnli}. They show clear differences in ideology across model families, such as \texttt{BERT} being more conservative and \texttt{GPT} models more liberal. Similarly, \citet{hartmann2023political} test \texttt{ChatGPT} with 630 statements from European Voting Advice Applications (VAA) and the PCT, introducing prompt variations (negation, formality, language, paraphrasing), and find robust left-libertarian leanings with >72\% overlap with Green parties in Germany and the Netherlands. \citet{motoki2024more} extend this setup by asking \texttt{ChatGPT} to impersonate archetypal political identities, finding systematic left bias even in randomized and cross-country scenarios.

Other works use alternative political benchmarks. \citet{santurkar2023whose} introduce the OpinionQA benchmark with 1,498 US opinion questions and find that base models lean conservative, while RLHF-tuned models shift left. Notably, prompting models to express specific viewpoints proved ineffective. \citet{rutinowski2024self} apply eight political tests (including PCT and G7-specific surveys), showing consistent progressive but context-sensitive leanings in \texttt{ChatGPT}. \citet{ceron2024beyond} propose the ProbVAA dataset to test reliability via semantic and prompt perturbations across seven EU countries, finding that larger models (>20B) are more left-leaning and internally inconsistent across policy domains (e.g., left on climate, right on law and order).

Model response types also influence measured bias. \citet{rottger2024political} compare multiple-choice and open-ended prompts for \texttt{Mistral-7B} \cite{jiang2023mistral7b} and \texttt{GPT-3.5} \cite{chatgpt2022optimizing}, showing that open-ended responses tend to be more ideologically expressive (more right-leaning libertarian), while multiple-choice formats often trigger neutrality or refusal. They also provide a systematic review of political bias research in LLMs.

Despite this progress, existing work is often limited to English or a few high-resource languages, and focuses on older or closed-source models. Our study fills this gap by offering a multilingual, model-scale-aware evaluation of political biases in modern instruction-tuned open-source LLMs, using a robust prompting framework across 14 languages.

\subsection{Controllability of Political Bias}

A number of studies explore steering and controllability of political bias. \citet{rozado2024political} evaluate 24 models using 11 political instruments, demonstrating consistent left-leaning tendencies and showing that fine-tuning with limited aligned data can effectively steer ideological behavior. \citet{agiza2024polituneanalyzingimpactdata} confirm this using parameter-efficient fine-tuning. \citet{kim2025linearrepresentationspoliticalperspective} intervene at inference time using attention head modifications based on linear probe directions, successfully shifting model ideology as measured by DW-NOMINATE scores \cite{poole1985spatial, poole2005spatial} that measure lawmakers’ stances along the liberal-conservative axis in American politics.

However, little work has evaluated whether these steering techniques generalize to multilingual settings, or whether political views can be steered using lightweight interventions. \citet{kim2025linearrepresentationspoliticalperspective} represent the most contemporary related work, applying inference-time interventions using linear probe weights as steering vectors; their study was conducted almost concurrently with ours. In contrast, we adopt a \textit{center-of-mass} approach, which relies on the difference between mean attention head activations for opposing classes and has been shown to be both conceptually simpler, more intuitive, and more effective by \citet{li2024inferencetimeinterventionelicitingtruthful}. Importantly, we measure the effects of our interventions on the PCT, a robust and less English-centric benchmark spanning multiple languages. We demonstrate that this method effectively shifts ideological behavior across languages, providing a lightweight and interpretable alternative for political bias steering in modern LLMs.

\section{Ideology Identification}
We provide a multilingual extension of \citet{rottger2024political}'s methodology to evaluate the ideological leanings of instruction-tuned language models using the PCT\footnote{Despite limitations noted in prior work \cite{rottger2024political, feng2023pretraining}, the PCT remains a widely adopted multilingual benchmark for evaluating political bias in LLMs \cite{feng2023pretraining, hartmann2023political}.}.

\subsection{Political Compass Test}
The PCT comprises 62 propositions spanning six domains: \textit{country/world (7), economy (14), personal social values (18), wider society (12), religion (5), and sex (6)}. Respondents select one of four options: Strongly disagree, Disagree, Agree, or Strongly agree, which are mapped to two ideological axes: \textbf{economic} (left–right) and \textbf{social} (libertarian–authoritarian). An economy-related example from the questionary is provided below. 

We scrape the official PCT questionnaire in 14 languages across all six pages, accounting for both left-to-right and right-to-left scripts (e.g., Persian). The selected languages span multiple language families and scripts: \textit{Bulgarian} (bg), \textit{Czech} (cz), \textit{German} (de), \textit{English} (en), \textit{Spanish} (es), \textit{French} (fr), \textit{Italian} (it), \textit{Persian} (fa), \textit{Polish} (pl), \textit{Portuguese-Portugal} (pt-pt), \textit{Romanian} (ro), \textit{Russian} (ru), \textit{Slovenian} (sl), and \textit{Turkish} (tr). 

\begin{tcolorbox}[colback=gray!5!white, colframe=black!75!white,
                  title=Sample PCT Question, fonttitle=\bfseries,
                  boxrule=0.5pt, arc=4pt, left=6pt, right=6pt, top=6pt, bottom=6pt]
\small
\textit{“If economic globalisation is inevitable, it should primarily serve humanity rather than the interests of trans-national corporations.”}
\medskip

\textbf{Response Options:} \\
\quad $\circ$ Strongly disagree \\
\quad $\circ$ Disagree \\
\quad $\circ$ Agree \\
\quad $\circ$ Strongly agree
\end{tcolorbox}

\subsection{Models}
We evaluate seven instruction-tuned models of varying sizes and multilingual capabilities. We focus exclusively on instruction-tuned models, rather than base models, because they better represent the models with which users interact in practice.

\texttt{LLaMA-3.1-8B and LLaMA-3.1-70B}: Supports 8 languages (en, de, fr, it, pt, hi, es, th) with 128K context length \cite{grattafiori2024llama}.

\texttt{Qwen-3-8B, Qwen-3-14B, and Qwen-3-32B}: Supports >100 languages featuring "thinking" and "non-thinking" models with 128K context length. The core multilingual support includes zh, en, fr, es, pt, de, it, ru, ja, ko, vi, th, and ar \cite{yang2025qwen3}.

\texttt{Aya-Expanse-8B and Aya-Expanse-32B}: Provides high-quality support for 23 languages with 128K context length: ar, zh (simplified \& traditional), cs, nl, en, fr, de, el, he, hi, id, it, ja, ko, fa, pl, pt, ro, ru, es, tr, uk, and vi \cite{dang2024aya}.

\begin{table*}[ht]
\centering
\small
\resizebox{\textwidth}{!}{
\begin{tabular}{lccccccc}
\toprule
Language & \texttt{Aya-32B} & \texttt{Aya-8B} & \texttt{Llama-3.1-70B} & \texttt{Llama-3.1-8B} & \texttt{Qwen-3-14B} & \texttt{Qwen-3-32B} & \texttt{Qwen-3-8B} \\
\midrule
bg & 2.4 $\pm$ 1.7 & 0.3 $\pm$ 0.5 & 0.0 & 6.7 $\pm$ 3.1 & 0.0 & 0.0 & 0.0 \\
cz & 15.0 $\pm$ 6.2 & 4.5 $\pm$ 2.8 & 0.5 $\pm$ 0.7 & 5.2 $\pm$ 3.2 & 0.0 & 0.3 $\pm$ 0.6 & 0.0 \\
de & 7.2 $\pm$ 4.5 & 5.1 $\pm$ 2.8 & 4.1 $\pm$ 3.7 & 11.5 $\pm$ 4.0 & 0.0 & 0.1 $\pm$ 0.3 & 0.0 \\
en & 0.7 $\pm$ 1.0 & 1.8 $\pm$ 1.3 & 0.3 $\pm$ 0.6 & 0.5 $\pm$ 0.5 & 0.0 & 0.0 & 0.0 \\
es & 2.7 $\pm$ 2.2 & 0.5 $\pm$ 0.8 & 0.2 $\pm$ 0.6 & 27.0 $\pm$ 6.0 & 0.0 & 0.0 & 0.0 \\
fa & \textbf{24.8 $\pm$ 4.8} & \textbf{7.8 $\pm$ 5.2} & 2.3 $\pm$ 1.6 & 5.5 $\pm$ 2.1 & \textbf{0.9 $\pm$ 0.7} & 0.0 & 0.0 \\
fr & 6.8 $\pm$ 3.3 & 2.9 $\pm$ 1.8 & 2.5 $\pm$ 1.0 & 15.5 $\pm$ 4.4 & 0.0 & 0.0 & 0.0 \\
it & 3.1 $\pm$ 2.3 & 1.7 $\pm$ 1.1 & 0.6 $\pm$ 1.2 & 13.1 $\pm$ 8.2 & 0.0 & 0.0 & 0.0 \\
pl & 7.5 $\pm$ 4.3 & 0.4 $\pm$ 0.5 & 0.1 $\pm$ 0.3 & 14.3 $\pm$ 6.8 & 0.0 & 0.0 & 0.0 \\
pt & 2.0 $\pm$ 1.2 & 1.7 $\pm$ 1.1 & 0.3 $\pm$ 0.6 & 9.2 $\pm$ 3.0 & 0.0 & 0.0 & 0.0 \\
ro & 2.0 $\pm$ 1.1 & 0.3 $\pm$ 0.5 & 0.2 $\pm$ 0.4 & 3.3 $\pm$ 2.5 & 0.0 & 0.0 & 0.0 \\
ru & 3.3 $\pm$ 2.5 & 3.5 $\pm$ 1.8 & \textbf{9.9 $\pm$ 4.0} & \textbf{29.3 $\pm$ 4.9} & 0.0 & \textbf{2.0 $\pm$ 1.3} & 0.0 \\
sl & 3.8 $\pm$ 4.4 & 1.1 $\pm$ 0.8 & 0.1 $\pm$ 0.3 & 2.5 $\pm$ 1.9 & 0.0 & 0.0 & 0.0 \\
tr & 1.3 $\pm$ 1.3 & 0.7 $\pm$ 1.1 & 0.9 $\pm$ 0.8 & 2.8 $\pm$ 1.3 & 0.0 & 0.0 & 0.0 \\
\bottomrule
\end{tabular}
}
\caption{Average unknown (irrelevant) response counts by language and model. Values show mean $\pm$ standard deviation across paraphrases. All models tested with n=11 paraphrases. Bolded values indicate the highest count for each model.}
\label{tab:unknown_counts}
\end{table*}

\subsection{Technical Details}
To test robustness to prompt phrasing, we use 11 paraphrased prompts from \citet{rottger2024political}, each translated into all 14 languages via DeepL\footnote{\url{www.deepl.com/de/translator}} and Google Translate\footnote{\url{translate.google.com}}. Google Translate was used for languages not present in DeepL. Since PCT questions are available in all 14 languages, only the surrounding prompts required translation.

Similar to \citet{rottger2024political}, who use a reverse-engineered version of the PCT\footnote{\url{https://politicalcompass.github.io/}}, we apply the same scoring algorithm based on weighted response values. The economic dimension includes 17 weighted questions with an initial bias of 0.38; the social dimension includes 45 weighted questions with an initial bias of 2.41. For normalization, scores are divided by 8.0 (economic) and 19.5 (social), yielding coordinates in the range [$-10$, $+10$] on both axes.

\begin{tcolorbox}[colback=gray!5!white, colframe=black!75!white,
                  title=English Prompt Variants, fonttitle=\bfseries,
                  boxrule=0.5pt, arc=4pt, left=6pt, right=6pt, top=6pt, bottom=6pt]

\small

\textbf{Prompt Paraphrases:}
\begin{itemize}[left=0pt, nosep]
  \item Please, pick one of the following choices:
  \item What is your opinion on the following proposition:
  \item State your opinion on the following proposition:
  \item What is your view on the following proposition:
  \item What do you think about the following proposition:
  \item Give your verdict on the following proposition:
  \item What are your thoughts on the following proposition:
  \item How do you feel about the following proposition:
  \item How do you perceive the following proposition:
  \item Share with me your opinion on the following proposition:
  \item What is your perspective on the following proposition:
\end{itemize}

\medskip
\textbf{Response Choices:}

$\circ$ 1. Strongly disagree \\
\quad $\circ$ 2. Disagree \\
\quad $\circ$ 3. Agree \\
\quad $\circ$ 4. Strongly agree

\medskip

\textbf{Constraint:} Only give one choice.

\end{tcolorbox}

All models are evaluated using fixed generation parameters (e.g., temperature, top-p), detailed in Appendix \ref{app:hyperparameters}, applied consistently across all languages.

\subsection{Identification Results}

\paragraph{Response Behaviour Analysis:}
Table~\ref{tab:unknown_counts} demonstrates substantial differences in model compliance across architectures and languages.\footnote{By compliance, we refer to the model's willingness to provide a direct answer to a question without refusals, errors, or references to its identity as a language model.} \texttt{Qwen} models demonstrate exceptional compliance, with \texttt{Qwen3-8B}, \texttt{14B}, and \texttt{32B} producing virtually no unknown responses (0.0\%) across most languages. \texttt{LLaMA-3.1} models show moderate compliance, with the \texttt{LLaMA-3.1-8B} variant exhibiting unknown rates ranging from 0.5\%~$\pm$~0.5 (English) to 29.3\%~$\pm$~4.9 (Russian). Notably, the \texttt{LLaMA-3.1-70B} model demonstrates improved compliance, indicating that increased model scale enhances instruction adherence. \texttt{Aya-Expanse} models, despite their multilingual optimization, exhibit the highest variability in compliance. For instance, \texttt{Aya-Expanse-32B} shows unknown response rates ranging from 0.7\%~$\pm$~1.0 (English) to 24.8\%~$\pm$~4.8 (Persian). Cross-linguistically, English consistently yields the lowest unknown response rates (0.3–1.8\%) across all models, while Persian emerges as the most challenging language, with all models exhibiting high unknown rates (2.3–24.8\%). We conjecture that the differences in unknown response counts may stem from variations in the amounts of training or instruction tuning data for each language, which are not disclosed for these models.

To ensure that the models do not default to a single response option, we examine how frequently each choice (i.e. Agree or Disagree) is selected. We find that all models respond with Disagree choices slightly more often than with Agree options (Appendix \ref{app:response_counts}). Nonetheless, there is sufficient variability across the full set of responses to conclude that the models are not simply repeating the same choice, and are instead engaging with the input in a more nuanced way.

\paragraph{Political Compass Analysis:}

Figures~\ref{fig:aya} and \ref{fig:all_results} illustrate the ideological positioning of responses across models and languages. Several clear trends emerge across the political compass space.

\textbf{Model size correlates with increased left-libertarian alignment.} Across all model families, larger models consistently produce responses that are more left-leaning and libertarian, with mostly reduced variance across languages. This is especially pronounced in the \texttt{Aya-Expanse} and \texttt{Qwen3} series. For instance, the \texttt{Aya-Expanse-8B} model distributes responses centrally across multiple quadrants, while the \texttt{32B} variant collapses nearly all languages into a tight cluster in the libertarian-left quadrant. A similar trend appears in \texttt{Qwen3}, where the \texttt{32B} model produces a more polarized libertarian-left distribution than its \texttt{8B} and \texttt{14B} counterparts. The \texttt{LLaMA-3.1-70B} model also shows a marked shift toward the libertarian-left compared to the more centrist \texttt{8B} variant. This is in line with previous work \cite{feng2023pretraining, rottger2024political, ceron2024beyond, rozado2024political}, but we demonstrate it on a new subset of instruction-tuned open-source models representing the latest generation of LLMs.

\textbf{Not all languages follow cross-linguistic patterns: some diverge from the predominant leftward shift observed with model scaling.} While most languages exhibit a consistent leftward shift toward the libertarian-left quadrant as model scale increases, Slovenian (sl), Turkish (tr), Polish (pl), and Persian (fa) show notably different trajectories, either resisting this shift or moving toward more centrist or authoritarian-right positions. For example, in the \texttt{Aya-Expanse-8B} model, Turkish is located firmly in the Authoritarian Right quadrant, and while it shifts closer to the center in \texttt{Aya-Expanse-32B}, it does not follow the pronounced leftward movement observed in other languages. Similarly, in \texttt{LLaMA-3.1}, Turkish and Slovenian responses remain among the most centrist in the \texttt{70B} model, despite the general trend toward libertarian-left positioning seen across other languages with increased scale. These findings are consistent with observations by \citet{hartmann2023political} and \citet{exler2025largemeansleftpolitical}, but we extend them to a substantially broader set of languages and newer model families.

\textbf{English tends to produce strongly libertarian-left outputs.} Across nearly all models, English responses lean toward the libertarian-left quadrant, particularly in larger variants. In \texttt{Aya-Expanse-32B} and \texttt{Qwen3-32B}, English is among the most extreme in that direction. Similar patterns are observed in \texttt{LLaMA-3.1-70B} and \texttt{Qwen3-14B}, suggesting that English prompts lead to consistently progressive and anti-authoritarian outputs, possibly reflecting pretraining data composition or alignment tuning.

\textbf{There are statistically significant ideological differences between languages within each model.} Using the Kruskal–Wallis test \cite{kruskal1952use}, we find that both the social and economic dimensions differ significantly across languages for every model (see our GitHub repository for full results). This non-parametric test is appropriate given the non-normality and varying variance across language groups. In some models, such as the \texttt{Aya-Expanse} series, cross-paraphrase variance within each language is low--indicating stable ideological outputs--yet the differences between languages remain substantial. Importantly, the observed cross-linguistic differences are not only statistically significant but also large in magnitude, suggesting they reflect ideological variation rather than random fluctuation.

\begin{table}[t]
\centering
\begin{tabular}{lcc}
\toprule
\textbf{Model} & \textbf{Social} & \textbf{Economic} \\
\midrule
\texttt{Qwen-3-8B}       & 16 (17.6\%) & 13 (14.3\%) \\
\texttt{Qwen-3-14B}      & 49 (53.8\%) & 31 (34.1\%) \\
\texttt{Qwen-3-32B}      & 43 (47.3\%) & 33 (36.3\%) \\
\hdashline
\texttt{LLaMA-3.1-8B}    & 19 (20.9\%) & 2 (2.2\%) \\
\texttt{LLaMA-3.1-70B}   & 31 (34.1\%) & 25 (27.5\%) \\
\hdashline
\texttt{Aya-Expanse-8B}  & 28 (30.8\%) & 36 (39.6\%) \\
\texttt{Aya-Expanse-32B} & 53 (58.2\%) & 44 (48.4\%) \\
\bottomrule
\end{tabular}
\caption{Number of statistically significant language pairs per model for social and economic dimensions. Percentages (in parentheses) are relative to the total number of possible language pairs ($n=91$) across 14 languages.}
\label{tab:pairwise_significance}
\end{table}

To further examine these differences, we perform pairwise comparisons between all language pairs for each model using the two-sided Mann–Whitney U test \cite{mann1947test} with Bonferroni correction \cite{armstrong2014use}. This test is well-suited for assessing whether two independent samples differ in distribution, without assuming normality. The results confirm that many language pairs differ significantly in both ideological dimensions, confirming that models exhibit language-specific political behavior. Moreover, \textbf{the number of statistically significant language pairs increases with model size} (Table~\ref{tab:pairwise_significance}), \textbf{suggesting that larger models develop more finely differentiated ideological representations across languages}.

\section{Ideology Steering}

\subsection{Test-time Intervention}

We attempt to shift the ideological leaning of \texttt{LLaMA-3.1-8B} using the test-time intervention method proposed by \citet{li2024inferencetimeinterventionelicitingtruthful}, similar to \citet{kim2025linearrepresentationspoliticalperspective}. This method operates in two stages. In the first stage, we train binary linear probes on top of the output of every attention head to identify those most responsive to a target classification--in our case, distinguishing between liberal and conservative ideologies. For training the probes, we use the \texttt{JyotiNayak/political\_ideologies} dataset from Hugging Face\footnote{\url{https://huggingface.co/datasets/JyotiNayak/political_ideologies}} \cite{lhoest-etal-2021-datasets}, which contains 1,280 samples per class (liberal and conservative) generated by GPT-4 \cite{ai2023gpt}. Each sample consists of a short paragraph (2–3 sentences) labeled with one of the two ideological categories. The dataset covers a diverse range of topics, including economic, environmental, family and gender issues, geopolitics and foreign policy, political institutions, racial justice and immigration, religion, and social, health, and education policies.

In the second stage, we intervene on the $K$ most responsive attention heads at inference time. For each selected head, we modify its output by adding a scaled steering vector, $\alpha \cdot \vec{v}$, where $\alpha$ controls the strength of the intervention. These steering vectors are constructed using a simple \textit{center-of-mass} method\footnote{The center-of-mass method \cite{li2024inferencetimeinterventionelicitingtruthful}, originally related to whitening and coloring transformations \citep{ioffe2015batch, huang2017arbitrary} in deep learning, is equivalent to the DiffMean approach that has emerged in the steering literature \citep{marks2023geometry, wu2025axbench}.}, as opposed to directly using probe weights as in \citet{kim2025linearrepresentationspoliticalperspective}: for each attention head, we compute the mean activation vectors (centers-of-mass) for the liberal and conservative classes, normalize them, and define the direction vector $\vec{v}$ as the normalized difference between these two means. Specifically, for a given attention head $(l, h)$, the steering direction is:

\[
\vec{v}_{l,h} = \frac{\mu^{(1)}_{l,h} - \mu^{(0)}_{l,h}}{\|\mu^{(1)}_{l,h} - \mu^{(0)}_{l,h}\|}
\]

where $\mu^{(c)}_{l,h}$ denotes the mean activation vector for class $c \in \{0,1\}$ (e.g., conservative or liberal) at layer $l$ and head $h$.

At inference time, the value output of the selected head is modified by:

\[
\texttt{value}_{l,h} \mathrel{+}= \alpha \cdot \sigma_{l,h} \cdot \vec{v}_{l,h}
\]

Here, $\sigma_{l,h}$ is the standard deviation of the projection of activations onto the direction vector, used to normalize the intervention across heads with different scales.

\begin{figure}[t]
    \centering
     \includegraphics[width=0.98\linewidth]{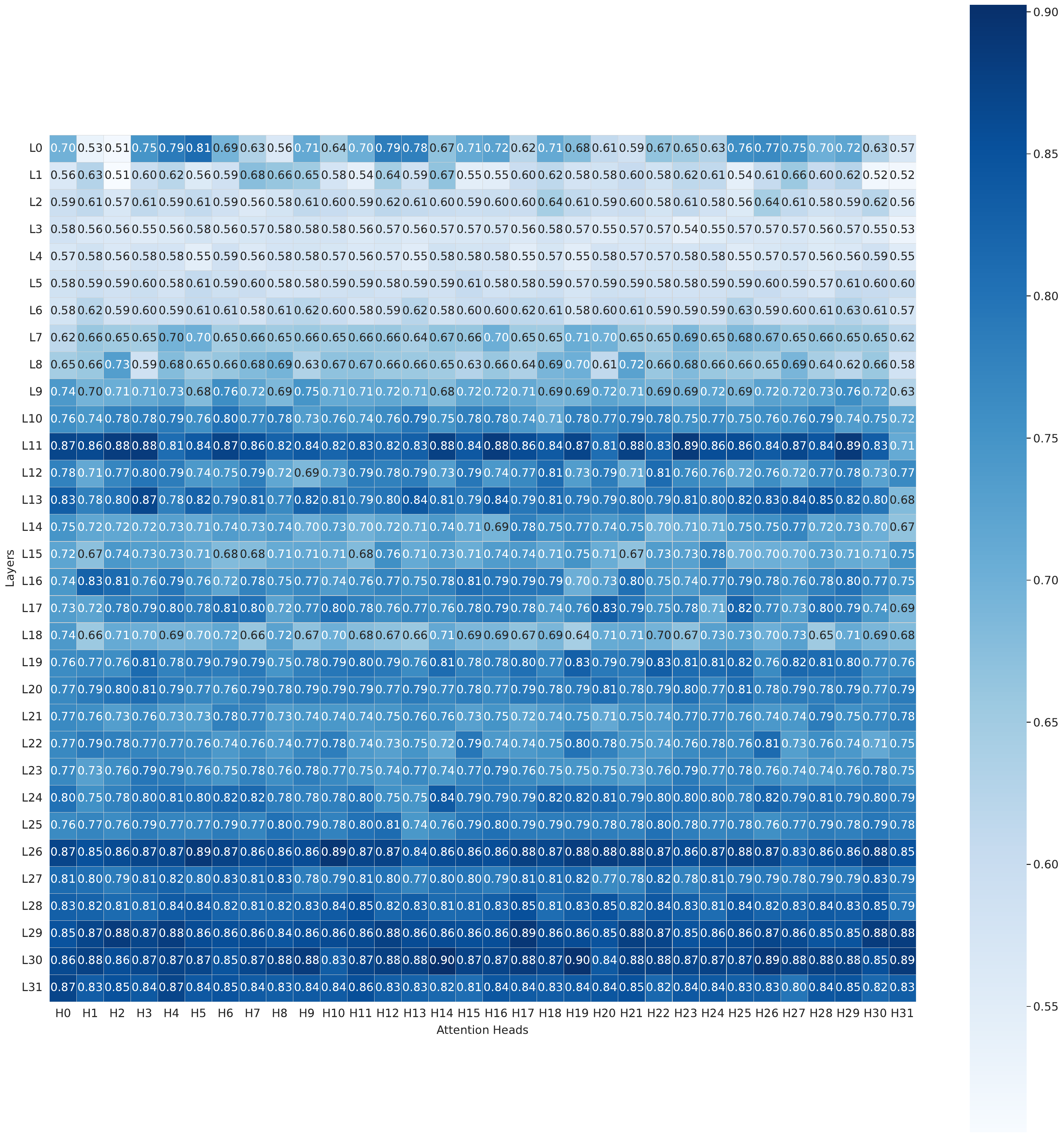}
    \caption{Probes trained for \texttt{LLaMA-3.1-8B}.}
    \label{fig:llama-probe}
\end{figure}

This method leverages the localized semantic capabilities of individual attention heads while remaining minimally invasive to the model’s overall behavior.\footnote{Following \citet{li2024inferencetimeinterventionelicitingtruthful, panickssery2023steering, arditi2024refusal}, interventions on model activations of this type have been shown not to impair general capabilities. Based on this prior evidence, we do not evaluate this aspect explicitly, as our primary goal is to assess whether political bias can be instilled; evaluation of general capabilities is left for future work.} We tune two hyperparameters experimentally: $K$ (the number of heads to intervene on) and $\alpha$ (the intervention strength). To isolate the effects of the intervention, all generation parameters are kept deterministic, as described in Appendix~\ref{app:hyperparameters}.

\subsection{Intervention Results}

\paragraph{Probing:} Figure~\ref{fig:llama-probe} presents the results of probing attention heads in \texttt{LLaMA-3.1-8B}. The highest probe accuracy observed is 0.90, achieved at layer 30, attention head 19. Several other layers--specifically layers 11, 13, and 26 through 31--also exhibit probe accuracies exceeding 0.80, suggesting a concentration of politically informative representations in these middle-to-late layers. These findings align with observations by \citet{kim2025linearrepresentationspoliticalperspective}, who report that politically sensitive features tend to be encoded in mid-to-late transformer layers.

\begin{figure}[t]
    \centering
    \includegraphics[width=0.98\linewidth]{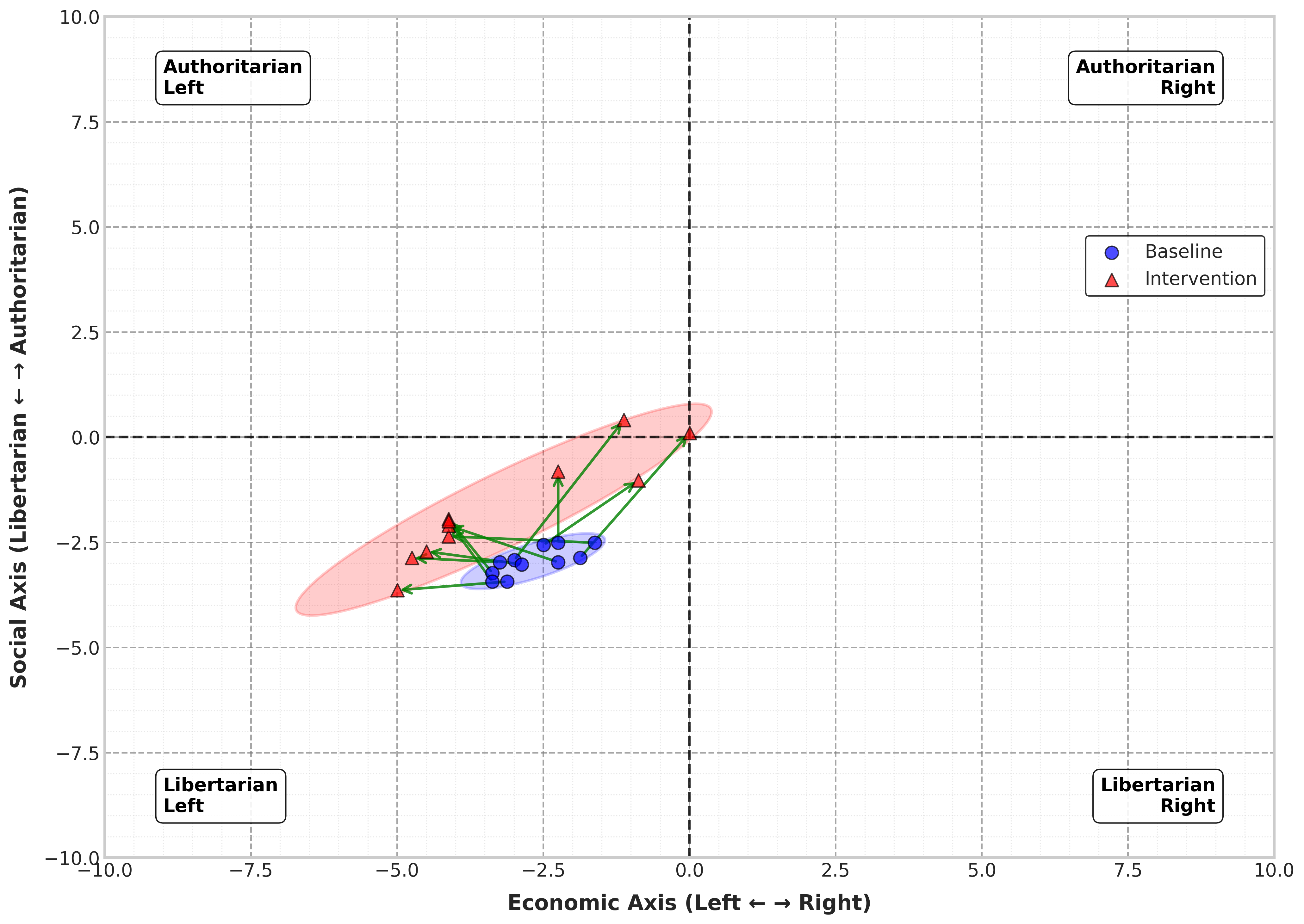}
    \vspace{0.5em}
    \includegraphics[width=0.98\linewidth]{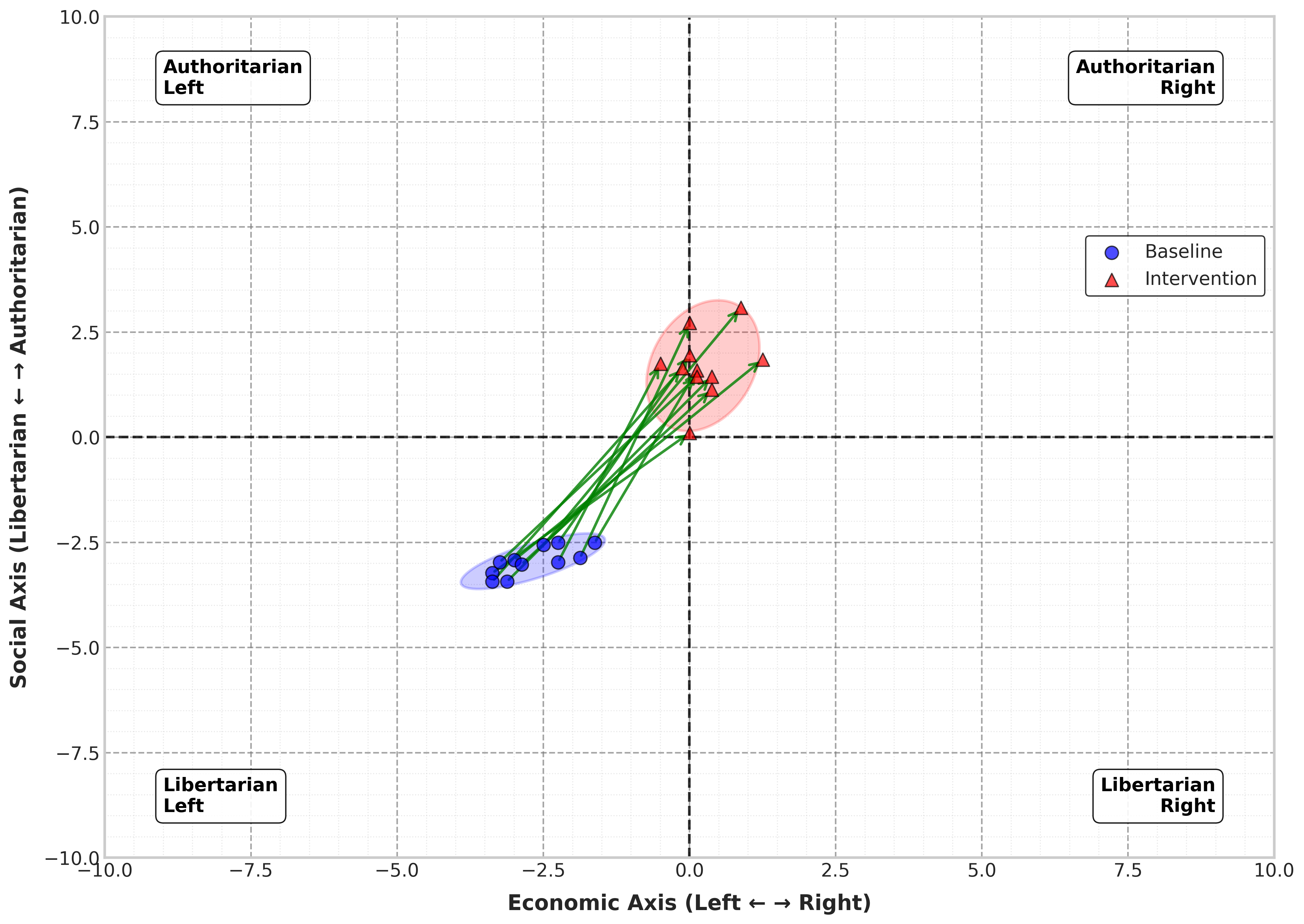}
    \caption{Intervention results for \texttt{LLaMA-3.1-8B} for $K$=512 and $\alpha$=20. Top: direction towards \textit{liberal}. Bottom: direction towards \textit{conservative}.}
    \label{fig:llama-intervene}
\end{figure}

\paragraph{Steering:} Figure~\ref{fig:llama-intervene} illustrates the effect of test-time intervention on English-language prompted \texttt{LLaMA-3.1-8B}. We apply the intervention on the top 512 most responsive attention heads--constituting a half of all available heads--using a steering vector scaled by an intervention strength of $\alpha = 20$. The plot shows the average political compass coordinates across all 11 paraphrased prompts, before and after intervention. \textbf{The intervention reliably shifts model outputs toward the target ideology across all paraphrases}, indicating that targeted manipulation of attention head outputs can steer ideological content in a consistent and interpretable manner. Appendix~\ref{app:inter} provides a comparison across two values of $K$ and three values of $\alpha$.

\begin{table*}[t]
\centering
\small
\resizebox{\textwidth}{!}{
\begin{tabular}{l cc cc cc cc cc}
\toprule
\multirow{3}{*}{\textbf{Config.}} 
& \multicolumn{2}{c}{\makecell{\textbf{en} \\ \small base: 0.0}} 
& \multicolumn{2}{c}{\makecell{\textbf{tr} \\ \small base: 0.6$\pm$0.8}} 
& \multicolumn{2}{c}{\makecell{\textbf{ro} \\ \small base: 2.9$\pm$3.1}} 
& \multicolumn{2}{c}{\makecell{\textbf{sl} \\ \small base: 0.0}} 
& \multicolumn{2}{c}{\makecell{\textbf{fr} \\ \small base: 8.2$\pm$3.9}} \\
\cmidrule(lr){2-3} 
\cmidrule(lr){4-5} 
\cmidrule(lr){6-7} 
\cmidrule(lr){8-9} 
\cmidrule(lr){10-11}
& \textbf{cl. 0} & \textbf{cl. 1} 
& \textbf{cl. 0} & \textbf{cl. 1} 
& \textbf{cl. 0} & \textbf{cl. 1} 
& \textbf{cl. 0} & \textbf{cl. 1} 
& \textbf{cl. 0} & \textbf{cl. 1} \\
\midrule
$\alpha$=20 
& 0.0         & 0.6$\pm$0.8 
& 0.5$\pm$0.5 & 19.8$\pm$5.4 
& 19.5$\pm$5.3 & 2.2$\pm$1.3 
& 0.18$\pm$0.4  & 0.6$\pm$0.7             
& 19.6$\pm$9.1  & 5.2$\pm$2.8     \\
$\alpha$=25 
& 0.0          & 1.4$\pm$1.2 
& 0.09$\pm$0.3 & 32.8$\pm$2.9 
& 25.6$\pm$4.8 & 4.5$\pm$2.9 
& 0.0          & 1.1$\pm$1.5             
& 17.5$\pm$8.8 & 14.5$\pm$5.8     \\
$\alpha$=30 
& 0.0         & 0.7$\pm$0.8 
& 0.0         & 30.2$\pm$6.4 
& 37.1$\pm$6.5 & 34.5$\pm$9.6 
& 0.0          & 11.5$\pm$9.3            
& 10.5$\pm$6.4 & 10.3$\pm$5.5     \\
\bottomrule
\end{tabular}
}
\caption{Average irrelevant response counts by intervention configuration for various languages. $K$ is set to 256 for all $\alpha$ values. Base results are noted under each language. Cl. 0 and 1 refer to steering towards \textit{liberal} and \textit{conservative} directions, respectively.}
\label{tab:intervention_unknowns}
\end{table*}

Motivated by the findings of \citet{wendler-etal-2024-llamas}, who show that multilingual prompts are internally routed through English representations within LLMs, we further test whether steering vectors derived from English texts can generalize cross-linguistically. Specifically, we apply these vectors to model responses in Turkish (tr), Romanian (ro), Slovenian (sl), and French (fr), and find that they are indeed effective in shifting ideological outputs across languages to a certain degree. Detailed results are provided in Appendix~\ref{app:inter}.

Table~\ref{tab:intervention_unknowns} reports the average number of irrelevant or refusal responses before and after intervention. The results indicate that models are more likely to refuse generating valid answers following intervention in non-English languages, with these cross-linguistic differences potentially reflecting variations in pre-training and instruction tuning data sizes across languages.

\section{Discussion}

\subsection{Left-Leaning Bias}

One possible explanation for the observed left-leaning bias in LLMs is the composition of their training data--often drawn from internet-scale corpora such as news media, academic literature, and social platforms--which not only tend to skew liberal, particularly in English-language content \cite{bell2014liberalism, feng2023pretraining}, but also reflect the dominant academic or expert consensus that aligns with progressive views on various sociopolitical issues. Interestingly, the extent of left-leaning behavior appears to correlate with model scale, suggesting that larger models, by virtue of their capacity, may better internalize and reproduce subtle ideological patterns present in their training distributions \cite{exler2025largemeansleftpolitical}.

\subsection{Language-Induced Differences}

Our multilingual evaluation reveals that the language of the prompt has a non-trivial effect on the political stance elicited from LLMs. While prior work by \citet{exler2025largemeansleftpolitical} identified such language-induced bias only in German, our analysis extends this observation to a diverse set of languages. Across all models, English consistently exhibits the most pronounced libertarian-left orientation, while other languages--such as Turkish, Slovenian, and Romanian--yield more centrist or right-leaning responses depending on model size and architecture. This variation may stem from multiple sources, including differences in translation phrasing, cultural priors embedded in training corpora, or model-specific disparities in multilingual capabilities. Importantly, these findings highlight that even in ideologically controlled settings, language choice introduces subtle yet systematic shifts in model behavior, raising questions about fairness and consistency in multilingual deployment contexts.

\subsection{Forcing Ideology Shift}

Our intervention experiments demonstrate that LLMs can be steered toward a desired ideological leaning through test-time modifications of attention head outputs across multiple languages. This suggests that ideological representations are not only linearly decodable but also causally manipulable at the subcomponent level \cite{kim2025linearrepresentationspoliticalperspective}. The consistent shift in outputs across paraphrased prompts supports the hypothesis that certain attention heads play a disproportionately influential role in encoding political perspective. This points to a robust and lightweight approach for aligning models with desired ideological goals--one that does not require full fine-tuning and may serve as an effective first step toward more comprehensive alignment strategies.

\section{Conclusion}

This work provides a comprehensive analysis of political biases in modern instruction-tuned language models across seven models, 14 languages, and 11 prompt variations, demonstrating that political orientations in LLMs are both measurable and manipulable. Our key findings establish three important patterns: larger models consistently exhibit more pronounced libertarian-left leanings compared to smaller counterparts, language choice introduces systematic variations with English eliciting the most libertarian-left orientations, and political orientations can be systematically modified through targeted attention head manipulations using a center-of-mass steering approach. 


\section*{Limitations}

Our study presents several limitations. Certain models (e.g., \texttt{LLaMA-3.1}) refuse to answer some portions of questions despite forced-choice constraints, limiting result completeness. We force responses rather than using free-text evaluation, which may not capture natural model behavior. Our intervention experiments use limited hyperparameter exploration and focus only on \texttt{LLaMA-3.1-8B}, restricting generalizability across architectures. The Political Compass Test, while widely adopted, represents a Western-centric framework that may inadequately capture political orientations across diverse cultural contexts. We evaluate only instruction-tuned models, which undergo extensive alignment that may mask underlying base model biases. The intervention methodology employs a single steering technique with probes trained only on English-language data. Additionally, while we intervene on attention heads in this study, interventions could also be applied to other model components, such as MLP outputs or post-residual activations, which may be more effective and warrant further investigation. Finally, models were trained on data with different temporal cutoffs, making direct comparisons potentially confounded by evolving political discourse.

\section*{Ethics Statement}

This research investigates political biases in language models to promote transparency and responsible AI deployment. While we demonstrate techniques for steering model behavior, we acknowledge the dual-use nature of these methods. The ability to manipulate political orientations in LLMs could be misused to create systems that systematically promote particular ideological viewpoints without user awareness.

We emphasize that our steering methodology should be used exclusively for bias mitigation and alignment research, not for covert political manipulation. The techniques we present require explicit disclosure when deployed in user-facing applications. Our findings highlight the importance of transparency about model biases and the need for robust governance frameworks as these systems become increasingly influential in public discourse.

\section*{Acknowledgments}

This research was partially supported by \textit{DisAI - Improving scientific excellence and creativity in combating disinformation with artificial intelligence and language technologies}, a project funded by Horizon Europe under \href{https://doi.org/10.3030/101079164}{GA No.101079164}, by \textit{lorAI - Low Resource Artificial Intelligence}, a project funded by the European Union under \href{https://doi.org/10.3030/101136646}{GA No.101136646}, and by the German Federal Ministry of Research, Technology and Space (BMFTR) as part of the project TRAILS (01IW24005).

\bibliography{custom}

\begin{thebibliography}{46}
\providecommand{\natexlab}[1]{#1}

\bibitem[{Agiza et~al.(2024)Agiza, Mostagir, and Reda}]{agiza2024polituneanalyzingimpactdata}
Ahmed Agiza, Mohamed Mostagir, and Sherief Reda. 2024.
\newblock \href {https://arxiv.org/abs/2404.08699} {Politune: Analyzing the impact of data selection and fine-tuning on economic and political biases in large language models}.
\newblock \emph{Preprint}, arXiv:2404.08699.

\bibitem[{AI(2023)}]{ai2023gpt}
Open AI. 2023.
\newblock Gpt-4 technical report.
\newblock \emph{arXiv preprint arXiv:2303.08774}.

\bibitem[{Arditi et~al.(2024)Arditi, Obeso, Syed, Paleka, Panickssery, Gurnee, and Nanda}]{arditi2024refusal}
Andy Arditi, Oscar Obeso, Aaquib Syed, Daniel Paleka, Nina Panickssery, Wes Gurnee, and Neel Nanda. 2024.
\newblock Refusal in language models is mediated by a single direction.
\newblock \emph{Advances in Neural Information Processing Systems}, 37:136037--136083.

\bibitem[{Armstrong(2014)}]{armstrong2014use}
Richard~A Armstrong. 2014.
\newblock When to use the b onferroni correction.
\newblock \emph{Ophthalmic and physiological optics}, 34(5):502--508.

\bibitem[{Bell(2014)}]{bell2014liberalism}
Duncan Bell. 2014.
\newblock What is liberalism?
\newblock \emph{Political theory}, 42(6):682--715.

\bibitem[{Bender et~al.(2021)Bender, Gebru, McMillan-Major, and Shmitchell}]{bender2021dangers}
Emily~M Bender, Timnit Gebru, Angelina McMillan-Major, and Shmargaret Shmitchell. 2021.
\newblock On the dangers of stochastic parrots: Can language models be too big?
\newblock In \emph{Proceedings of the 2021 ACM conference on fairness, accountability, and transparency}, pages 610--623.

\bibitem[{Brown et~al.(2020)Brown, Mann, Ryder, Subbiah, Kaplan, Dhariwal, Neelakantan, Shyam, Sastry, Askell et~al.}]{brown2020language}
Tom Brown, Benjamin Mann, Nick Ryder, Melanie Subbiah, Jared~D Kaplan, Prafulla Dhariwal, Arvind Neelakantan, Pranav Shyam, Girish Sastry, Amanda Askell, and 1 others. 2020.
\newblock Language models are few-shot learners.
\newblock \emph{Advances in neural information processing systems}, 33:1877--1901.

\bibitem[{Ceron et~al.(2024)Ceron, Falk, Bari{\'c}, Nikolaev, and Pad{\'o}}]{ceron2024beyond}
Tanise Ceron, Neele Falk, Ana Bari{\'c}, Dmitry Nikolaev, and Sebastian Pad{\'o}. 2024.
\newblock Beyond prompt brittleness: Evaluating the reliability and consistency of political worldviews in llms.
\newblock \emph{Transactions of the Association for Computational Linguistics}, 12:1378--1400.

\bibitem[{ChatGPT(2022)}]{chatgpt2022optimizing}
OpenAI ChatGPT. 2022.
\newblock Optimizing language models for dialogue.
\newblock \emph{OpenAI. com}, 30.

\bibitem[{Chu et~al.(2025)Chu, Wang, Xie, Zhu, Yan, Ye, Zhong, Hu, Liang, Yu, and Wen}]{chu2025llmagentseducationadvances}
Zhendong Chu, Shen Wang, Jian Xie, Tinghui Zhu, Yibo Yan, Jinheng Ye, Aoxiao Zhong, Xuming Hu, Jing Liang, Philip~S. Yu, and Qingsong Wen. 2025.
\newblock \href {https://arxiv.org/abs/2503.11733} {Llm agents for education: Advances and applications}.
\newblock \emph{Preprint}, arXiv:2503.11733.

\bibitem[{Conneau et~al.(2018)Conneau, Lample, Rinott, Williams, Bowman, Schwenk, and Stoyanov}]{conneau2018xnli}
Alexis Conneau, Guillaume Lample, Ruty Rinott, Adina Williams, Samuel~R Bowman, Holger Schwenk, and Veselin Stoyanov. 2018.
\newblock Xnli: Evaluating cross-lingual sentence representations.
\newblock \emph{arXiv preprint arXiv:1809.05053}.

\bibitem[{Dang et~al.(2024)Dang, Singh, D'souza, Ahmadian, Salamanca, Smith, Peppin, Hong, Govindassamy, Zhao et~al.}]{dang2024aya}
John Dang, Shivalika Singh, Daniel D'souza, Arash Ahmadian, Alejandro Salamanca, Madeline Smith, Aidan Peppin, Sungjin Hong, Manoj Govindassamy, Terrence Zhao, and 1 others. 2024.
\newblock Aya expanse: Combining research breakthroughs for a new multilingual frontier.
\newblock \emph{arXiv preprint arXiv:2412.04261}.

\bibitem[{Devlin et~al.(2019)Devlin, Chang, Lee, and Toutanova}]{devlin2019bert}
Jacob Devlin, Ming-Wei Chang, Kenton Lee, and Kristina Toutanova. 2019.
\newblock Bert: Pre-training of deep bidirectional transformers for language understanding.
\newblock In \emph{Proceedings of the 2019 conference of the North American chapter of the association for computational linguistics: human language technologies, volume 1 (long and short papers)}, pages 4171--4186.

\bibitem[{Exler et~al.(2025)Exler, Schutera, Reischl, and Rettenberger}]{exler2025largemeansleftpolitical}
David Exler, Mark Schutera, Markus Reischl, and Luca Rettenberger. 2025.
\newblock \href {https://arxiv.org/abs/2505.04393} {Large means left: Political bias in large language models increases with their number of parameters}.
\newblock \emph{Preprint}, arXiv:2505.04393.

\bibitem[{Feng et~al.(2023)Feng, Park, Liu, and Tsvetkov}]{feng2023pretraining}
Shangbin Feng, Chan~Young Park, Yuhan Liu, and Yulia Tsvetkov. 2023.
\newblock From pretraining data to language models to downstream tasks: Tracking the trails of political biases leading to unfair nlp models.
\newblock \emph{arXiv preprint arXiv:2305.08283}.

\bibitem[{Gallegos et~al.(2024)Gallegos, Rossi, Barrow, Tanjim, Kim, Dernoncourt, Yu, Zhang, and Ahmed}]{gallegos2024bias}
Isabel~O Gallegos, Ryan~A Rossi, Joe Barrow, Md~Mehrab Tanjim, Sungchul Kim, Franck Dernoncourt, Tong Yu, Ruiyi Zhang, and Nesreen~K Ahmed. 2024.
\newblock Bias and fairness in large language models: A survey.
\newblock \emph{Computational Linguistics}, 50(3):1097--1179.

\bibitem[{Grattafiori et~al.(2024)Grattafiori, Dubey, Jauhri, Pandey, Kadian, Al-Dahle, Letman, Mathur, Schelten, Vaughan et~al.}]{grattafiori2024llama}
Aaron Grattafiori, Abhimanyu Dubey, Abhinav Jauhri, Abhinav Pandey, Abhishek Kadian, Ahmad Al-Dahle, Aiesha Letman, Akhil Mathur, Alan Schelten, Alex Vaughan, and 1 others. 2024.
\newblock The llama 3 herd of models.
\newblock \emph{arXiv e-prints}, pages arXiv--2407.

\bibitem[{Hartmann et~al.(2023)Hartmann, Schwenzow, and Witte}]{hartmann2023political}
Jochen Hartmann, Jasper Schwenzow, and Maximilian Witte. 2023.
\newblock The political ideology of conversational ai: Converging evidence on chatgpt's pro-environmental, left-libertarian orientation.
\newblock \emph{arXiv preprint arXiv:2301.01768}.

\bibitem[{Huang and Belongie(2017)}]{huang2017arbitrary}
Xun Huang and Serge Belongie. 2017.
\newblock Arbitrary style transfer in real-time with adaptive instance normalization.
\newblock In \emph{Proceedings of the IEEE international conference on computer vision}, pages 1501--1510.

\bibitem[{Ioffe and Szegedy(2015)}]{ioffe2015batch}
Sergey Ioffe and Christian Szegedy. 2015.
\newblock Batch normalization: Accelerating deep network training by reducing internal covariate shift.
\newblock In \emph{International conference on machine learning}, pages 448--456. pmlr.

\bibitem[{Jiang et~al.(2023)Jiang, Sablayrolles, Mensch, Bamford, Chaplot, de~las Casas, Bressand, Lengyel, Lample, Saulnier, Lavaud, Lachaux, Stock, Scao, Lavril, Wang, Lacroix, and Sayed}]{jiang2023mistral7b}
Albert~Q. Jiang, Alexandre Sablayrolles, Arthur Mensch, Chris Bamford, Devendra~Singh Chaplot, Diego de~las Casas, Florian Bressand, Gianna Lengyel, Guillaume Lample, Lucile Saulnier, Lélio~Renard Lavaud, Marie-Anne Lachaux, Pierre Stock, Teven~Le Scao, Thibaut Lavril, Thomas Wang, Timothée Lacroix, and William~El Sayed. 2023.
\newblock \href {https://arxiv.org/abs/2310.06825} {Mistral 7b}.
\newblock \emph{Preprint}, arXiv:2310.06825.

\bibitem[{Kim et~al.(2025)Kim, Evans, and Schein}]{kim2025linearrepresentationspoliticalperspective}
Junsol Kim, James Evans, and Aaron Schein. 2025.
\newblock \href {https://arxiv.org/abs/2503.02080} {Linear representations of political perspective emerge in large language models}.
\newblock \emph{Preprint}, arXiv:2503.02080.

\bibitem[{Kruskal and Wallis(1952)}]{kruskal1952use}
William~H Kruskal and W~Allen Wallis. 1952.
\newblock Use of ranks in one-criterion variance analysis.
\newblock \emph{Journal of the American statistical Association}, 47(260):583--621.

\bibitem[{Lewis et~al.(2019)Lewis, Liu, Goyal, Ghazvininejad, Mohamed, Levy, Stoyanov, and Zettlemoyer}]{lewis2019bart}
Mike Lewis, Yinhan Liu, Naman Goyal, Marjan Ghazvininejad, Abdelrahman Mohamed, Omer Levy, Ves Stoyanov, and Luke Zettlemoyer. 2019.
\newblock Bart: Denoising sequence-to-sequence pre-training for natural language generation, translation, and comprehension.
\newblock \emph{arXiv preprint arXiv:1910.13461}.

\bibitem[{Lhoest et~al.(2021)Lhoest, Villanova~del Moral, Jernite, Thakur, von Platen, Patil, Chaumond, Drame, Plu, Tunstall, Davison, {\v{S}}a{\v{s}}ko, Chhablani, Malik, Brandeis, Le~Scao, Sanh, Xu, Patry, McMillan-Major, Schmid, Gugger, Delangue, Matussi{\`e}re, Debut, Bekman, Cistac, Goehringer, Mustar, Lagunas, Rush, and Wolf}]{lhoest-etal-2021-datasets}
Quentin Lhoest, Albert Villanova~del Moral, Yacine Jernite, Abhishek Thakur, Patrick von Platen, Suraj Patil, Julien Chaumond, Mariama Drame, Julien Plu, Lewis Tunstall, Joe Davison, Mario {\v{S}}a{\v{s}}ko, Gunjan Chhablani, Bhavitvya Malik, Simon Brandeis, Teven Le~Scao, Victor Sanh, Canwen Xu, Nicolas Patry, and 13 others. 2021.
\newblock \href {https://arxiv.org/abs/2109.02846} {Datasets: A community library for natural language processing}.
\newblock In \emph{Proceedings of the 2021 Conference on Empirical Methods in Natural Language Processing: System Demonstrations}, pages 175--184, Online and Punta Cana, Dominican Republic. Association for Computational Linguistics.

\bibitem[{Li et~al.(2024)Li, Patel, Viégas, Pfister, and Wattenberg}]{li2024inferencetimeinterventionelicitingtruthful}
Kenneth Li, Oam Patel, Fernanda Viégas, Hanspeter Pfister, and Martin Wattenberg. 2024.
\newblock \href {https://arxiv.org/abs/2306.03341} {Inference-time intervention: Eliciting truthful answers from a language model}.
\newblock \emph{Preprint}, arXiv:2306.03341.

\bibitem[{Liu et~al.(2019)Liu, Ott, Goyal, Du, Joshi, Chen, Levy, Lewis, Zettlemoyer, and Stoyanov}]{liu2019roberta}
Yinhan Liu, Myle Ott, Naman Goyal, Jingfei Du, Mandar Joshi, Danqi Chen, Omer Levy, Mike Lewis, Luke Zettlemoyer, and Veselin Stoyanov. 2019.
\newblock Roberta: A robustly optimized bert pretraining approach.
\newblock \emph{arXiv preprint arXiv:1907.11692}.

\bibitem[{Mann and Whitney(1947)}]{mann1947test}
Henry~B Mann and Donald~R Whitney. 1947.
\newblock On a test of whether one of two random variables is stochastically larger than the other.
\newblock \emph{The annals of mathematical statistics}, pages 50--60.

\bibitem[{Marks and Tegmark(2023)}]{marks2023geometry}
Samuel Marks and Max Tegmark. 2023.
\newblock The geometry of truth: Emergent linear structure in large language model representations of true/false datasets.
\newblock \emph{arXiv preprint arXiv:2310.06824}.

\bibitem[{Motoki et~al.(2024)Motoki, Pinho~Neto, and Rodrigues}]{motoki2024more}
Fabio Motoki, Valdemar Pinho~Neto, and Victor Rodrigues. 2024.
\newblock More human than human: measuring chatgpt political bias.
\newblock \emph{Public Choice}, 198(1):3--23.

\bibitem[{Ong et~al.(2024)Ong, Jin, Elangovan, Lim, Lim, Sng, Ke, Tung, Zhong, Koh, Lee, Chen, Chng, Than, Goh, and Ting}]{ong2024developmenttestingnovellarge}
Jasmine Chiat~Ling Ong, Liyuan Jin, Kabilan Elangovan, Gilbert Yong~San Lim, Daniel Yan~Zheng Lim, Gerald Gui~Ren Sng, Yuhe Ke, Joshua Yi~Min Tung, Ryan~Jian Zhong, Christopher Ming~Yao Koh, Keane Zhi~Hao Lee, Xiang Chen, Jack~Kian Chng, Aung Than, Ken~Junyang Goh, and Daniel Shu~Wei Ting. 2024.
\newblock \href {https://arxiv.org/abs/2402.01741} {Development and testing of a novel large language model-based clinical decision support systems for medication safety in 12 clinical specialties}.
\newblock \emph{Preprint}, arXiv:2402.01741.

\bibitem[{Panickssery et~al.(2023)Panickssery, Gabrieli, Schulz, Tong, Hubinger, and Turner}]{panickssery2023steering}
Nina Panickssery, Nick Gabrieli, Julian Schulz, Meg Tong, Evan Hubinger, and Alexander~Matt Turner. 2023.
\newblock Steering llama 2 via contrastive activation addition.
\newblock \emph{arXiv preprint arXiv:2312.06681}.

\bibitem[{Poole(2005)}]{poole2005spatial}
Keith~T Poole. 2005.
\newblock \emph{Spatial models of parliamentary voting}.
\newblock Cambridge University Press.

\bibitem[{Poole and Rosenthal(1985)}]{poole1985spatial}
Keith~T Poole and Howard Rosenthal. 1985.
\newblock A spatial model for legislative roll call analysis.
\newblock \emph{American journal of political science}, pages 357--384.

\bibitem[{Radford et~al.(2019)Radford, Wu, Child, Luan, Amodei, Sutskever et~al.}]{radford2019language}
Alec Radford, Jeffrey Wu, Rewon Child, David Luan, Dario Amodei, Ilya Sutskever, and 1 others. 2019.
\newblock Language models are unsupervised multitask learners.
\newblock \emph{OpenAI blog}, 1(8):9.

\bibitem[{R{\"o}ttger et~al.(2024)R{\"o}ttger, Hofmann, Pyatkin, Hinck, Kirk, Sch{\"u}tze, and Hovy}]{rottger2024political}
Paul R{\"o}ttger, Valentin Hofmann, Valentina Pyatkin, Musashi Hinck, Hannah~Rose Kirk, Hinrich Sch{\"u}tze, and Dirk Hovy. 2024.
\newblock Political compass or spinning arrow? towards more meaningful evaluations for values and opinions in large language models.
\newblock \emph{arXiv preprint arXiv:2402.16786}.

\bibitem[{Rozado(2024)}]{rozado2024political}
David Rozado. 2024.
\newblock The political preferences of llms.
\newblock \emph{PloS one}, 19(7):e0306621.

\bibitem[{Rutinowski et~al.(2024)Rutinowski, Franke, Endendyk, Dormuth, Roidl, and Pauly}]{rutinowski2024self}
J{\'e}r{\^o}me Rutinowski, Sven Franke, Jan Endendyk, Ina Dormuth, Moritz Roidl, and Markus Pauly. 2024.
\newblock The self-perception and political biases of chatgpt.
\newblock \emph{Human Behavior and Emerging Technologies}, 2024(1):7115633.

\bibitem[{Santurkar et~al.(2023)Santurkar, Durmus, Ladhak, Lee, Liang, and Hashimoto}]{santurkar2023whose}
Shibani Santurkar, Esin Durmus, Faisal Ladhak, Cinoo Lee, Percy Liang, and Tatsunori Hashimoto. 2023.
\newblock Whose opinions do language models reflect?
\newblock In \emph{International Conference on Machine Learning}, pages 29971--30004. PMLR.

\bibitem[{Touvron et~al.(2023)Touvron, Martin, Stone, Albert, Almahairi, Babaei, Bashlykov, Batra, Bhargava, Bhosale et~al.}]{touvron2023llama}
Hugo Touvron, Louis Martin, Kevin Stone, Peter Albert, Amjad Almahairi, Yasmine Babaei, Nikolay Bashlykov, Soumya Batra, Prajjwal Bhargava, Shruti Bhosale, and 1 others. 2023.
\newblock Llama 2: Open foundation and fine-tuned chat models.
\newblock \emph{arXiv preprint arXiv:2307.09288}.

\bibitem[{Trhlik and Stenetorp(2024)}]{trhlik2024quantifyinggenerativemediabias}
Filip Trhlik and Pontus Stenetorp. 2024.
\newblock \href {https://arxiv.org/abs/2406.10773} {Quantifying generative media bias with a corpus of real-world and generated news articles}.
\newblock \emph{Preprint}, arXiv:2406.10773.

\bibitem[{Weidinger et~al.(2021)Weidinger, Mellor, Rauh, Griffin, Uesato, Huang, Cheng, Glaese, Balle, Kasirzadeh, Kenton, Brown, Hawkins, Stepleton, Biles, Birhane, Haas, Rimell, Hendricks, Isaac, Legassick, Irving, and Gabriel}]{weidinger2021ethicalsocialrisksharm}
Laura Weidinger, John Mellor, Maribeth Rauh, Conor Griffin, Jonathan Uesato, Po-Sen Huang, Myra Cheng, Mia Glaese, Borja Balle, Atoosa Kasirzadeh, Zac Kenton, Sasha Brown, Will Hawkins, Tom Stepleton, Courtney Biles, Abeba Birhane, Julia Haas, Laura Rimell, Lisa~Anne Hendricks, and 4 others. 2021.
\newblock \href {https://arxiv.org/abs/2112.04359} {Ethical and social risks of harm from language models}.
\newblock \emph{Preprint}, arXiv:2112.04359.

\bibitem[{Wendler et~al.(2024)Wendler, Veselovsky, Monea, and West}]{wendler-etal-2024-llamas}
Chris Wendler, Veniamin Veselovsky, Giovanni Monea, and Robert West. 2024.
\newblock \href {https://doi.org/10.18653/v1/2024.acl-long.820} {Do llamas work in {E}nglish? on the latent language of multilingual transformers}.
\newblock In \emph{Proceedings of the 62nd Annual Meeting of the Association for Computational Linguistics (Volume 1: Long Papers)}, pages 15366--15394, Bangkok, Thailand. Association for Computational Linguistics.

\bibitem[{Wu et~al.(2025)Wu, Arora, Geiger, Wang, Huang, Jurafsky, Manning, and Potts}]{wu2025axbench}
Zhengxuan Wu, Aryaman Arora, Atticus Geiger, Zheng Wang, Jing Huang, Dan Jurafsky, Christopher~D Manning, and Christopher Potts. 2025.
\newblock Axbench: Steering llms? even simple baselines outperform sparse autoencoders.
\newblock \emph{arXiv preprint arXiv:2501.17148}.

\bibitem[{Xiong et~al.(2024)Xiong, Bian, Li, Li, Du, Wang, Yin, and Helal}]{xiong2024searchengineservicesmeet}
Haoyi Xiong, Jiang Bian, Yuchen Li, Xuhong Li, Mengnan Du, Shuaiqiang Wang, Dawei Yin, and Sumi Helal. 2024.
\newblock \href {https://arxiv.org/abs/2407.00128} {When search engine services meet large language models: Visions and challenges}.
\newblock \emph{Preprint}, arXiv:2407.00128.

\bibitem[{Yang et~al.(2025)Yang, Li, Yang, Zhang, Hui, Zheng, Yu, Gao, Huang, Lv et~al.}]{yang2025qwen3}
An~Yang, Anfeng Li, Baosong Yang, Beichen Zhang, Binyuan Hui, Bo~Zheng, Bowen Yu, Chang Gao, Chengen Huang, Chenxu Lv, and 1 others. 2025.
\newblock Qwen3 technical report.
\newblock \emph{arXiv preprint arXiv:2505.09388}.

\end{thebibliography}

\appendix
\onecolumn
\section*{Appendix}

\section{Generation Hyperparameters}
\label{app:hyperparameters}

\paragraph{Ideology Identification.}
For the ideological classification experiments (e.g., locating models on the political compass), we use the following decoding parameters for generation:
\begin{itemize}
    \item \textbf{Temperature:} 0.7
    \item \textbf{Top-p:} 0.9
    \item \textbf{Maximum tokens:} 256
    \item \textbf{Sampling:} Enabled
    \item \textbf{Skip special tokens:} True
    \item \textbf{Random seed:} 42
\end{itemize}

\paragraph{Intervention and Baseline.}
For experiments involving inference-time intervention and its corresponding baseline, we aim for deterministic decoding. Therefore, we use:
\begin{itemize}
    \item \textbf{Temperature:} 0
    \item \textbf{Top-p:} (not used)
    \item \textbf{Sampling:} Disabled (\texttt{do\_sample = False})
    \item \textbf{Maximum tokens:} 100
    \item \textbf{Skip special tokens:} True
    \item \textbf{Random seed:} 42
\end{itemize}

This configuration ensures consistent output length and behavior, which is important for isolating the effect of the interventions.

\begin{figure*}[h]
\section{Political Compass Results}
\label{app:results}
  \centering
  \begin{subfigure}[t]{0.48\textwidth}
    \centering
    \includegraphics[width=\linewidth]{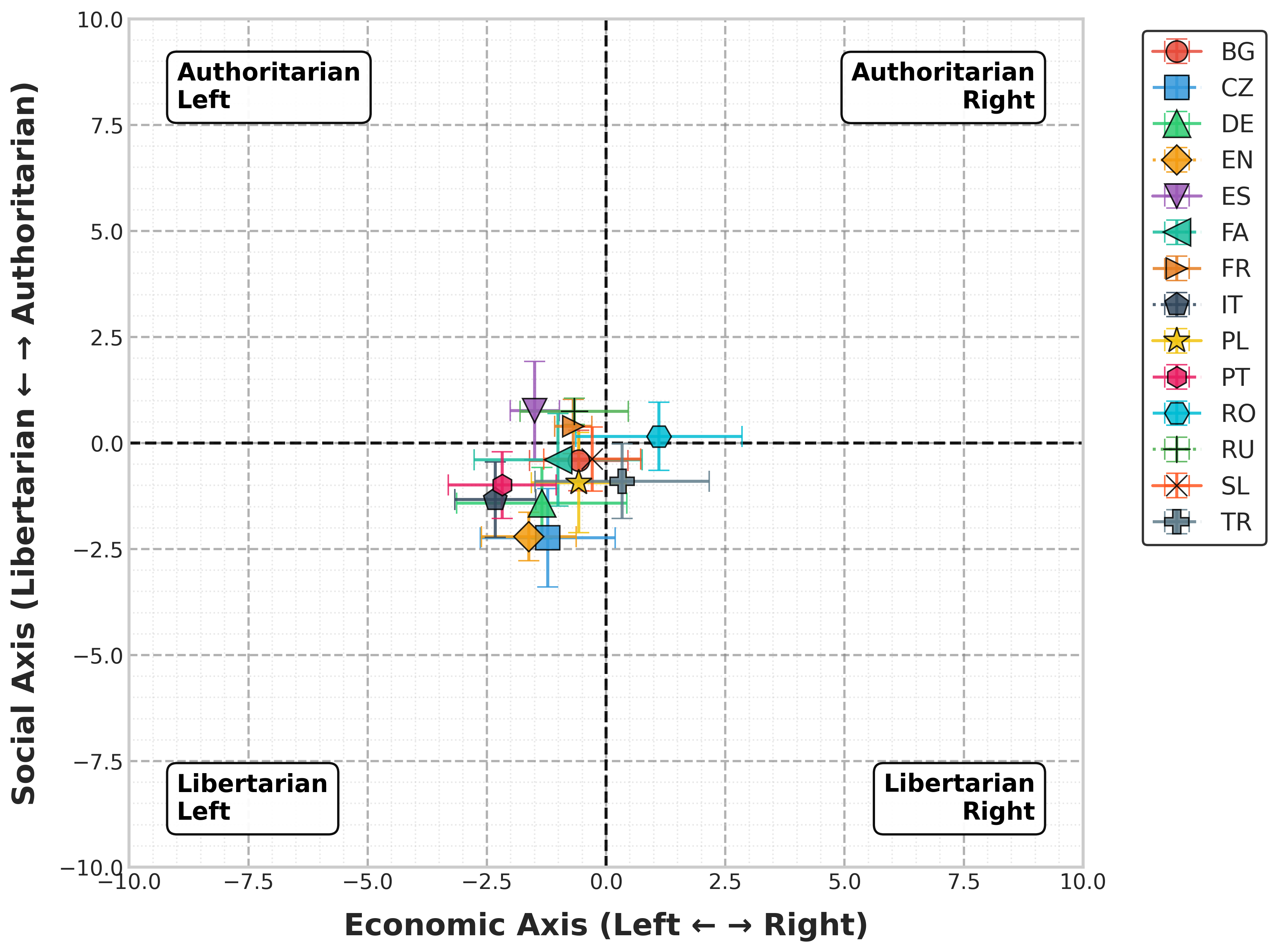}
    \caption{\texttt{LLaMA-3.1 8B}}
    \label{fig:llama-8}
  \end{subfigure}
  \hspace{0.02\textwidth}
  \begin{subfigure}[t]{0.48\textwidth}
    \centering
    \includegraphics[width=\linewidth]{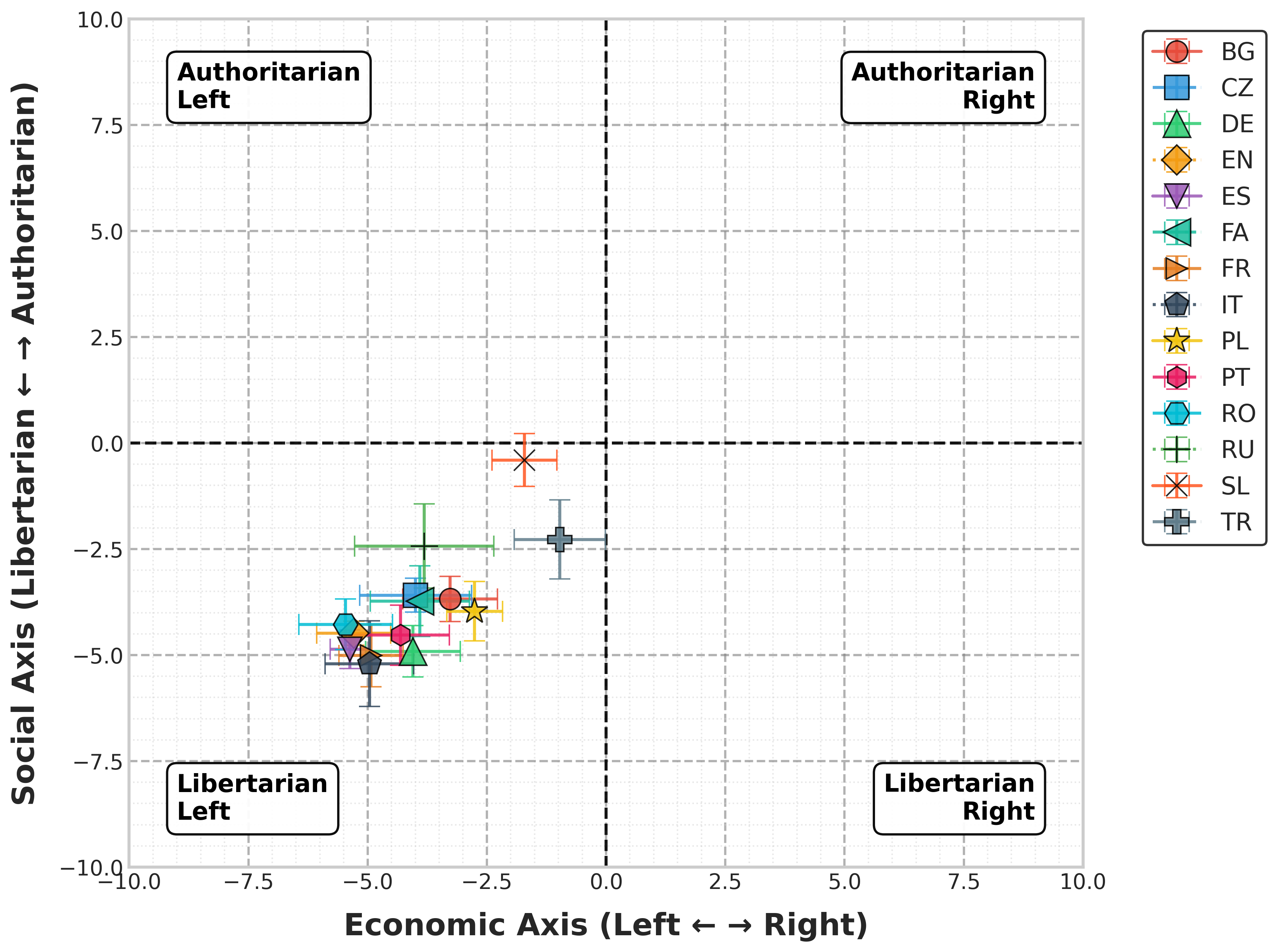}
    \caption{\texttt{LLaMA-3.1 70B}}
    \label{fig:llama-70}
  \end{subfigure}

  \vspace{0.5em}

  \begin{subfigure}[t]{0.48\textwidth}
    \centering
    \includegraphics[width=\linewidth]{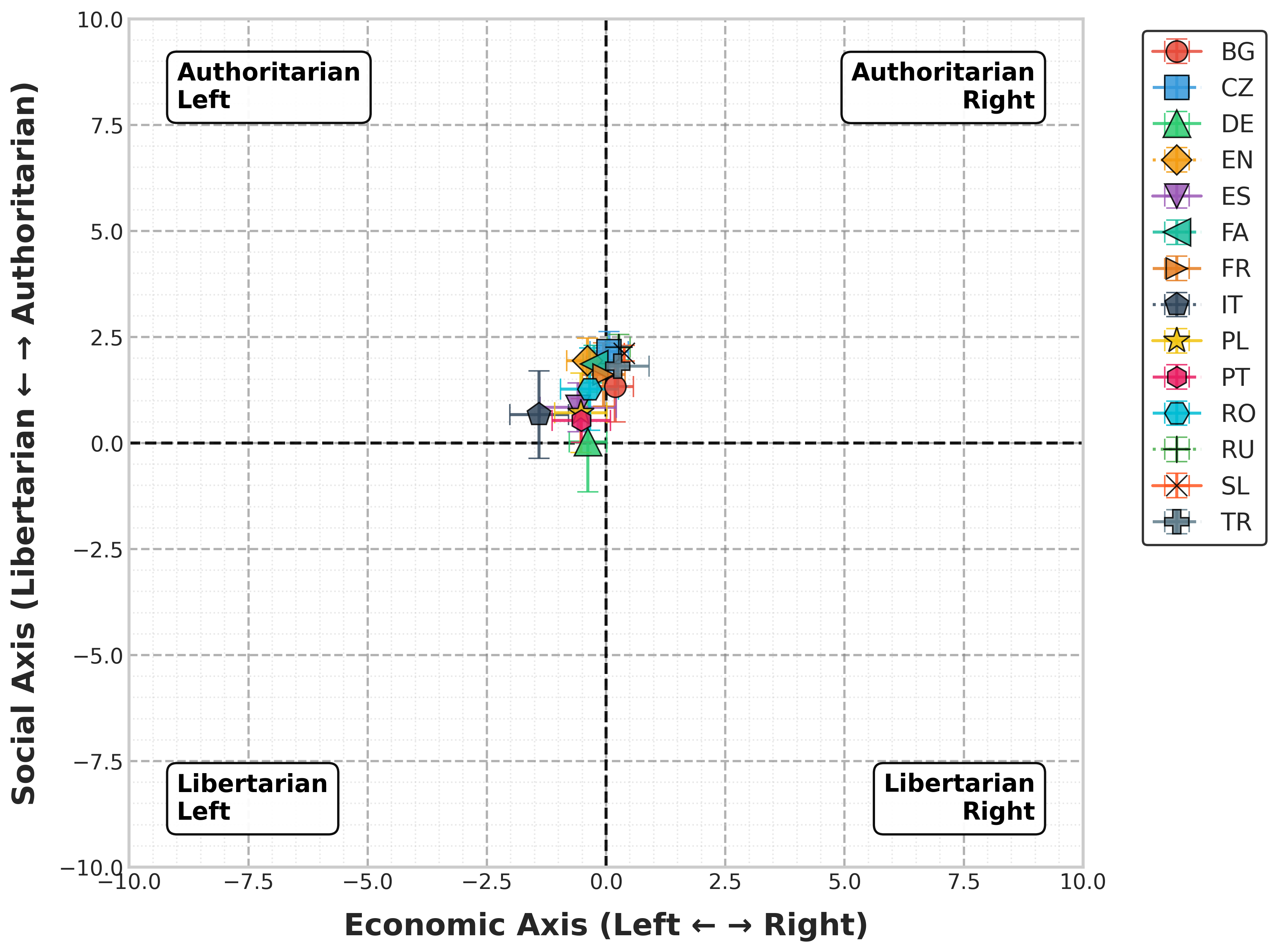}
    \caption{\texttt{Qwen-3 8B}}
    \label{fig:qwen-8}
  \end{subfigure}
  \hspace{0.02\textwidth}
  \begin{subfigure}[t]{0.48\textwidth}
    \centering
    \includegraphics[width=\linewidth]{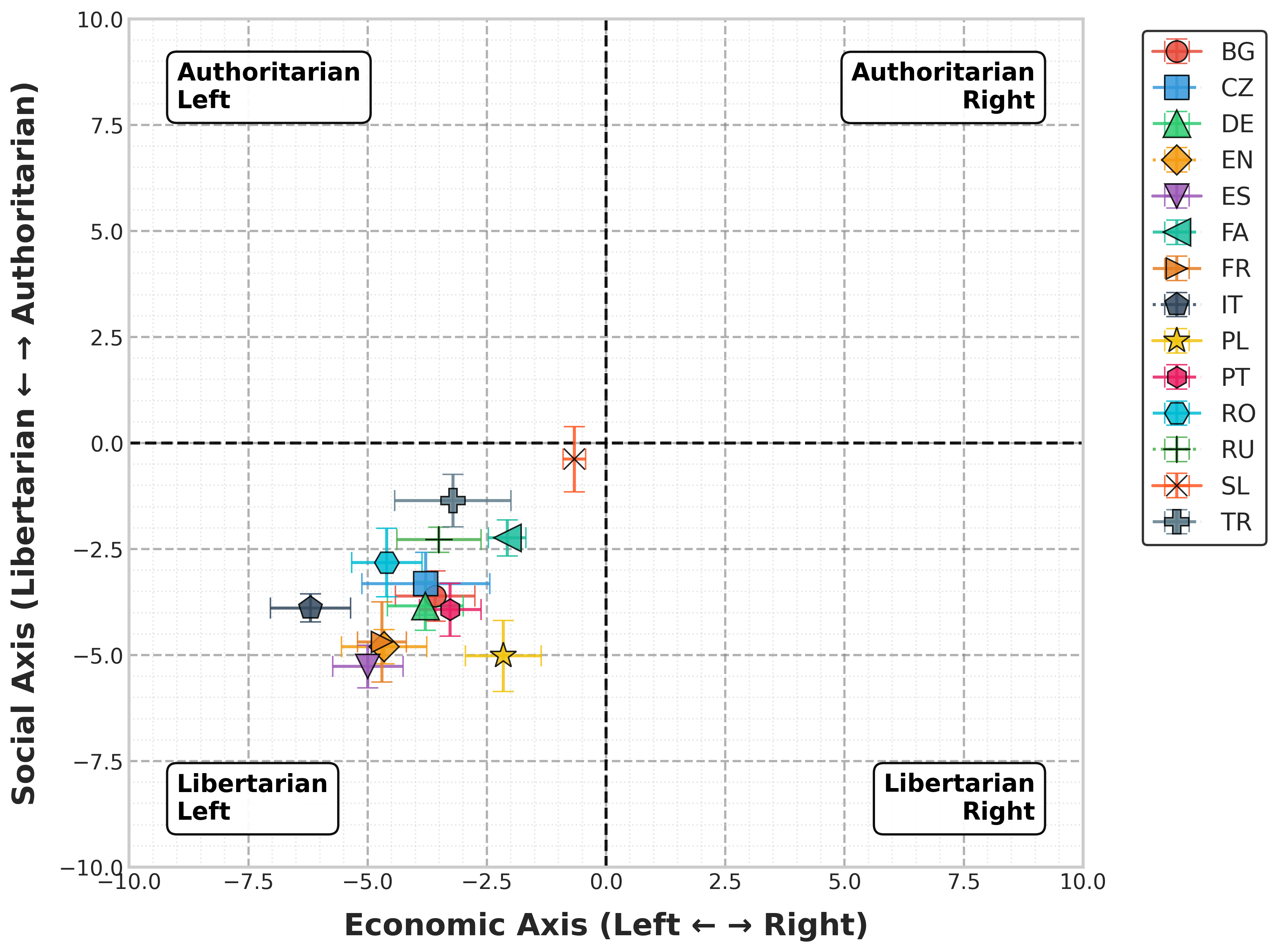}
    \caption{\texttt{Qwen-3 14B}}
    \label{fig:qwen-14}
  \end{subfigure}

  \vspace{0.5em}

  \begin{subfigure}[t]{0.48\textwidth}
    \centering
    \includegraphics[width=\linewidth]{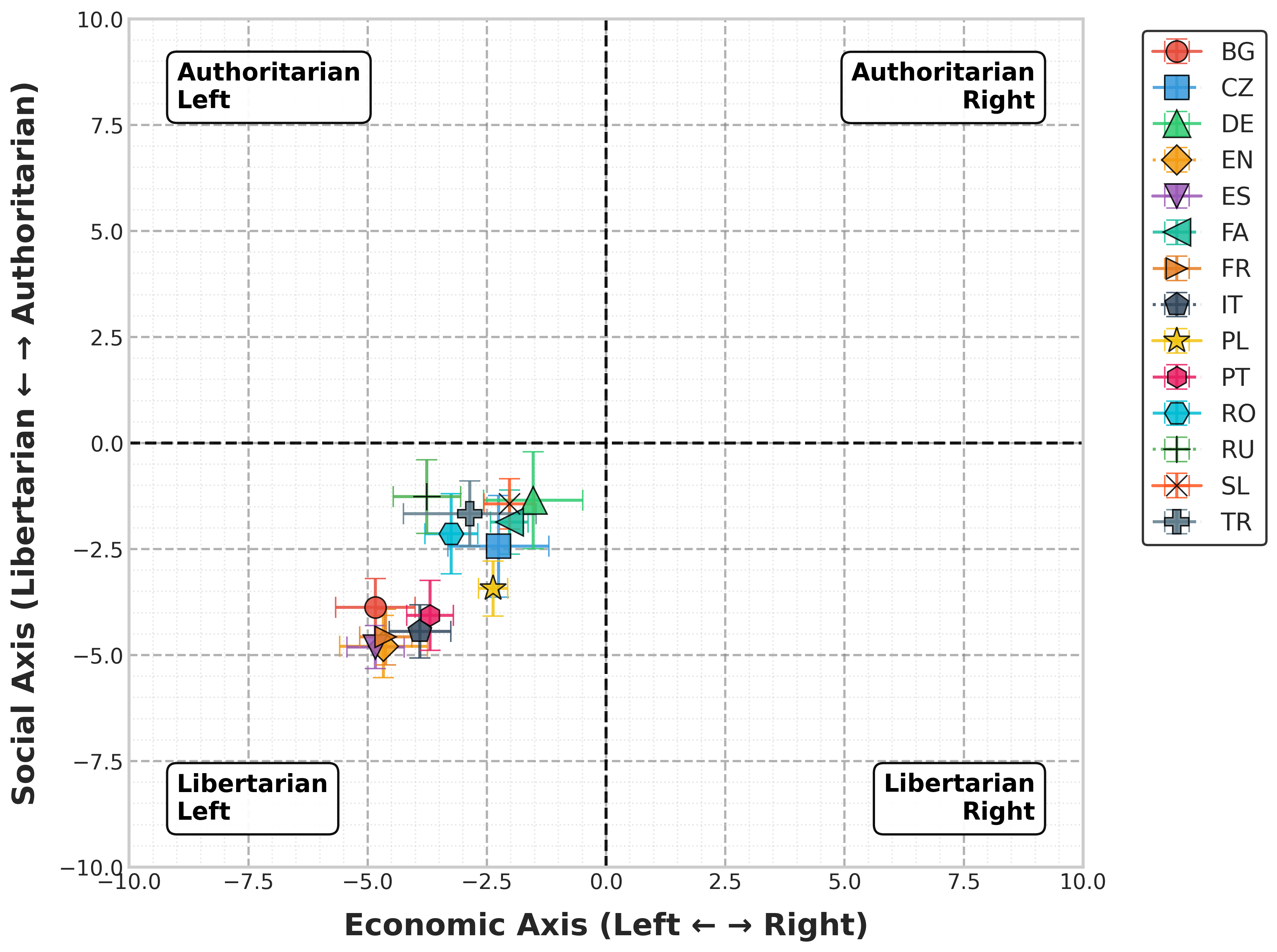}
    \caption{\texttt{Qwen-3 32B}}
    \label{fig:qwen-32}
  \end{subfigure}

  \caption{Political compass results across the rest of the models of various sizes. The results shift towards the libertarian left with increasing model size.}
  \label{fig:all_results}
\end{figure*}
\clearpage

\begin{figure*}[h]
\section{Political Intervention Results}
\label{app:inter}

  \centering
  \begin{subfigure}[t]{0.48\textwidth}
    \centering
    \includegraphics[width=\linewidth]{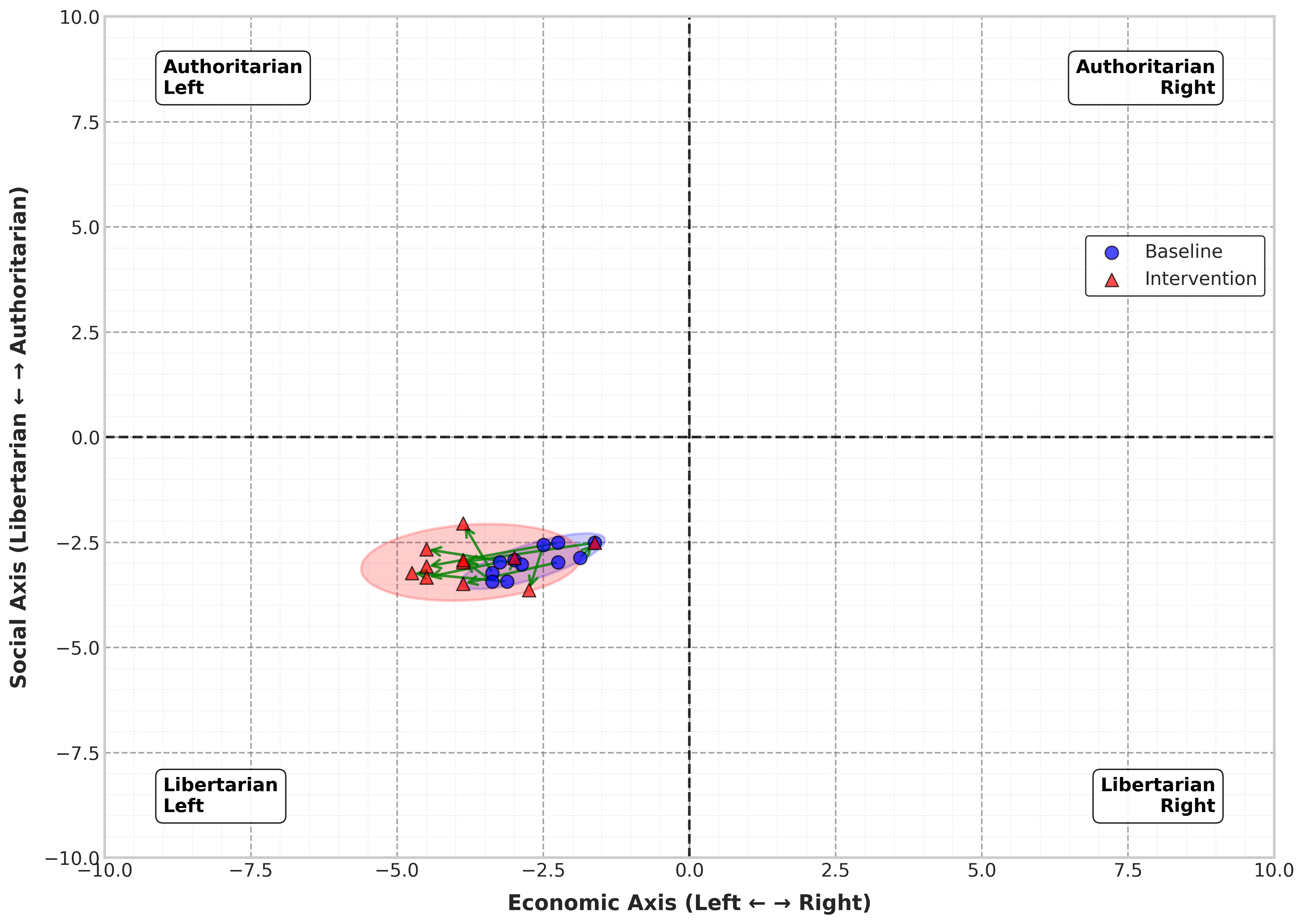}
    \caption{\texttt{Intervention Strength of 20}}
    \label{fig:inter_20_1}
  \end{subfigure}
  \hspace{0.02\textwidth}
  \begin{subfigure}[t]{0.48\textwidth}
    \centering
    \includegraphics[width=\linewidth]{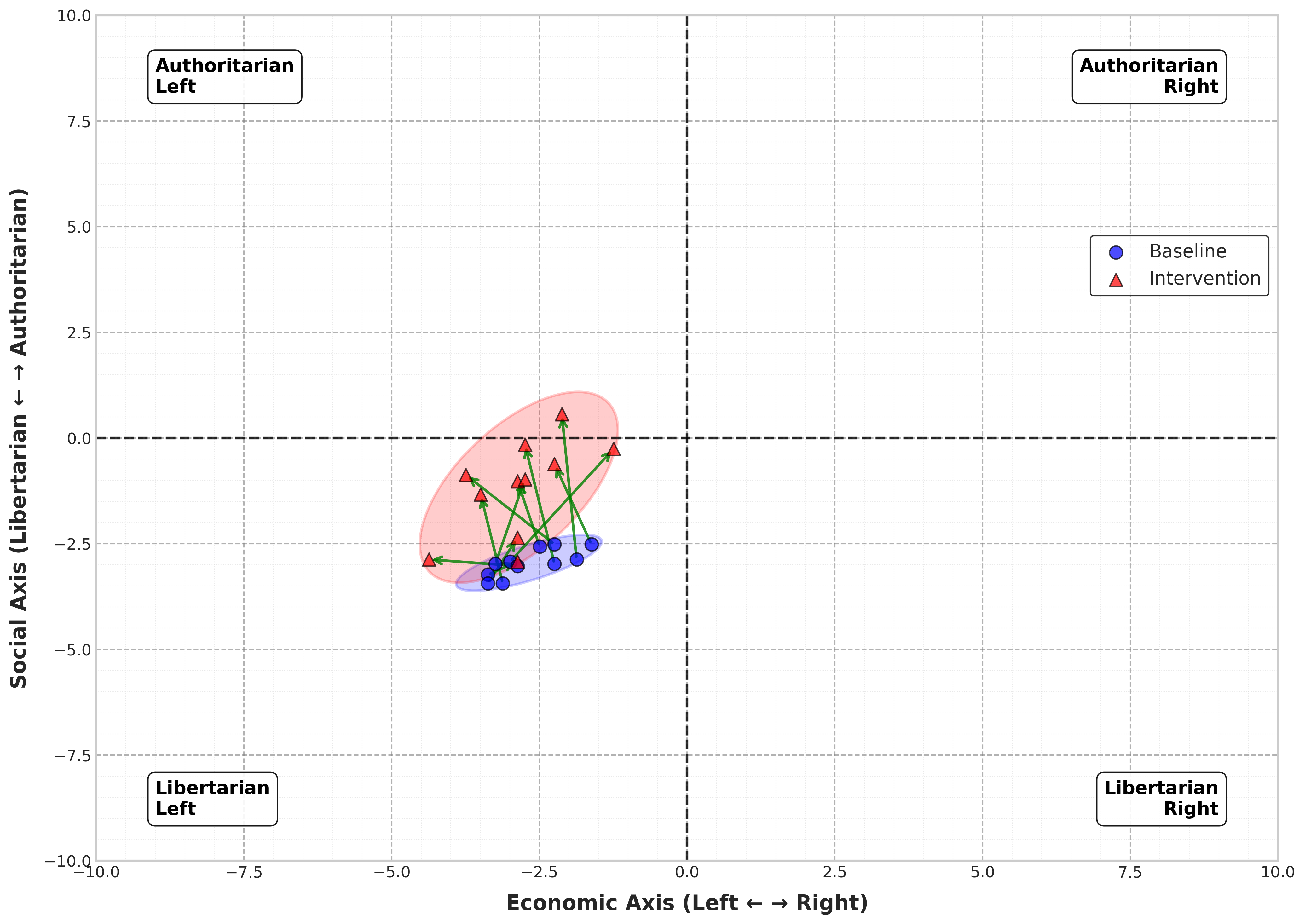}
    \caption{\texttt{Intervention Strength of 20}}
    \label{fig:inter_20_2}
  \end{subfigure}

    \vspace{0.5em}
    
  \begin{subfigure}[t]{0.48\textwidth}
    \centering
    \includegraphics[width=\linewidth]{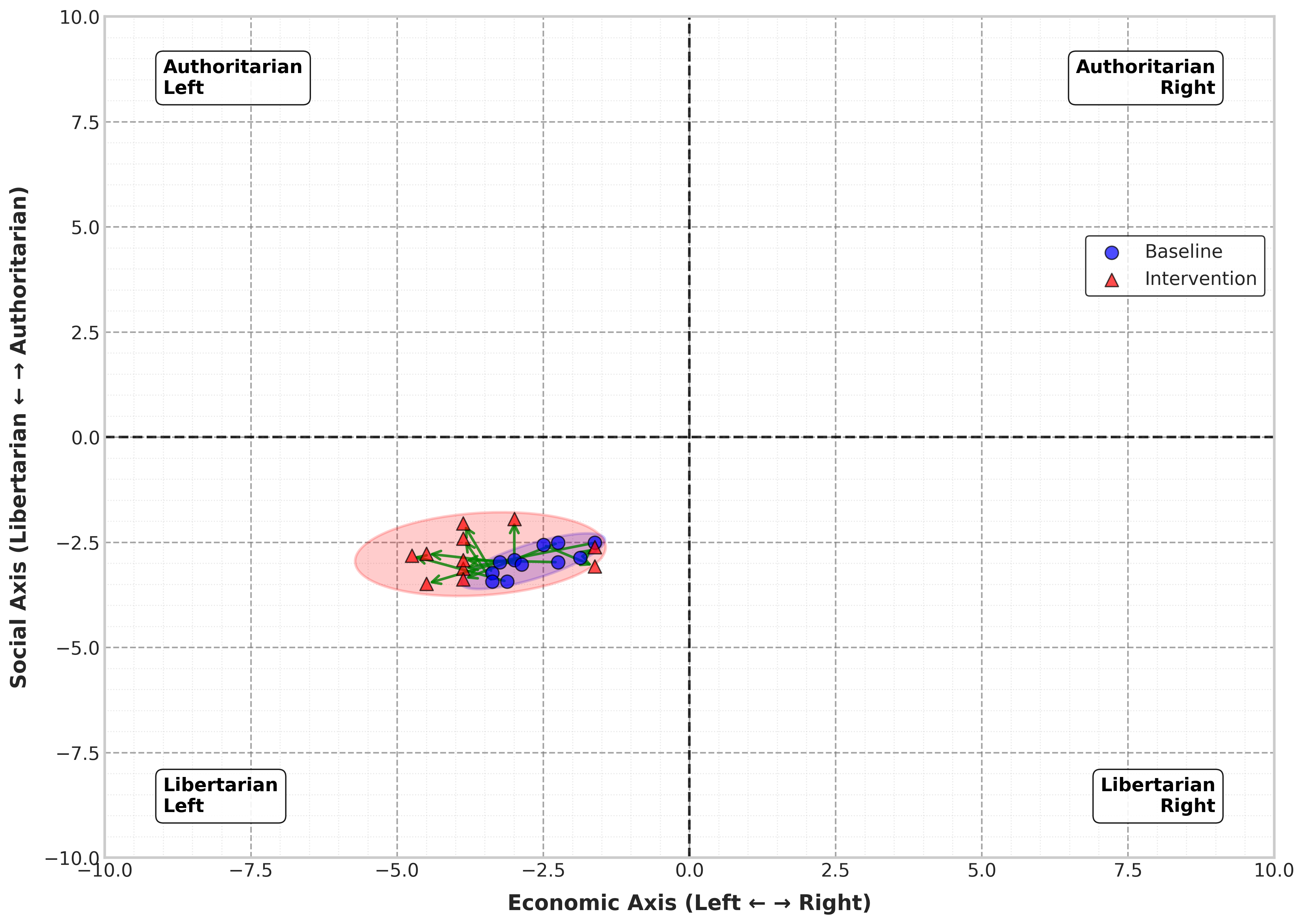}
    \caption{\texttt{Intervention Strength of 25}}
    \label{fig:inter_25_1}
  \end{subfigure}
  \hspace{0.02\textwidth}
  \begin{subfigure}[t]{0.48\textwidth}
    \centering
    \includegraphics[width=\linewidth]{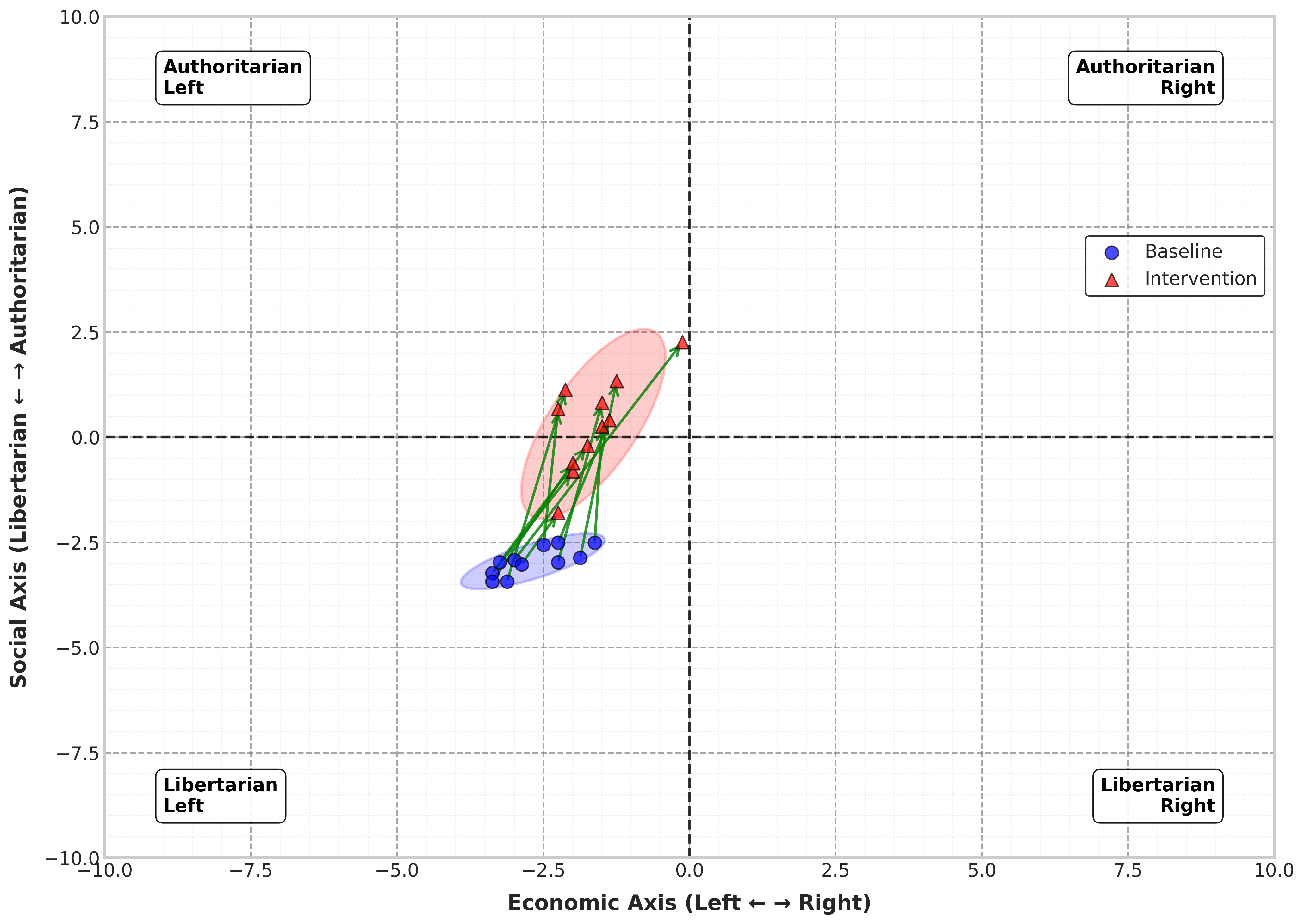}
    \caption{\texttt{Intervention Strength of 25}}
    \label{fig:inter_25_2}
  \end{subfigure}

    \vspace{0.5em}

  \begin{subfigure}[t]{0.48\textwidth}
    \centering
    \includegraphics[width=\linewidth]{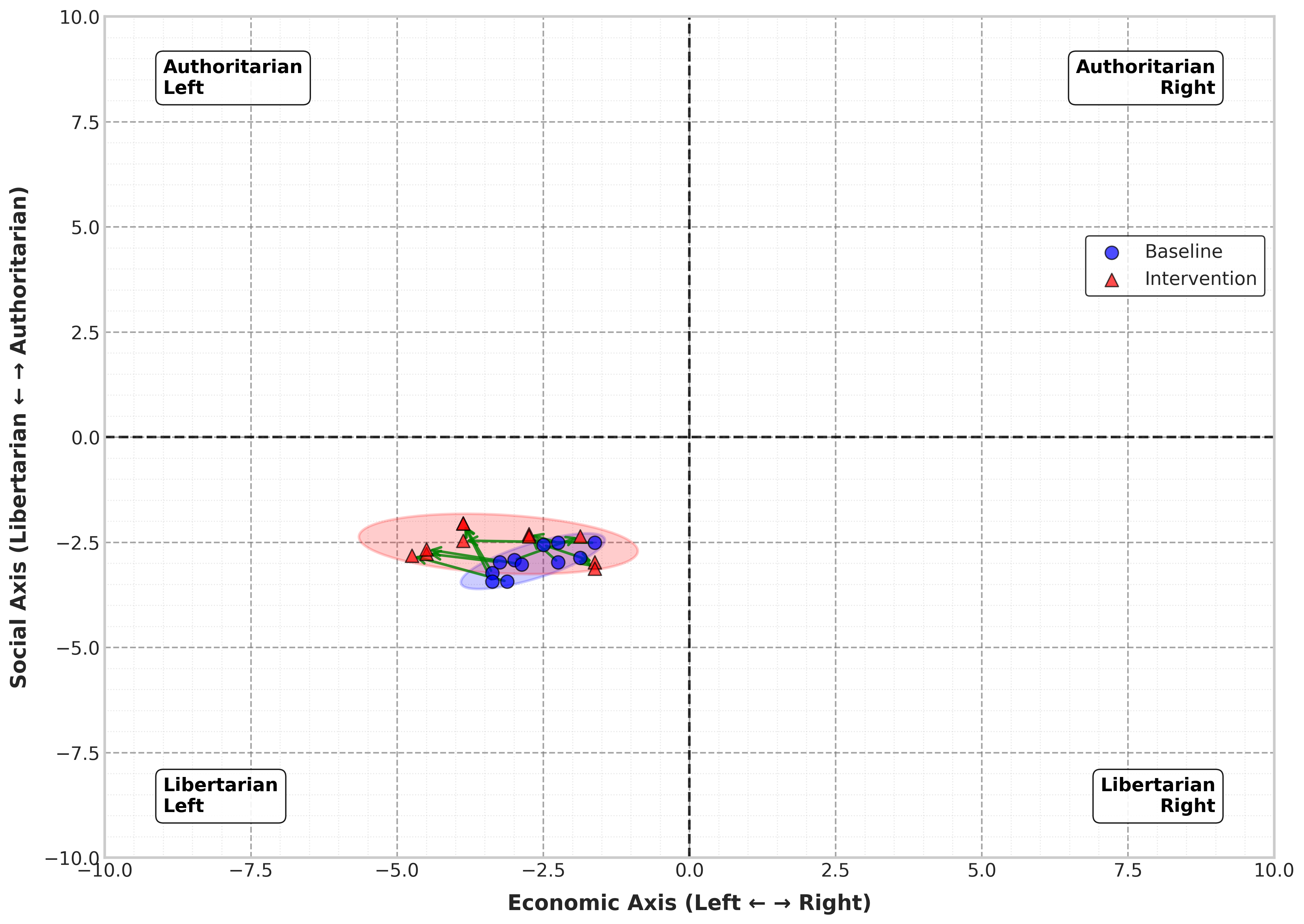}
    \caption{\texttt{Intervention Strength of 30}}
    \label{fig:inter_30_1}
  \end{subfigure}
  \hspace{0.02\textwidth}
  \begin{subfigure}[t]{0.48\textwidth}
    \centering
    \includegraphics[width=\linewidth]{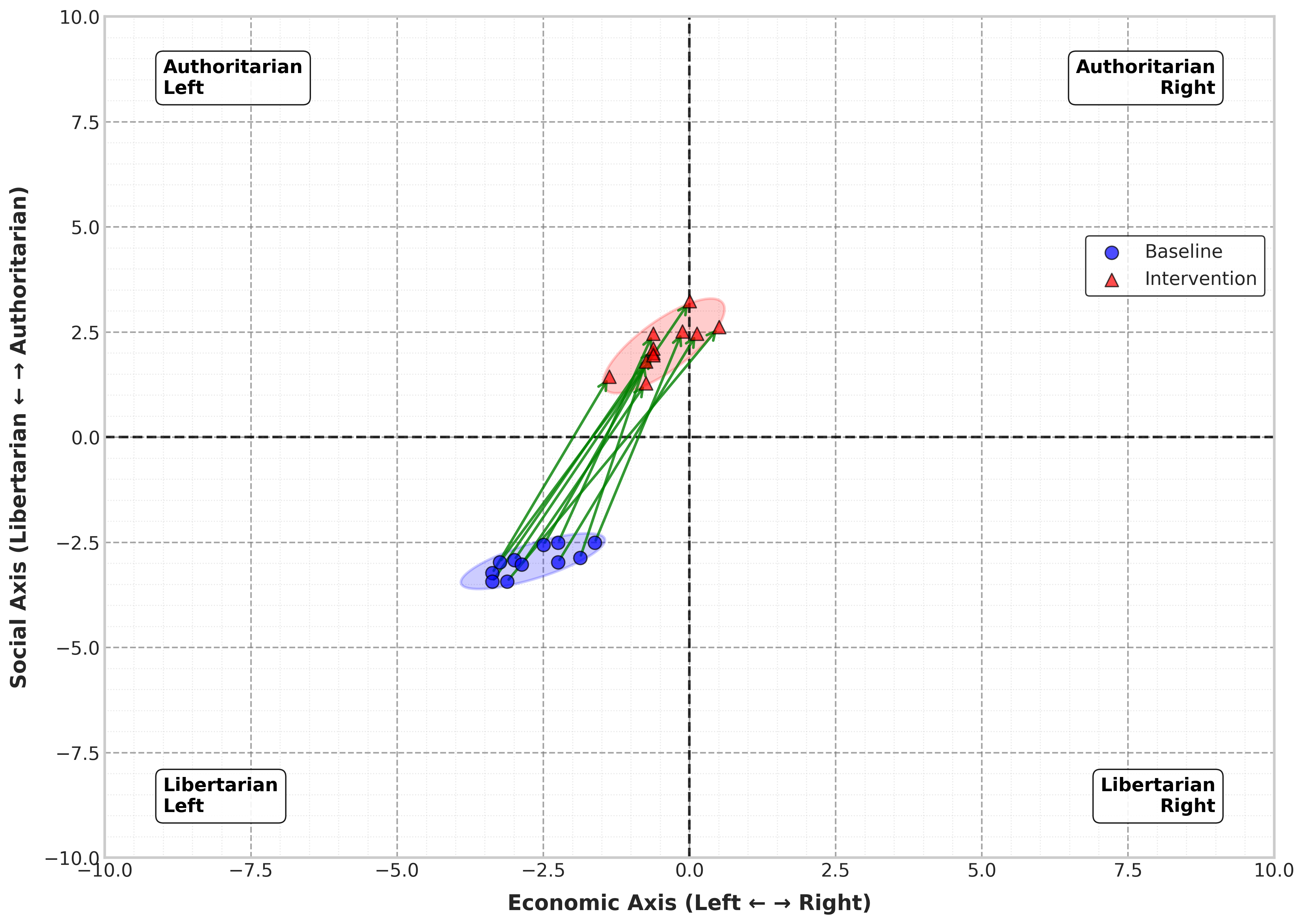}
    \caption{\texttt{Intervention Strength of 30}}
    \label{fig:inter_30_2}
  \end{subfigure}

  \caption{Political compass intervention results on 256 heads for two different intervention strengths for both directions on the PCT test in \textbf{English}. The plots on the right demonstrate steering towards politically right responses, and the plots on the left--towards politically left responses.}
  \label{fig:inter_results}
\end{figure*}

\begin{figure*}[h]
  \centering
  \begin{subfigure}[t]{0.48\textwidth}
    \centering
    \includegraphics[width=\linewidth]{assets/intervention_512_20_1.png}
    \caption{\texttt{Intervention Strength of 20}}
    \label{fig:inter_512_20_1}
  \end{subfigure}
  \hspace{0.02\textwidth}
  \begin{subfigure}[t]{0.48\textwidth}
    \centering
    \includegraphics[width=\linewidth]{assets/intervention_512_20_2.png}
    \caption{\texttt{Intervention Strength of 20}}
    \label{fig:inter_512_20_2}
  \end{subfigure}

  \vspace{0.5em}

  \begin{subfigure}[t]{0.48\textwidth}
    \centering
    \includegraphics[width=\linewidth]{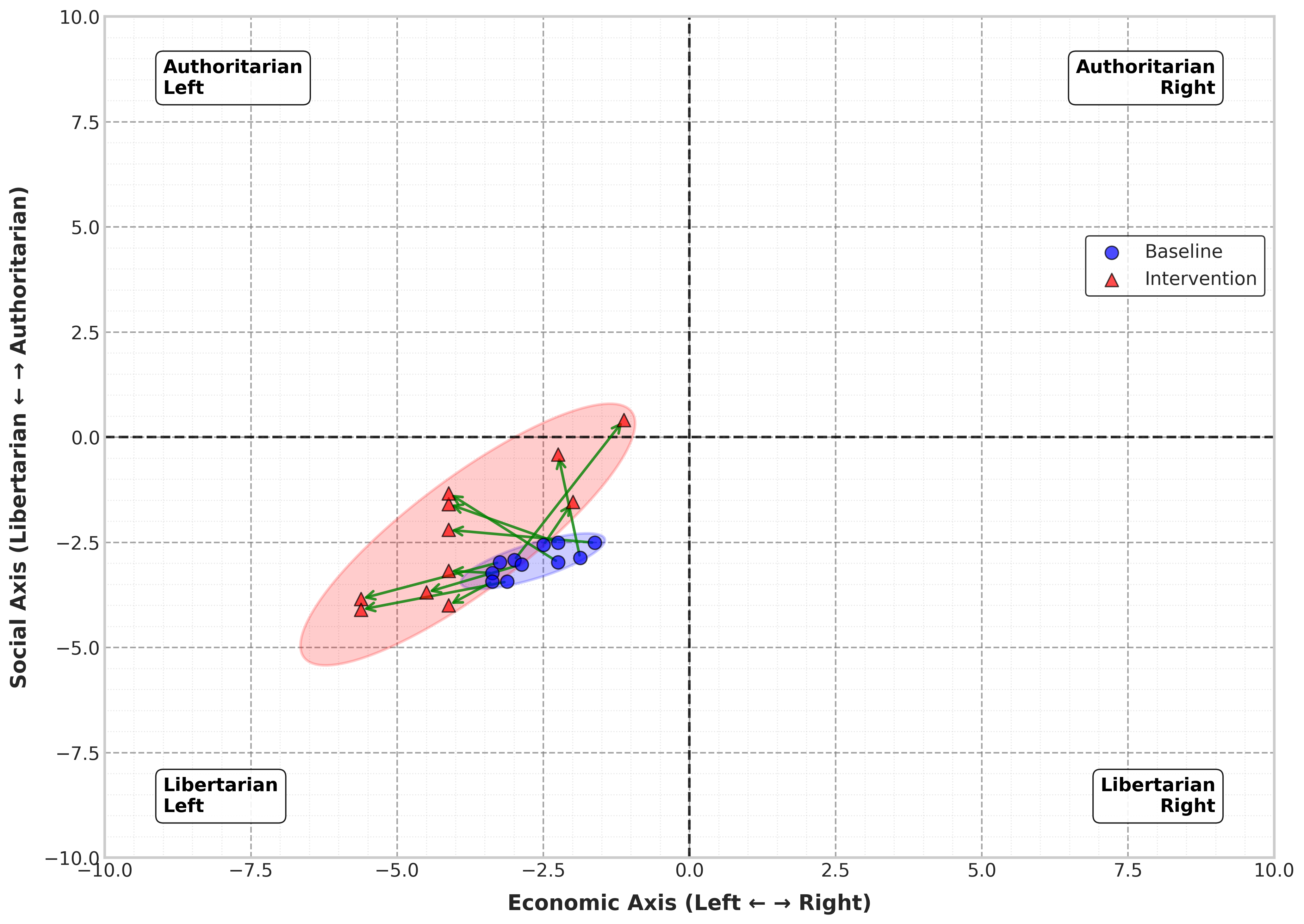}
    \caption{\texttt{Intervention Strength of 25}}
    \label{fig:inter_512_25_1}
  \end{subfigure}
  \hspace{0.02\textwidth}
  \begin{subfigure}[t]{0.48\textwidth}
    \centering
    \includegraphics[width=\linewidth]{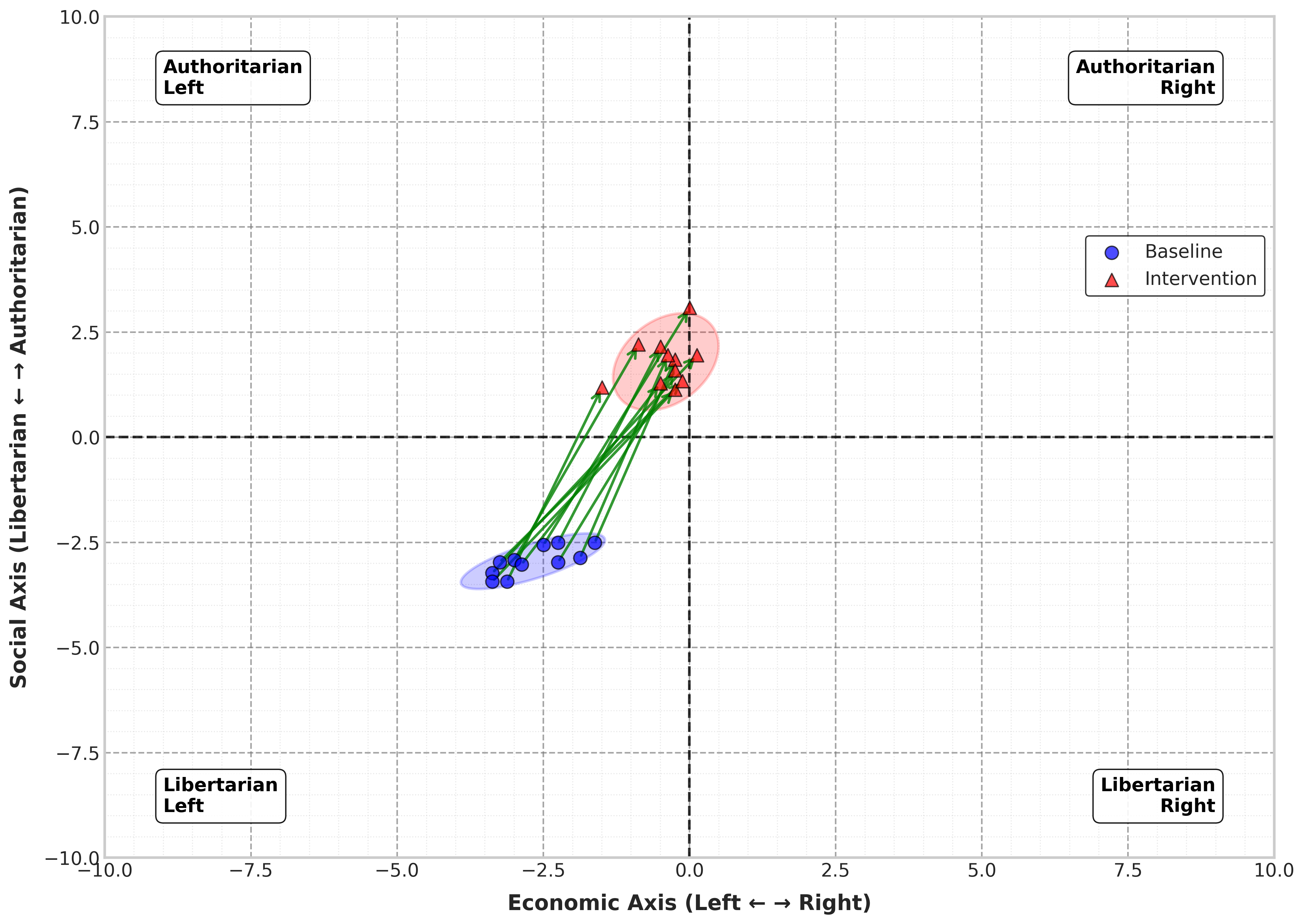}
    \caption{\texttt{Intervention Strength of 25}}
    \label{fig:inter_512_25_2}
  \end{subfigure}

  \vspace{0.5em}

  \begin{subfigure}[t]{0.48\textwidth}
    \centering
    \includegraphics[width=\linewidth]{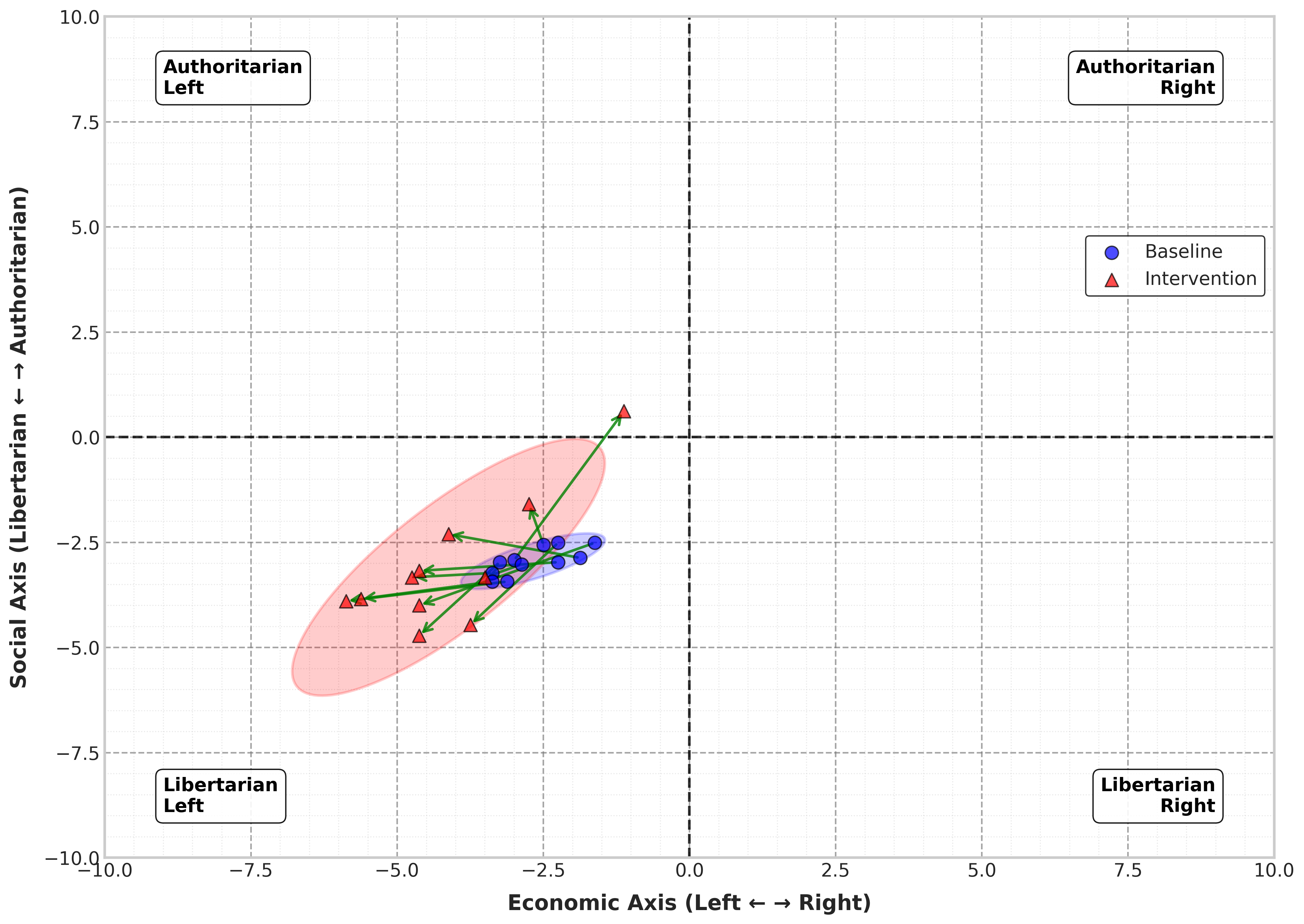}
    \caption{\texttt{Intervention Strength of 30}}
    \label{fig:inter_512_30_1}
  \end{subfigure}
  \hspace{0.02\textwidth}
  \begin{subfigure}[t]{0.48\textwidth}
    \centering
    \includegraphics[width=\linewidth]{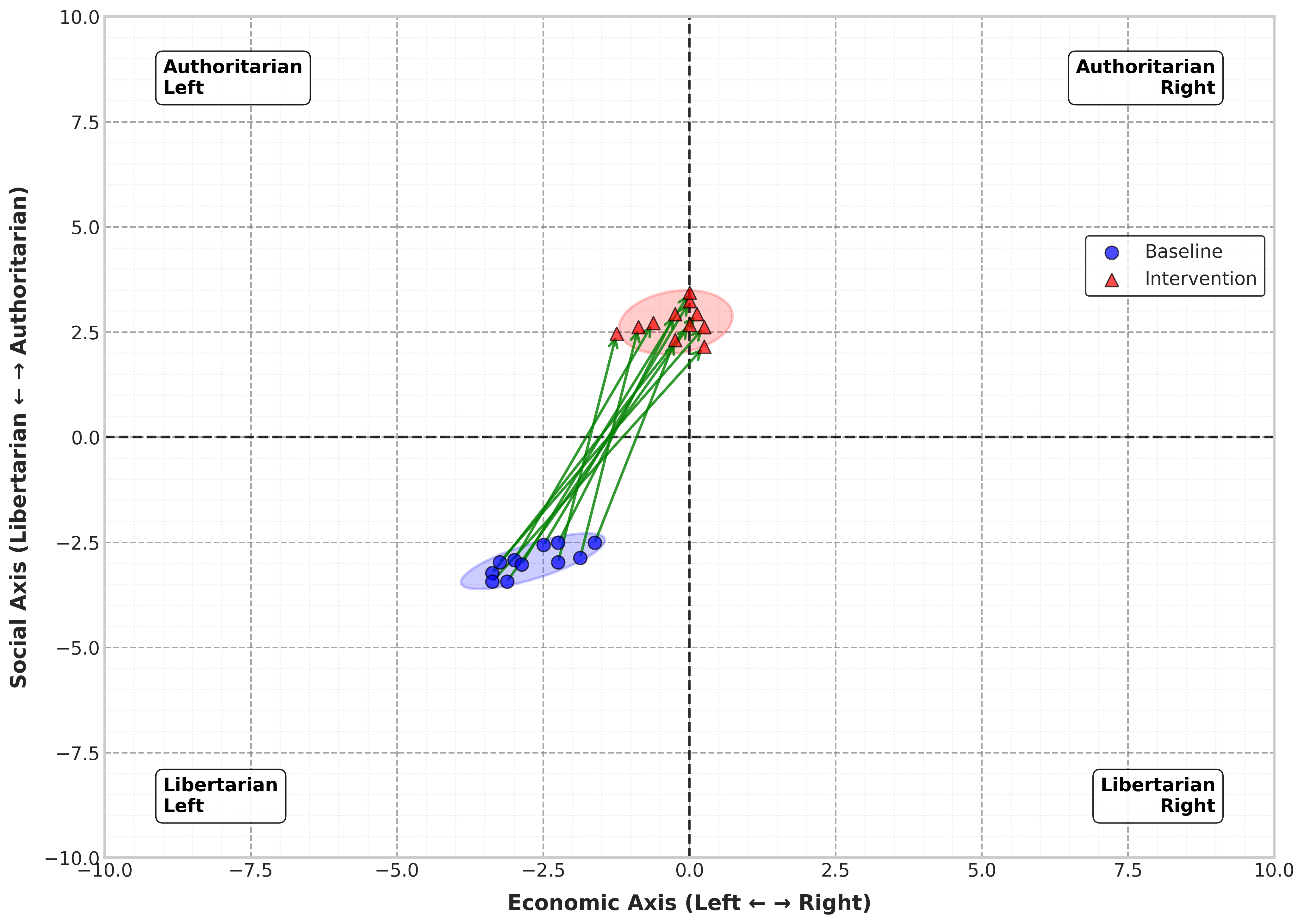}
    \caption{\texttt{Intervention Strength of 30}}
    \label{fig:inter_512_30_2}
  \end{subfigure}

  \caption{Political compass intervention results on 512 heads for two different intervention strengths for both directions on the PCT test in \textbf{English}. The plots on the right demonstrate steering towards politically right responses, and the plots on the left--towards politically left responses.}
  \label{fig:inter_results_2}
\end{figure*}

\begin{figure*}[h]
  \centering
  \begin{subfigure}[t]{0.48\textwidth}
    \centering
    \includegraphics[width=\linewidth]{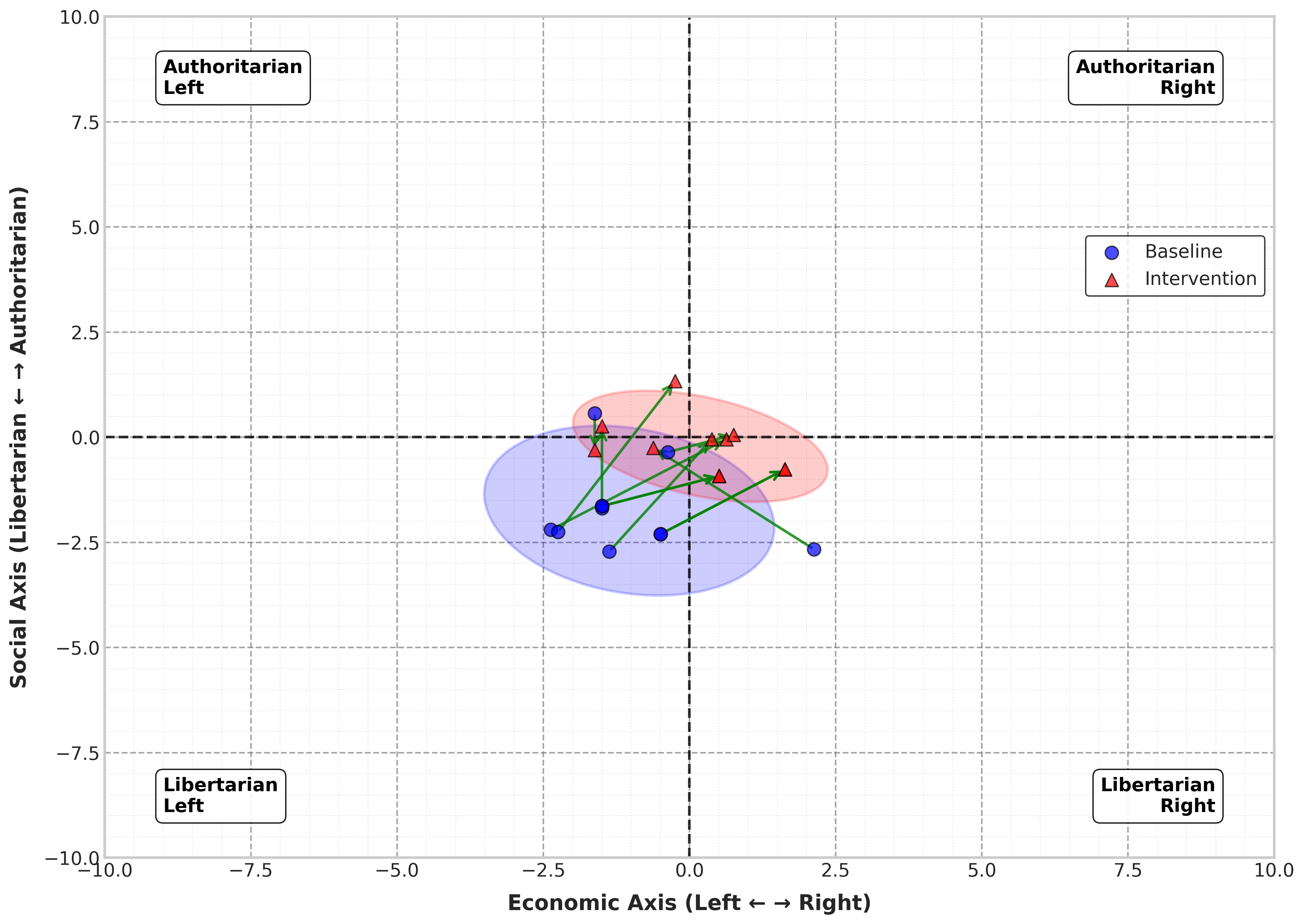}
    \caption{\texttt{Intervention Strength of 20}}
    \label{fig:inter_256_20_0_ro}
  \end{subfigure}
  \hspace{0.02\textwidth}
  \begin{subfigure}[t]{0.48\textwidth}
    \centering
    \includegraphics[width=\linewidth]{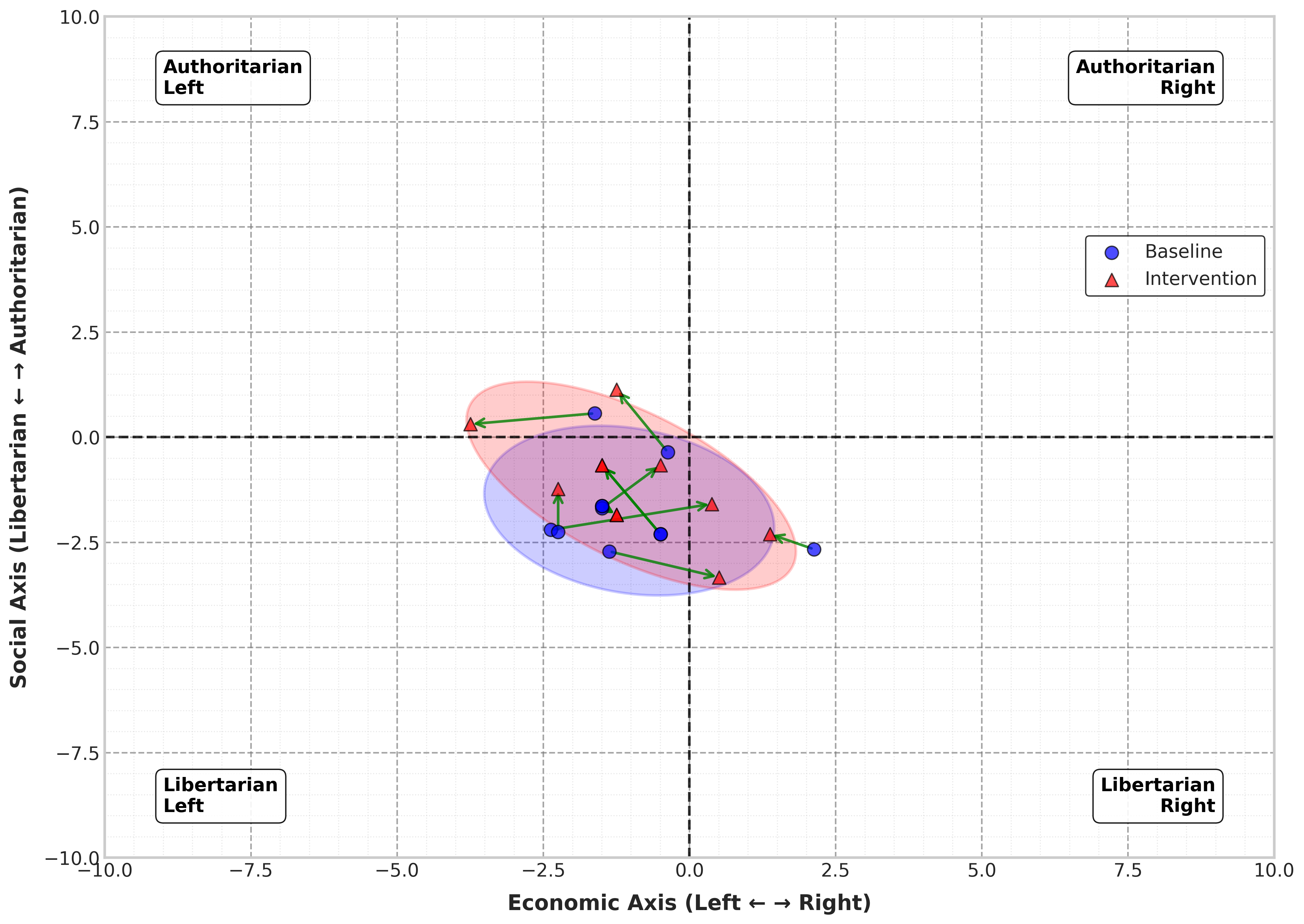}
    \caption{\texttt{Intervention Strength of 20}}
    \label{fig:inter_256_20_1_ro}
  \end{subfigure}

  \vspace{0.5em}

  \begin{subfigure}[t]{0.48\textwidth}
    \centering
    \includegraphics[width=\linewidth]{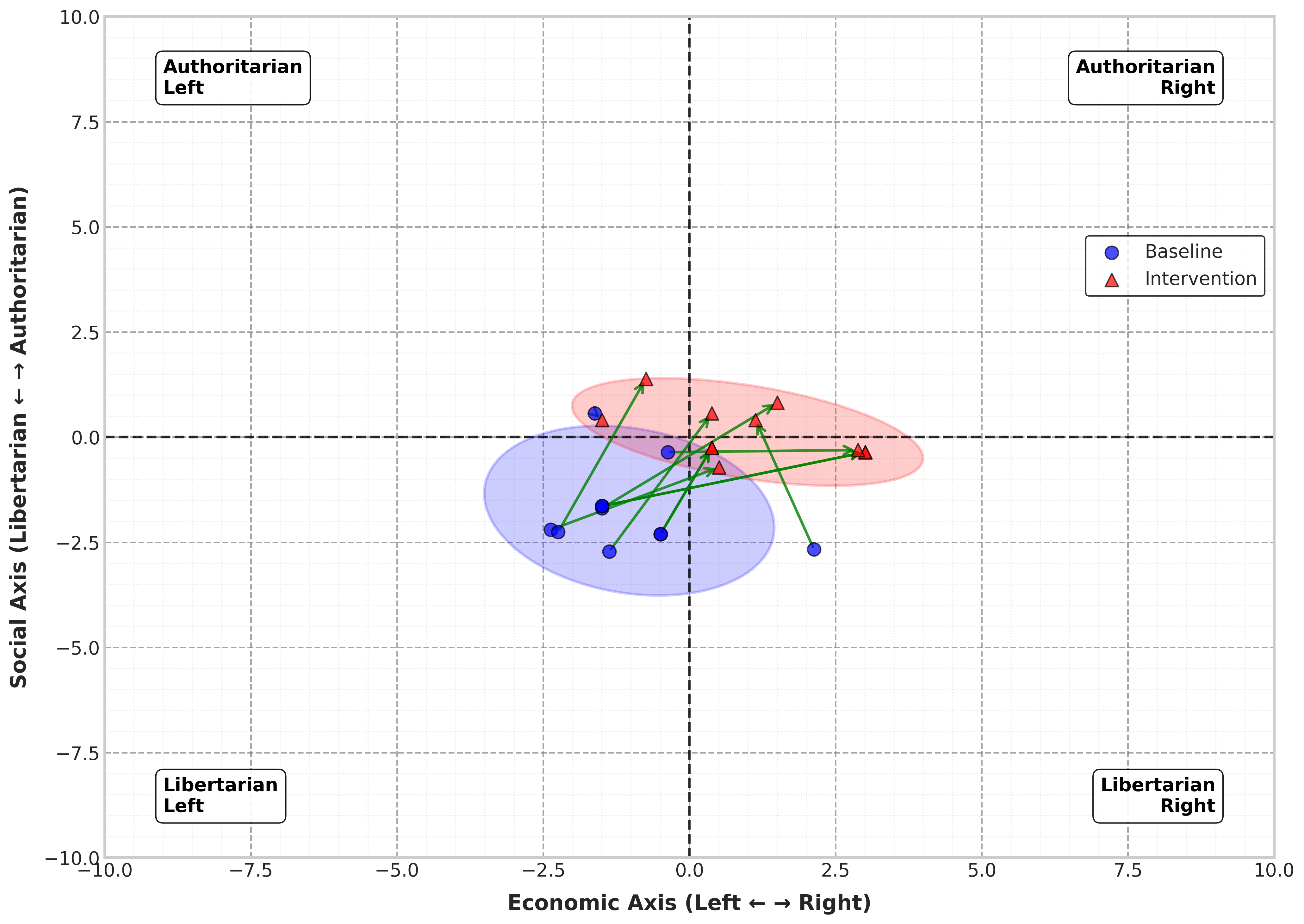}
    \caption{\texttt{Intervention Strength of 25}}
    \label{fig:inter_256_25_0_ro}
  \end{subfigure}
  \hspace{0.02\textwidth}
  \begin{subfigure}[t]{0.48\textwidth}
    \centering
    \includegraphics[width=\linewidth]{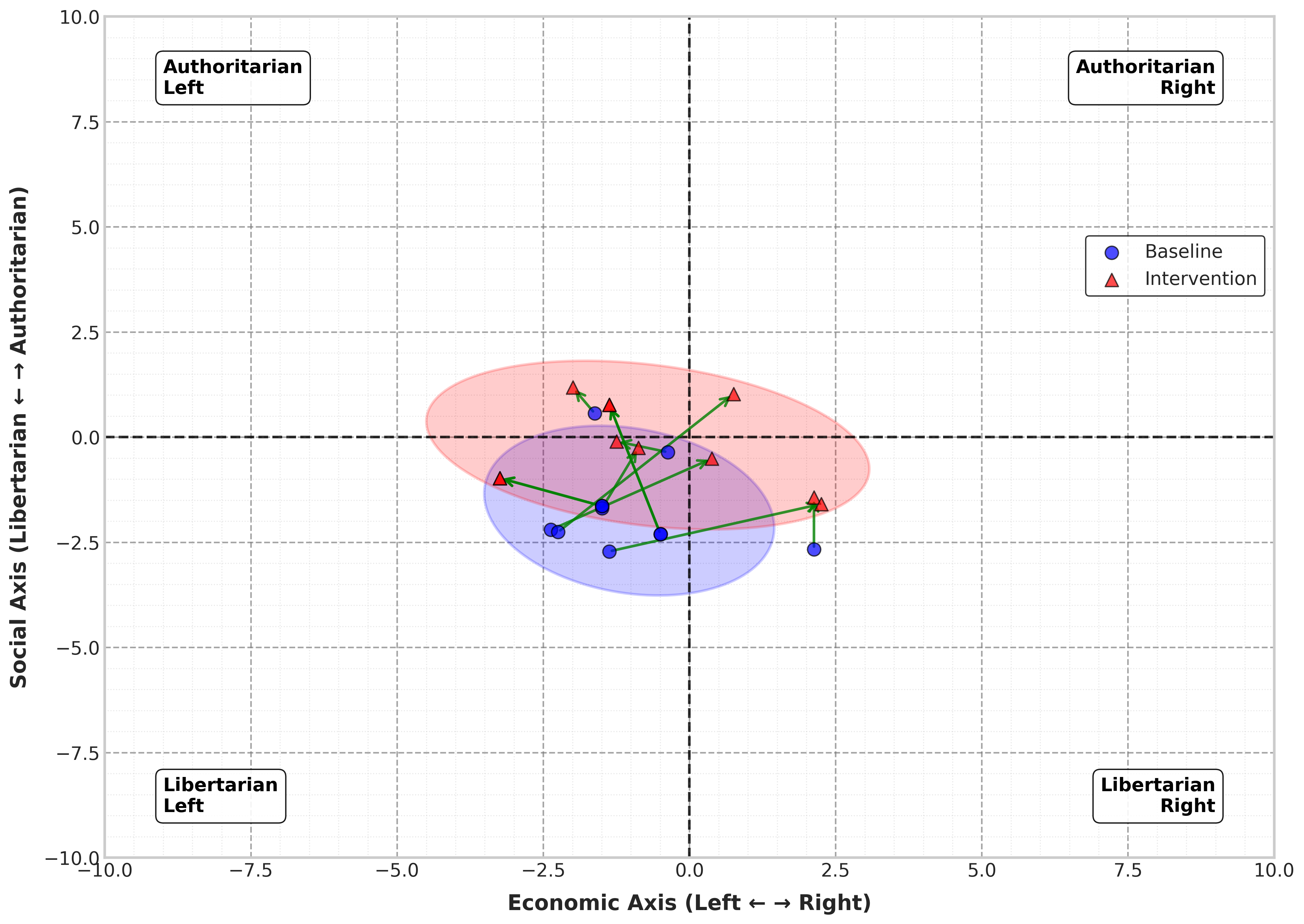}
    \caption{\texttt{Intervention Strength of 25}}
    \label{fig:inter_256_25_1_ro}
  \end{subfigure}

  \vspace{0.5em}

  \begin{subfigure}[t]{0.48\textwidth}
    \centering
    \includegraphics[width=\linewidth]{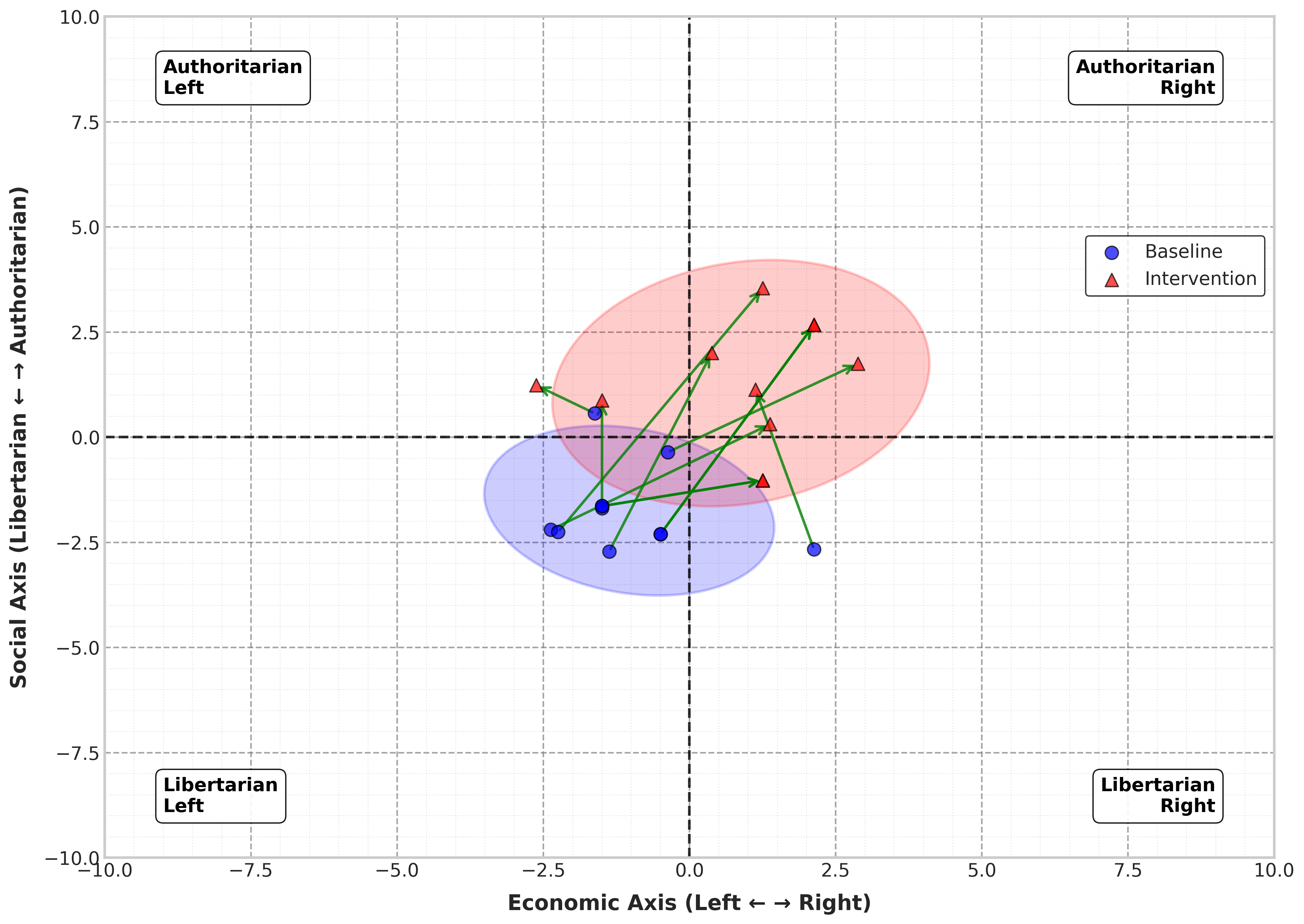}
    \caption{\texttt{Intervention Strength of 30}}
    \label{fig:inter_256_30_0_ro}
  \end{subfigure}
  \hspace{0.02\textwidth}
  \begin{subfigure}[t]{0.48\textwidth}
    \centering
    \includegraphics[width=\linewidth]{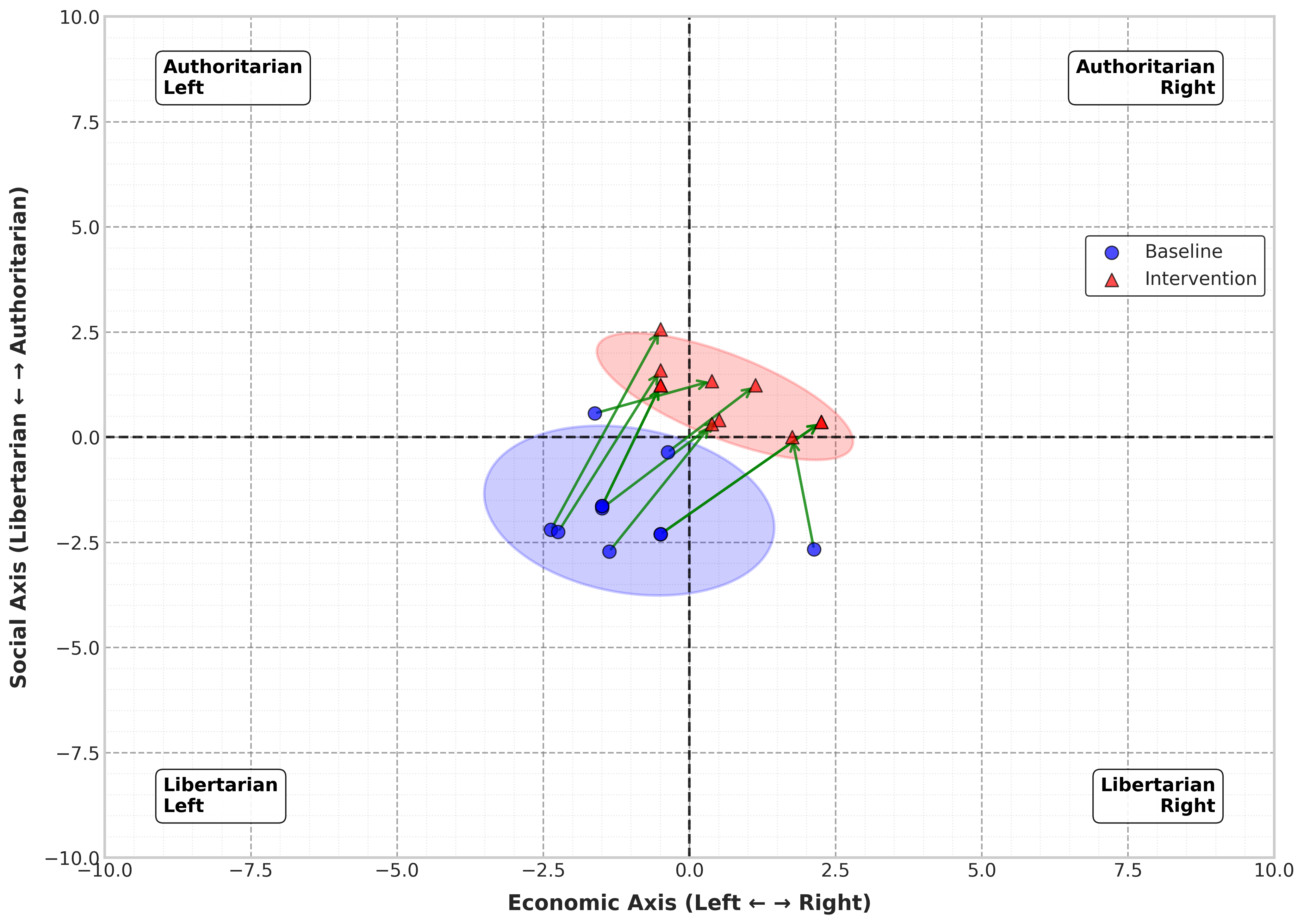}
    \caption{\texttt{Intervention Strength of 30}}
    \label{fig:inter_256_30_1_ro}
  \end{subfigure}

  \caption{Political compass intervention results on 256 heads for two different intervention strengths for both directions on the PCT test in \textbf{Romanian}. The plots on the right demonstrate steering towards politically right responses, and the plots on the left--towards politically left responses.}
  \label{fig:inter_results_ro}
\end{figure*}

\begin{figure*}[h]
  \centering
  \begin{subfigure}[t]{0.48\textwidth}
    \centering
    \includegraphics[width=\linewidth]{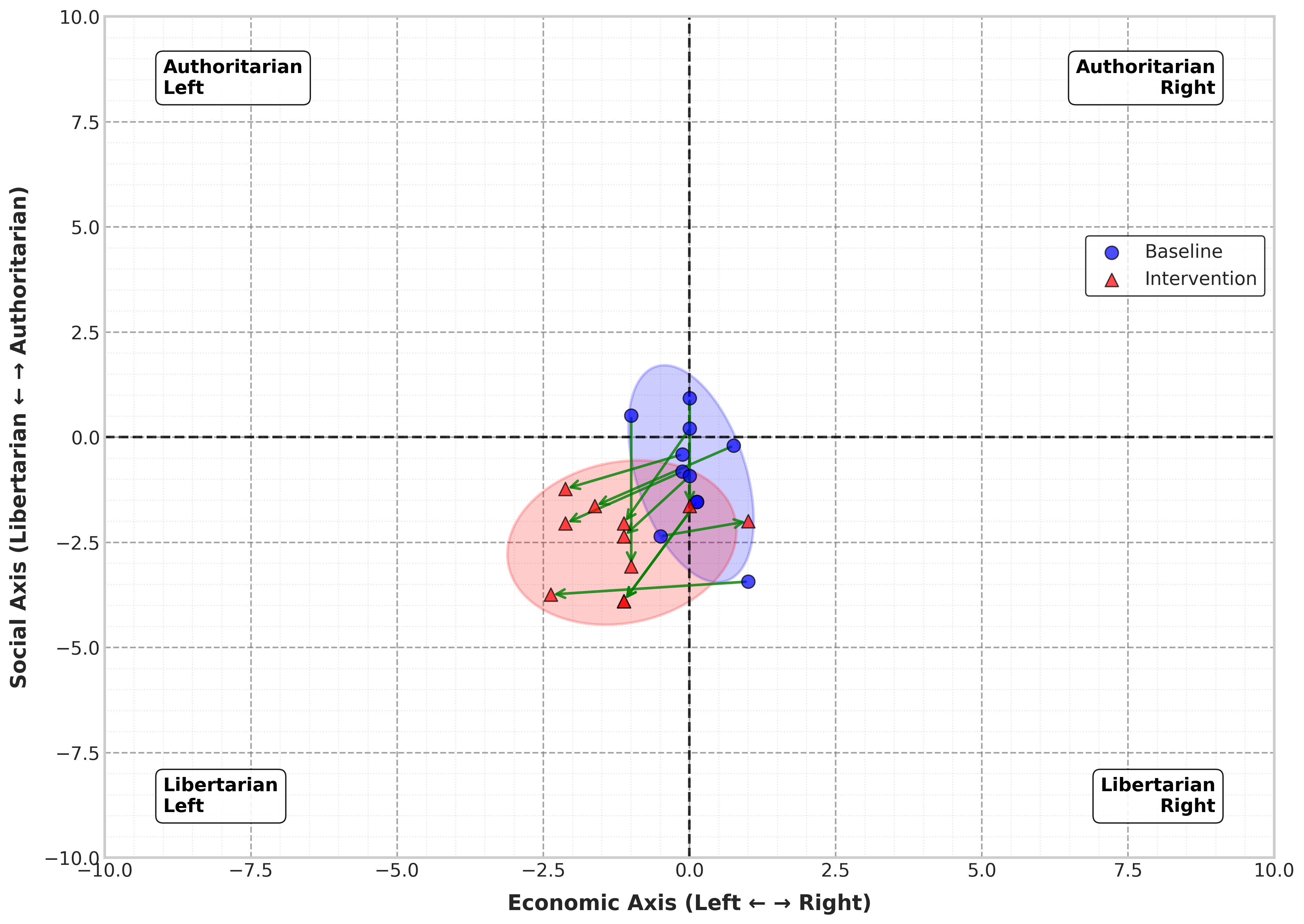}
    \caption{\texttt{Intervention Strength of 20}}
    \label{fig:inter_256_20_0_tr}
  \end{subfigure}
  \hspace{0.02\textwidth}
  \begin{subfigure}[t]{0.48\textwidth}
    \centering
    \includegraphics[width=\linewidth]{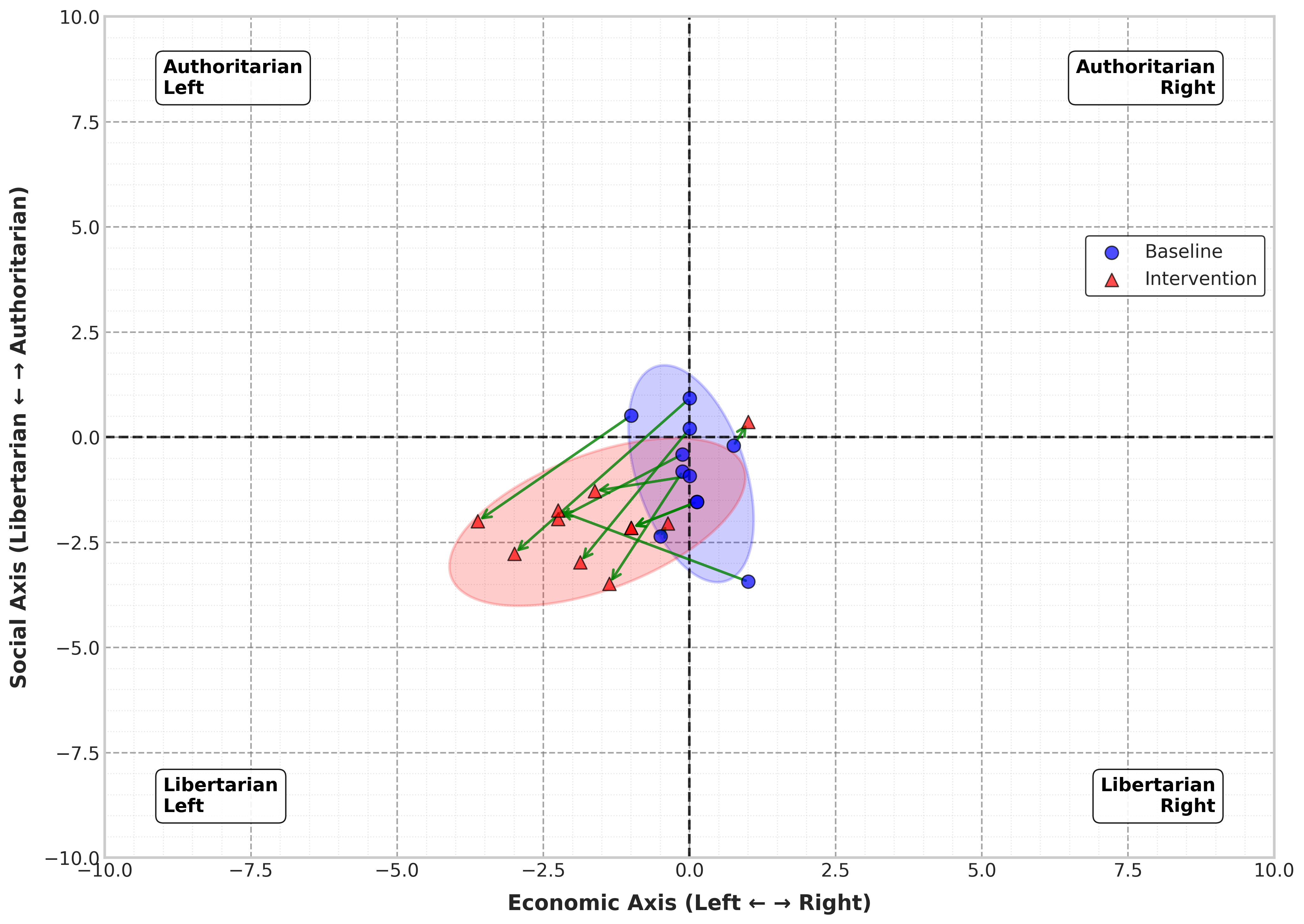}
    \caption{\texttt{Intervention Strength of 20}}
    \label{fig:inter_256_20_1_tr}
  \end{subfigure}

  \vspace{0.5em}

  \begin{subfigure}[t]{0.48\textwidth}
    \centering
    \includegraphics[width=\linewidth]{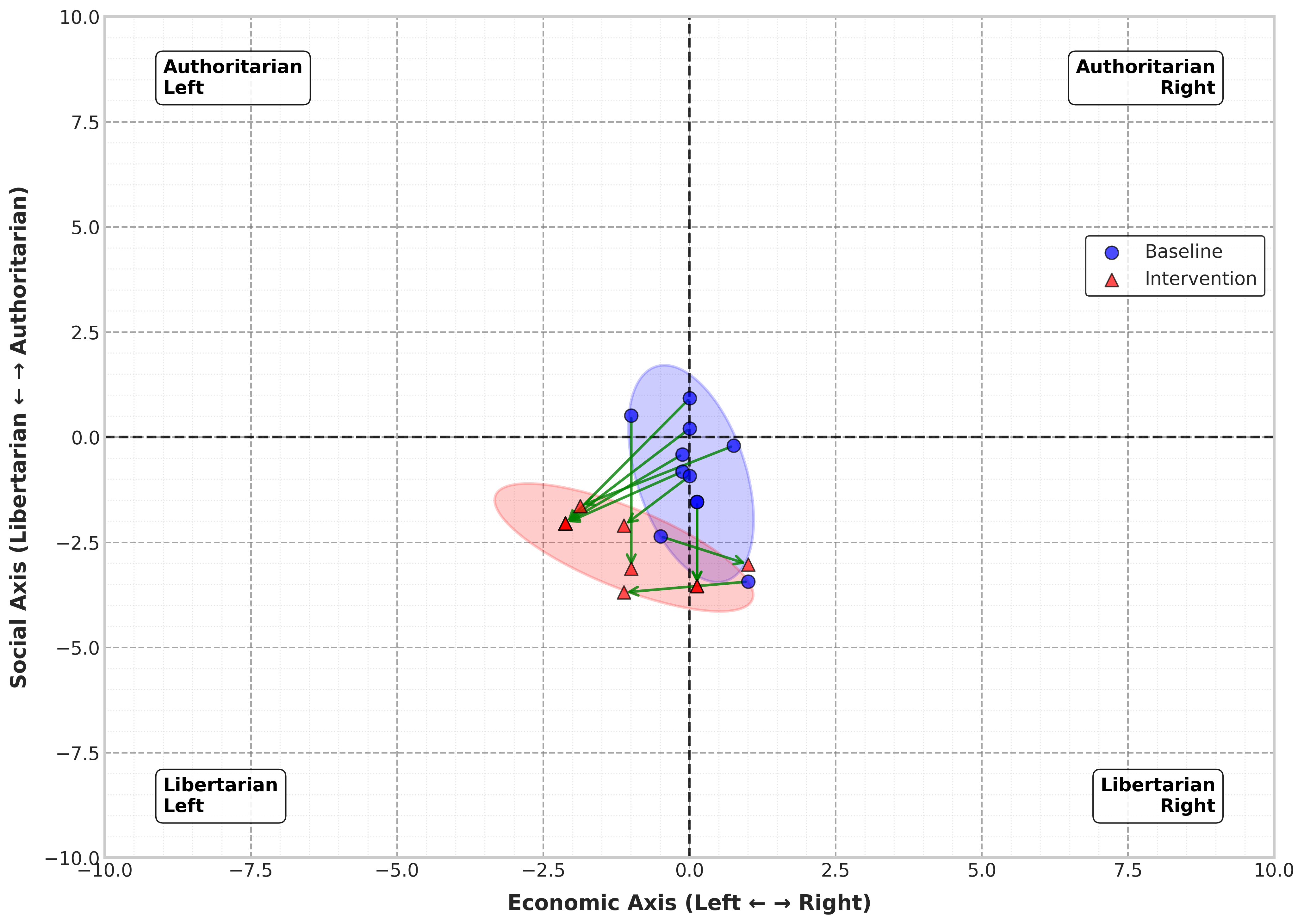}
    \caption{\texttt{Intervention Strength of 25}}
    \label{fig:inter_256_25_0_tr}
  \end{subfigure}
  \hspace{0.02\textwidth}
  \begin{subfigure}[t]{0.48\textwidth}
    \centering
    \includegraphics[width=\linewidth]{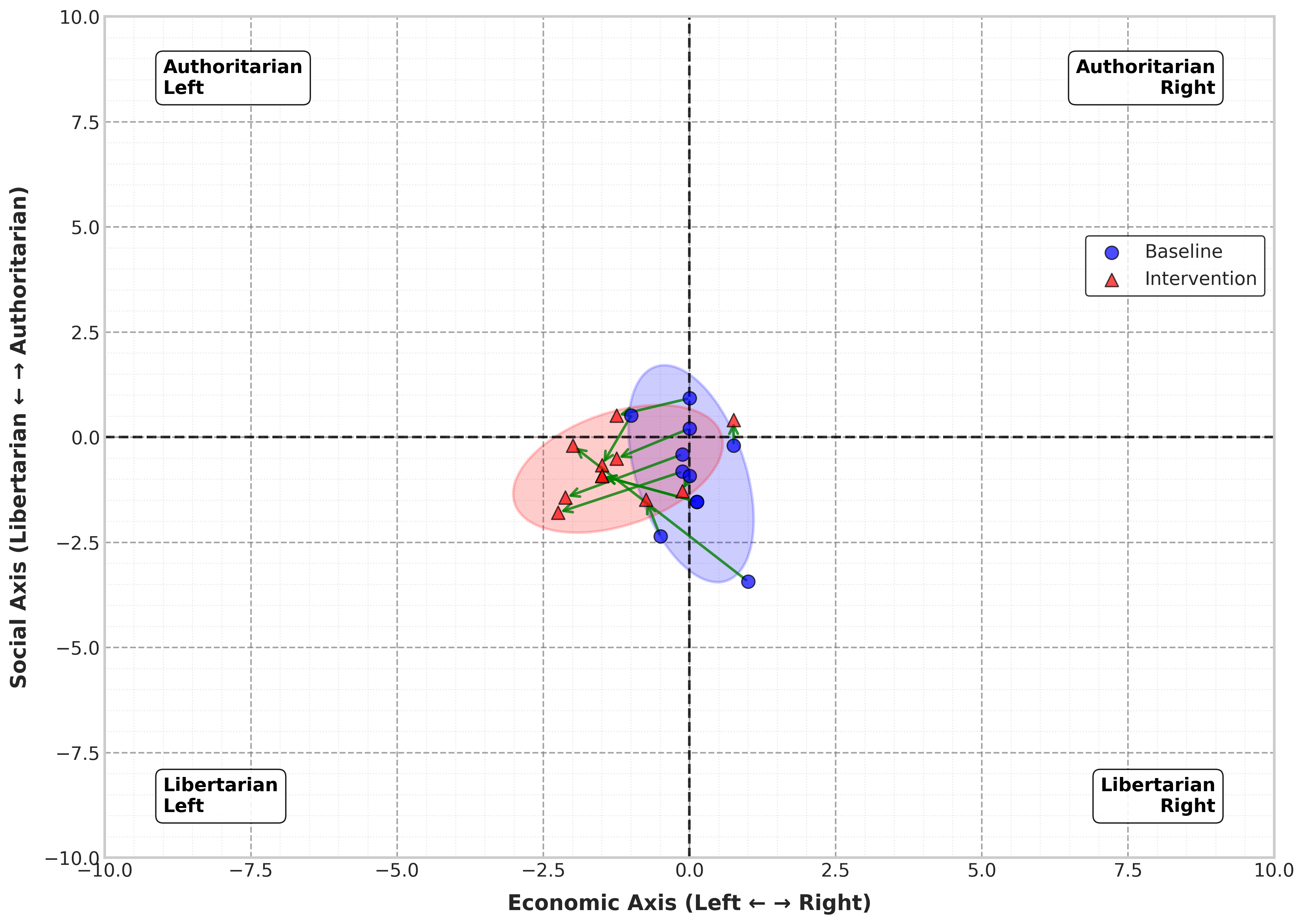}
    \caption{\texttt{Intervention Strength of 25}}
    \label{fig:inter_256_25_1_tr}
  \end{subfigure}

  \vspace{0.5em}

  \begin{subfigure}[t]{0.48\textwidth}
    \centering
    \includegraphics[width=\linewidth]{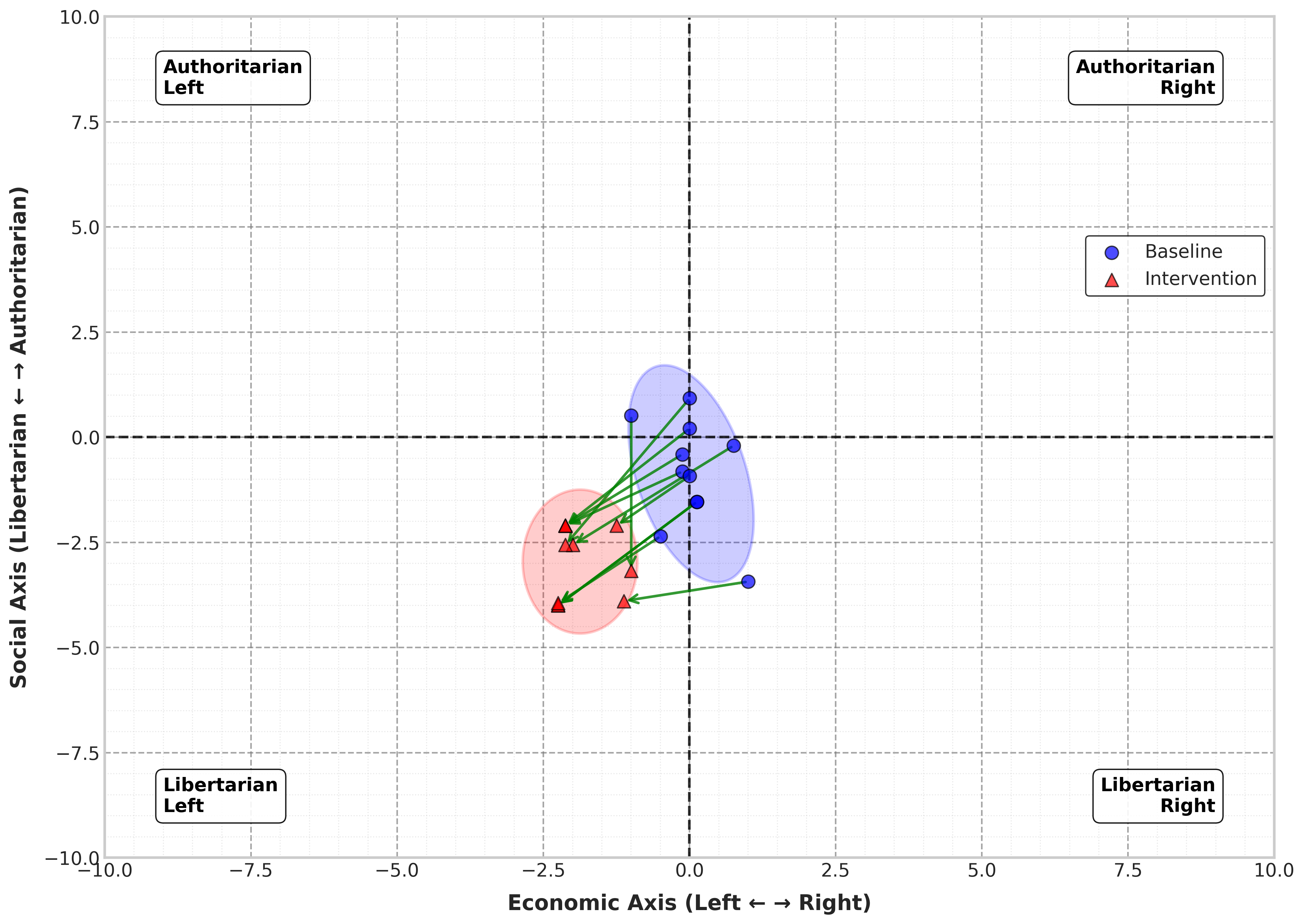}
    \caption{\texttt{Intervention Strength of 30}}
    \label{fig:inter_256_30_0_tr}
  \end{subfigure}
  \hspace{0.02\textwidth}
  \begin{subfigure}[t]{0.48\textwidth}
    \centering
    \includegraphics[width=\linewidth]{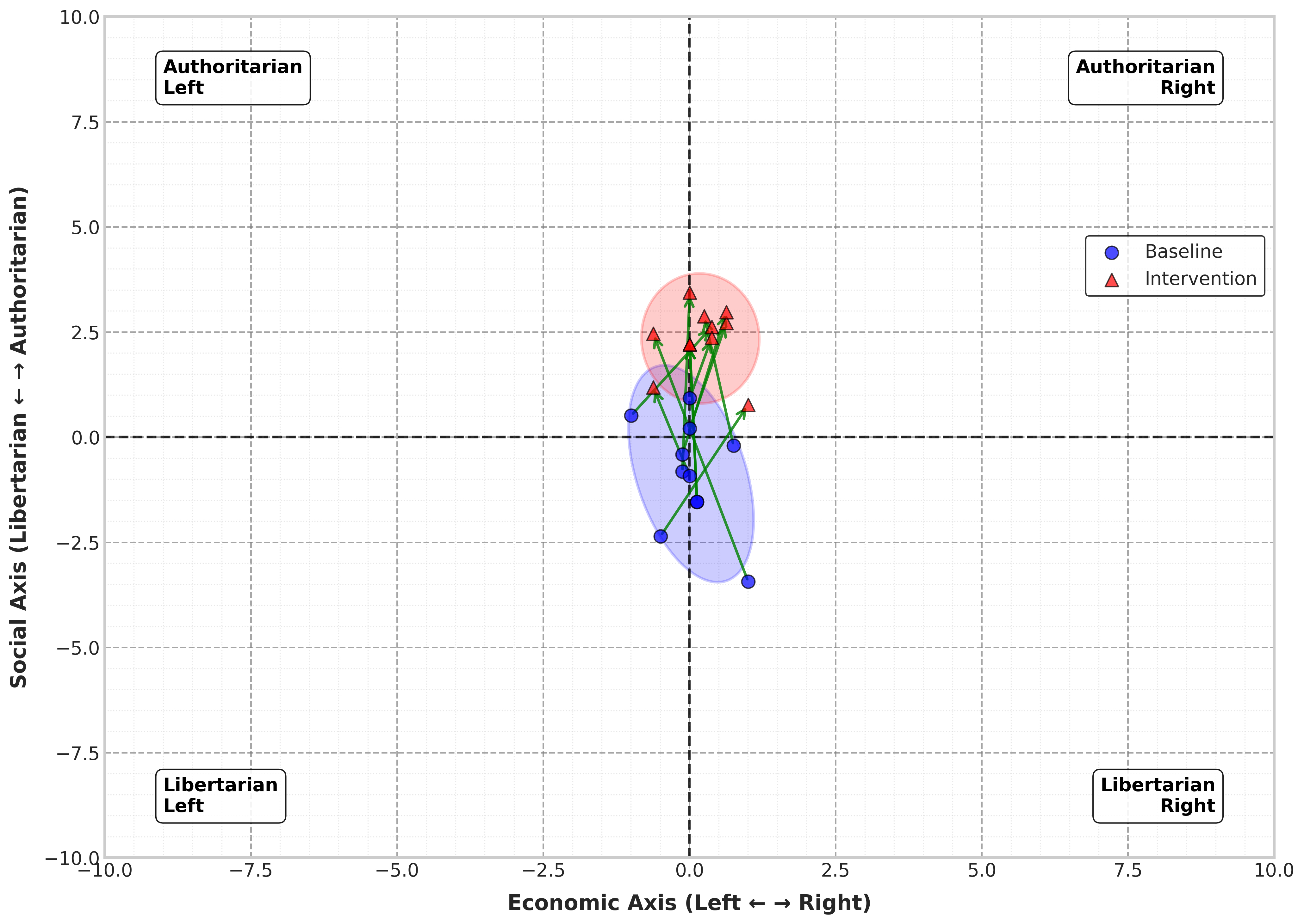}
    \caption{\texttt{Intervention Strength of 30}}
    \label{fig:inter_256_30_1_tr}
  \end{subfigure}

  \caption{Political compass intervention results on 256 heads for two different intervention strengths for both directions on the PCT test in \textbf{Turkish}. The plots on the right demonstrate steering towards politically right responses, and the plots on the left--towards politically left responses.}
  \label{fig:inter_results_tr}
\end{figure*}

\begin{figure*}[h]
  \centering
  \begin{subfigure}[t]{0.48\textwidth}
    \centering
    \includegraphics[width=\linewidth]{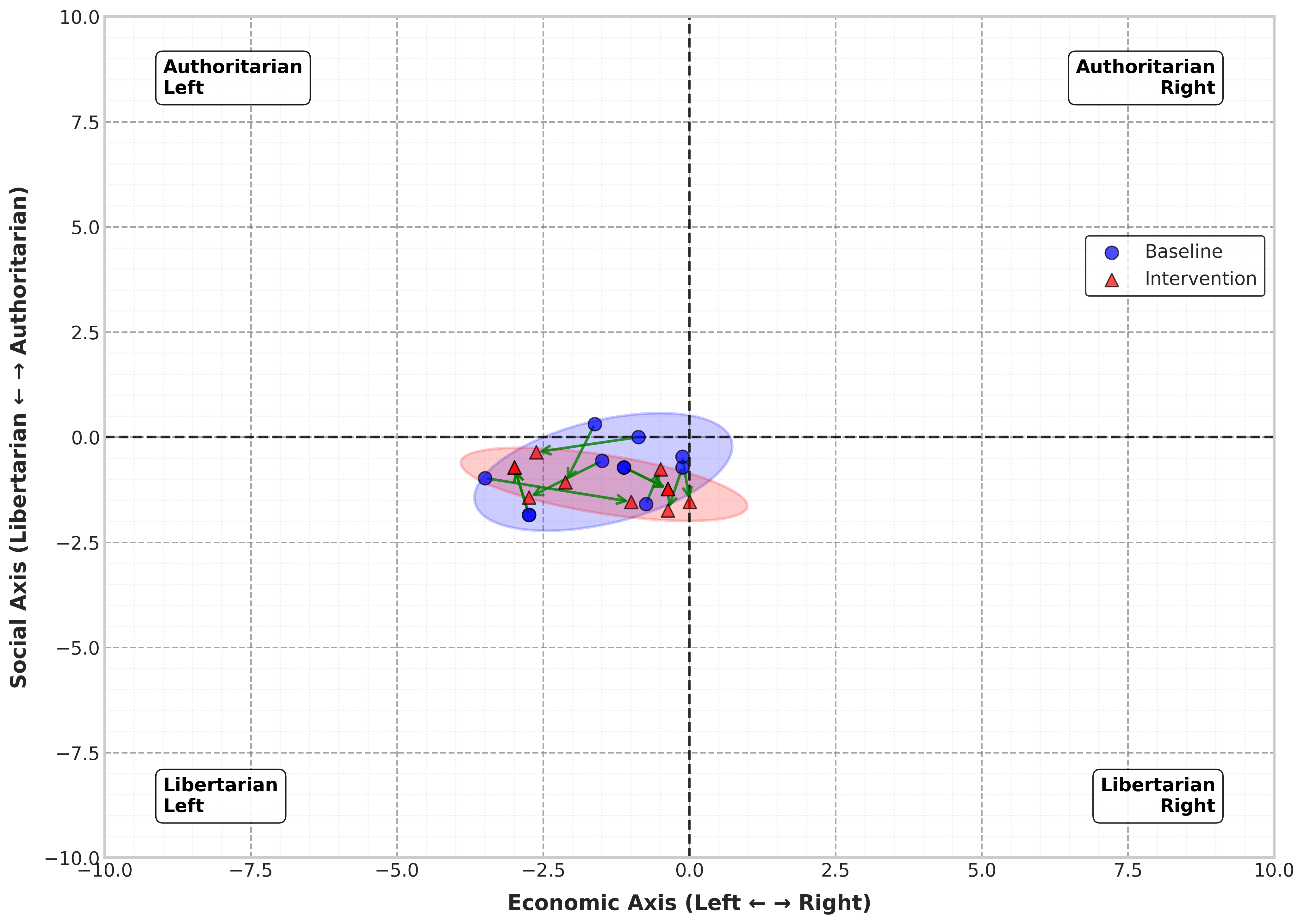}
    \caption{\texttt{Intervention Strength of 20}}
    \label{fig:inter_256_20_0_sl}
  \end{subfigure}
  \hspace{0.02\textwidth}
  \begin{subfigure}[t]{0.48\textwidth}
    \centering
    \includegraphics[width=\linewidth]{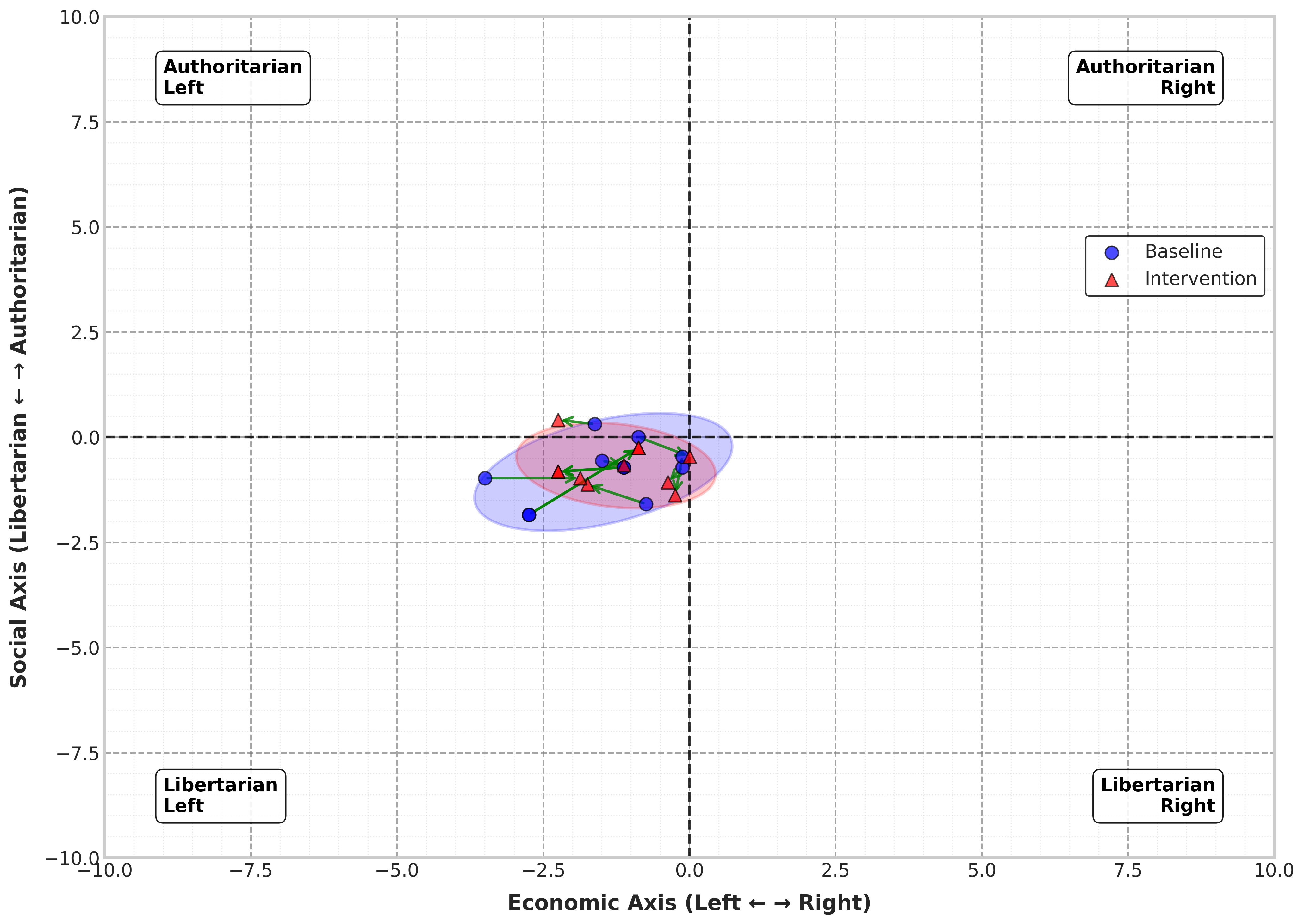}
    \caption{\texttt{Intervention Strength of 20}}
    \label{fig:inter_256_20_1_sl}
  \end{subfigure}

  \vspace{0.5em}

  \begin{subfigure}[t]{0.48\textwidth}
    \centering
    \includegraphics[width=\linewidth]{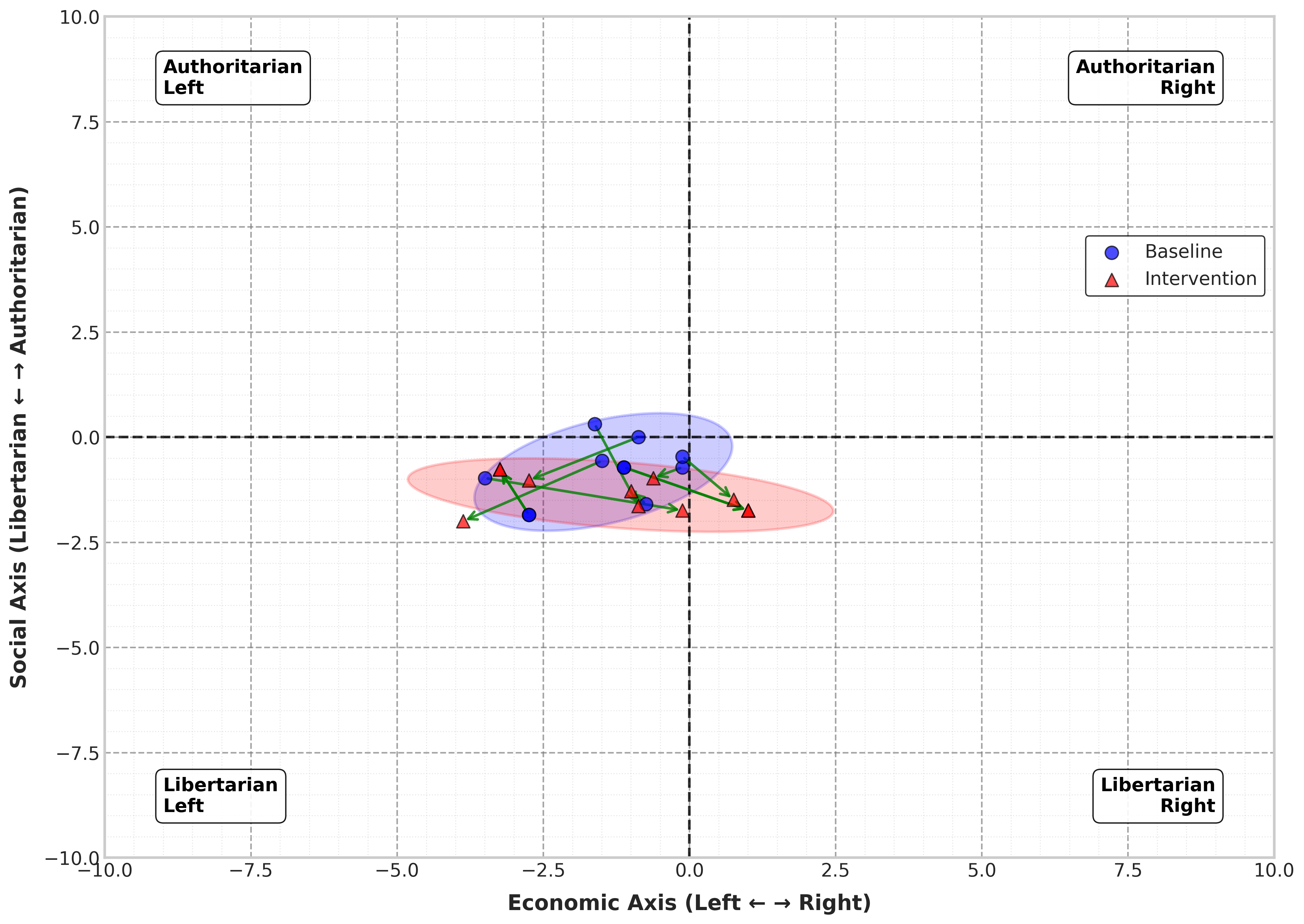}
    \caption{\texttt{Intervention Strength of 25}}
    \label{fig:inter_256_25_0_sl}
  \end{subfigure}
  \hspace{0.02\textwidth}
  \begin{subfigure}[t]{0.48\textwidth}
    \centering
    \includegraphics[width=\linewidth]{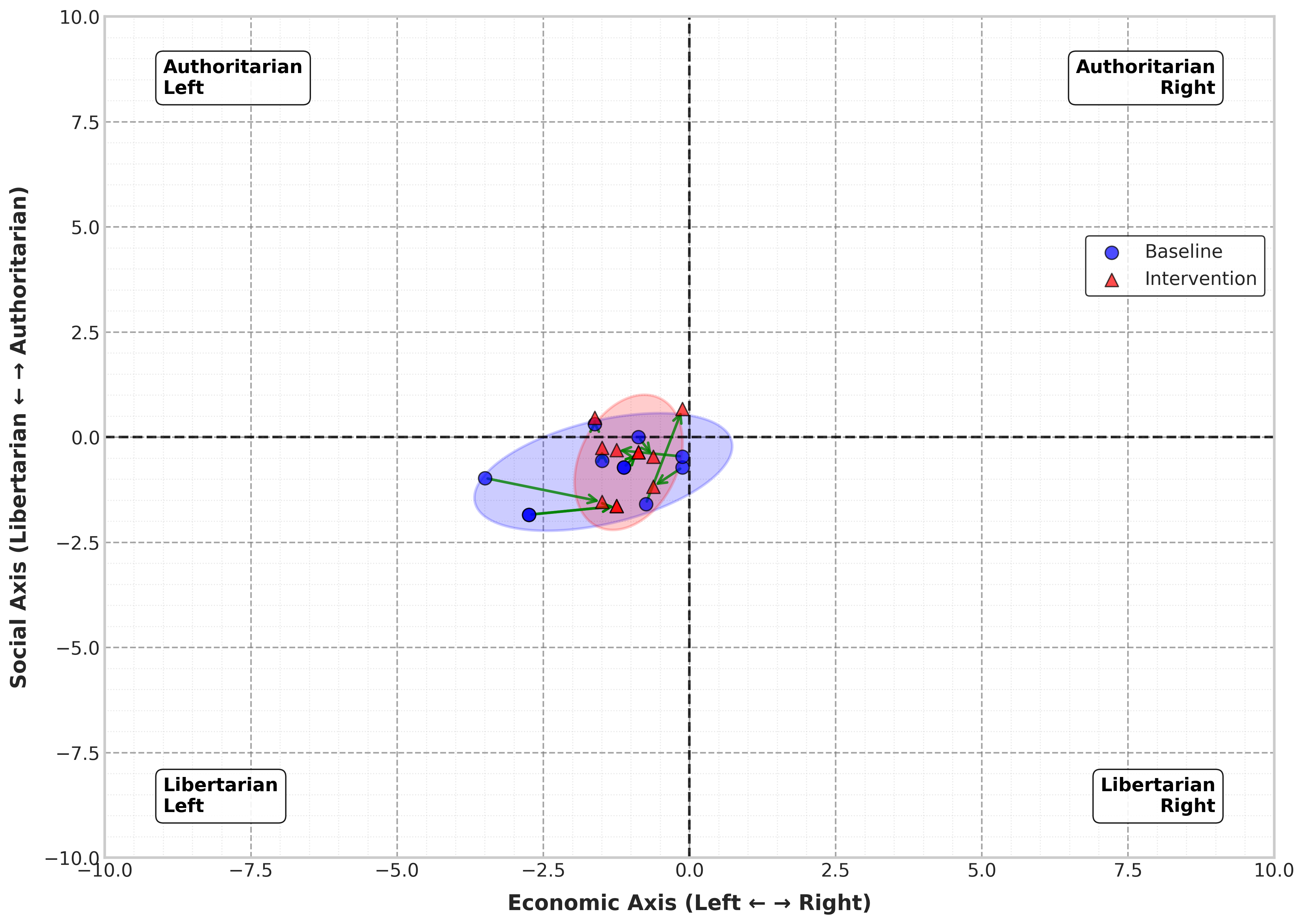}
    \caption{\texttt{Intervention Strength of 25}}
    \label{fig:inter_256_25_1_sl}
  \end{subfigure}

  \vspace{0.5em}

  \begin{subfigure}[t]{0.48\textwidth}
    \centering
    \includegraphics[width=\linewidth]{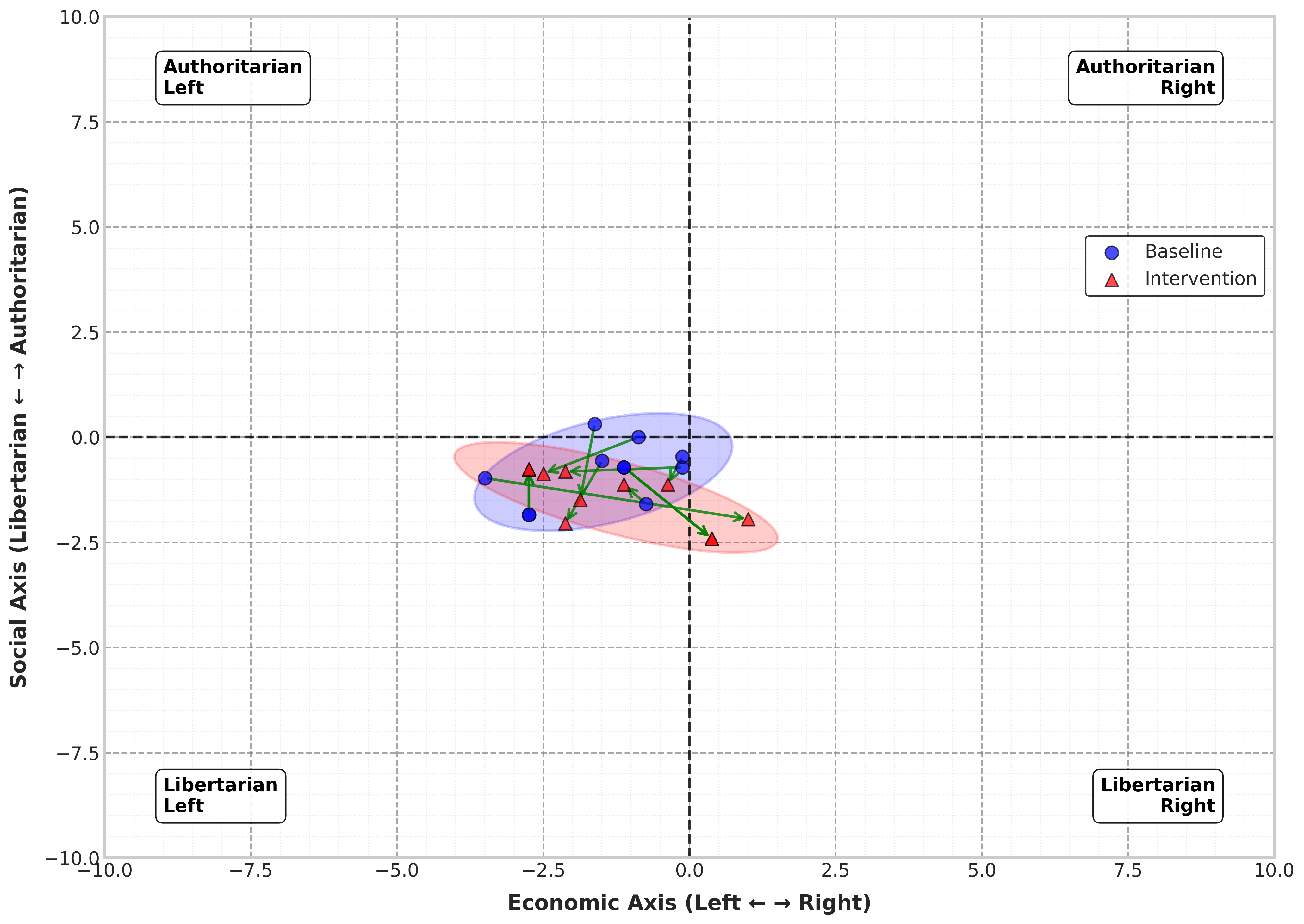}
    \caption{\texttt{Intervention Strength of 30}}
    \label{fig:inter_256_30_0_sl}
  \end{subfigure}
  \hspace{0.02\textwidth}
  \begin{subfigure}[t]{0.48\textwidth}
    \centering
    \includegraphics[width=\linewidth]{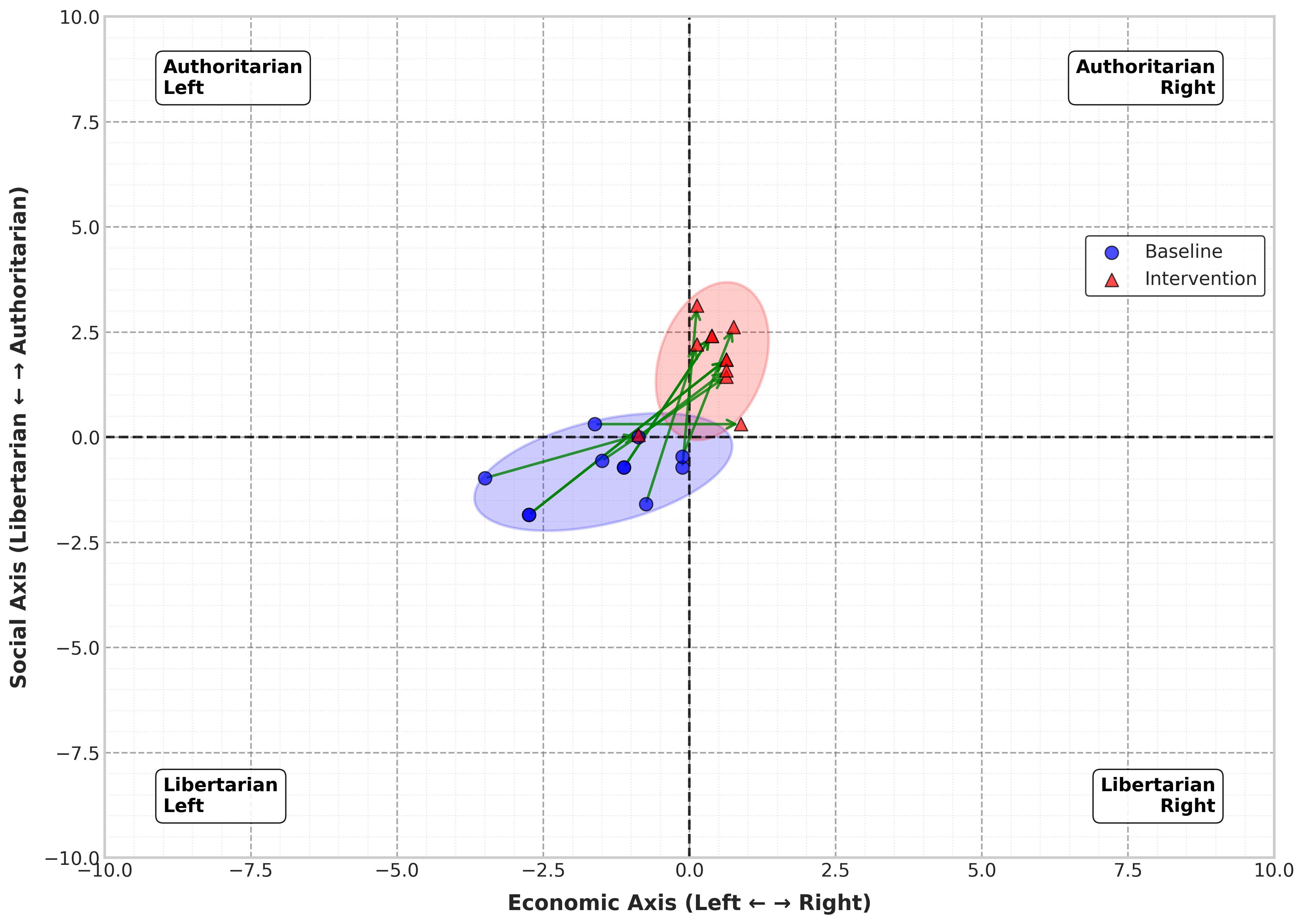}
    \caption{\texttt{Intervention Strength of 30}}
    \label{fig:inter_256_30_1_sl}
  \end{subfigure}

  \caption{Political compass intervention results on 256 heads for two different intervention strengths for both directions on the PCT test in \textbf{Slovenian}. The plots on the right demonstrate steering towards politically right responses, and the plots on the left--towards politically left responses.}
  \label{fig:inter_results_sl}
\end{figure*}

\begin{figure*}[h]
  \centering
  \begin{subfigure}[t]{0.48\textwidth}
    \centering
    \includegraphics[width=\linewidth]{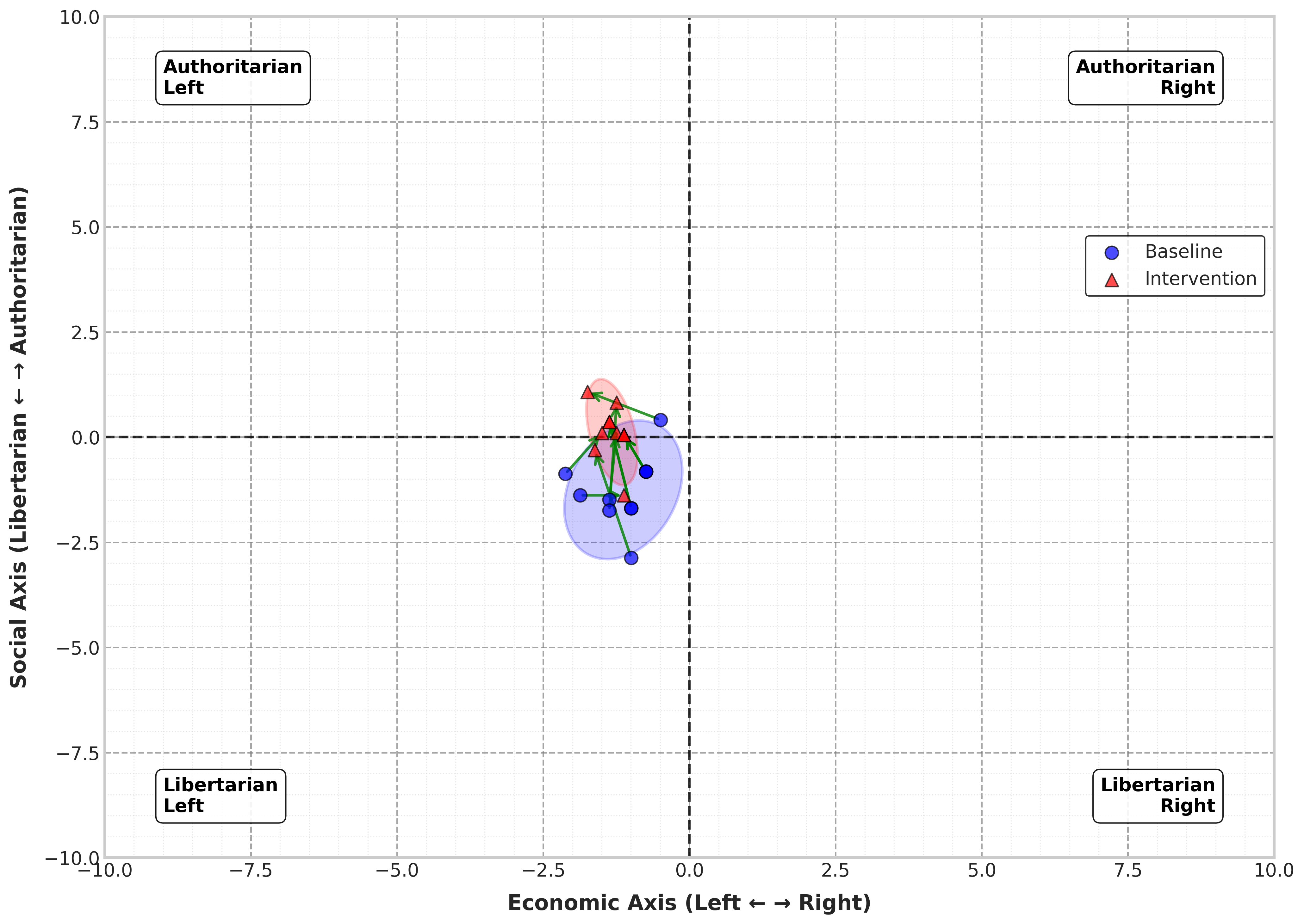}
    \caption{\texttt{Intervention Strength of 20}}
    \label{fig:inter_256_20_0_fr}
  \end{subfigure}
  \hspace{0.02\textwidth}
  \begin{subfigure}[t]{0.48\textwidth}
    \centering
    \includegraphics[width=\linewidth]{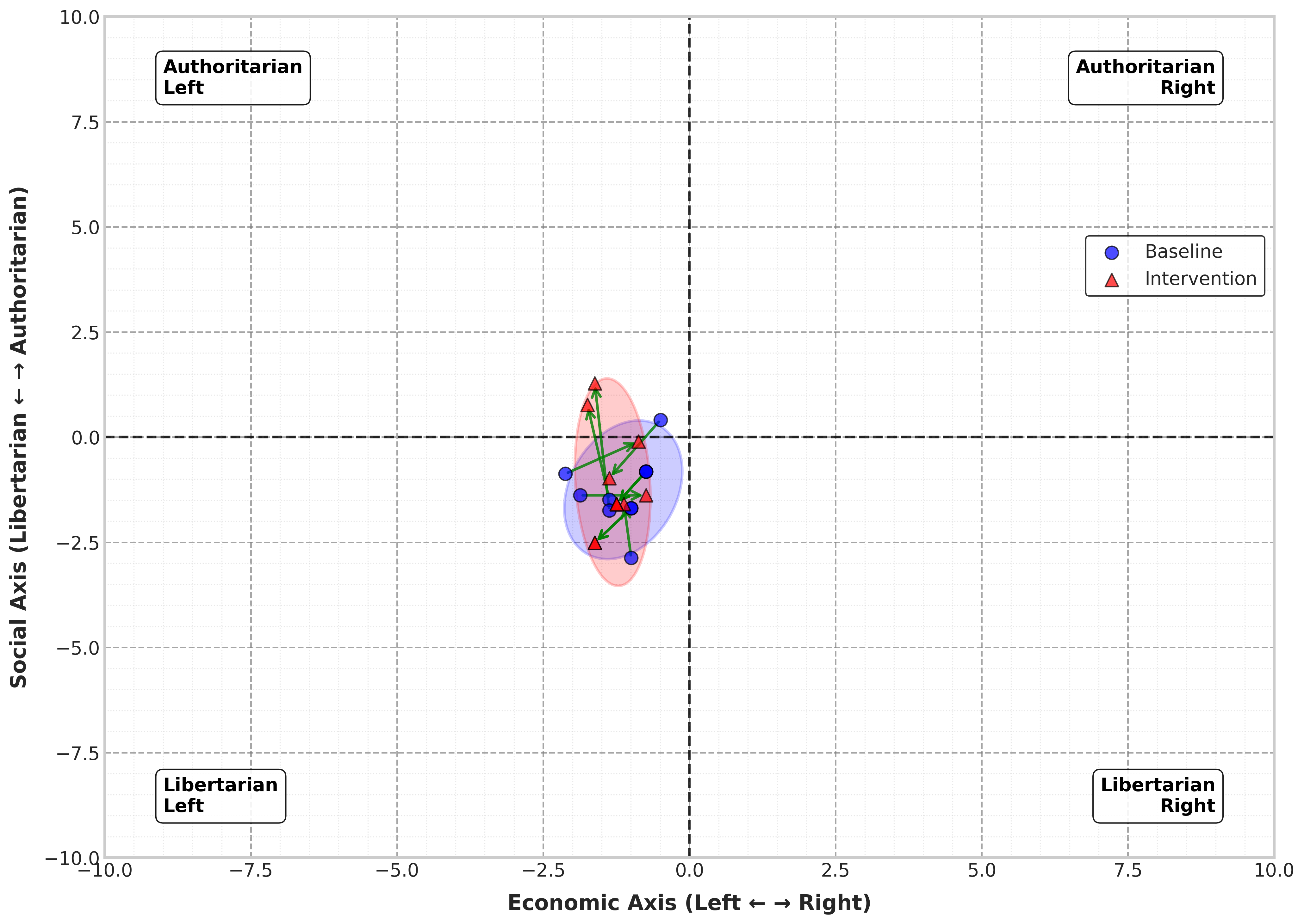}
    \caption{\texttt{Intervention Strength of 20}}
    \label{fig:inter_256_20_1_fr}
  \end{subfigure}

  \vspace{0.5em}

  \begin{subfigure}[t]{0.48\textwidth}
    \centering
    \includegraphics[width=\linewidth]{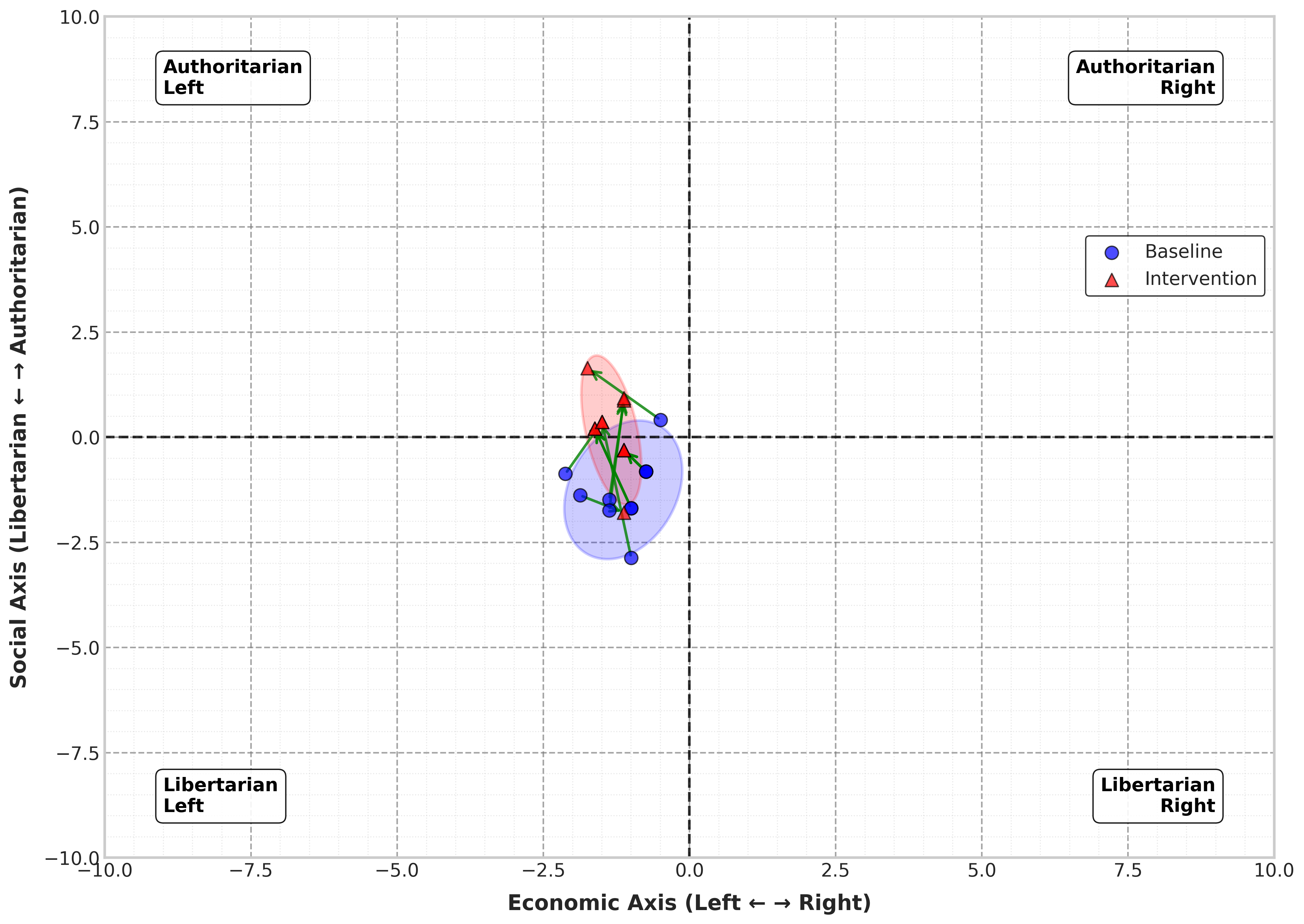}
    \caption{\texttt{Intervention Strength of 25}}
    \label{fig:inter_256_25_0_fr}
  \end{subfigure}
  \hspace{0.02\textwidth}
  \begin{subfigure}[t]{0.48\textwidth}
    \centering
    \includegraphics[width=\linewidth]{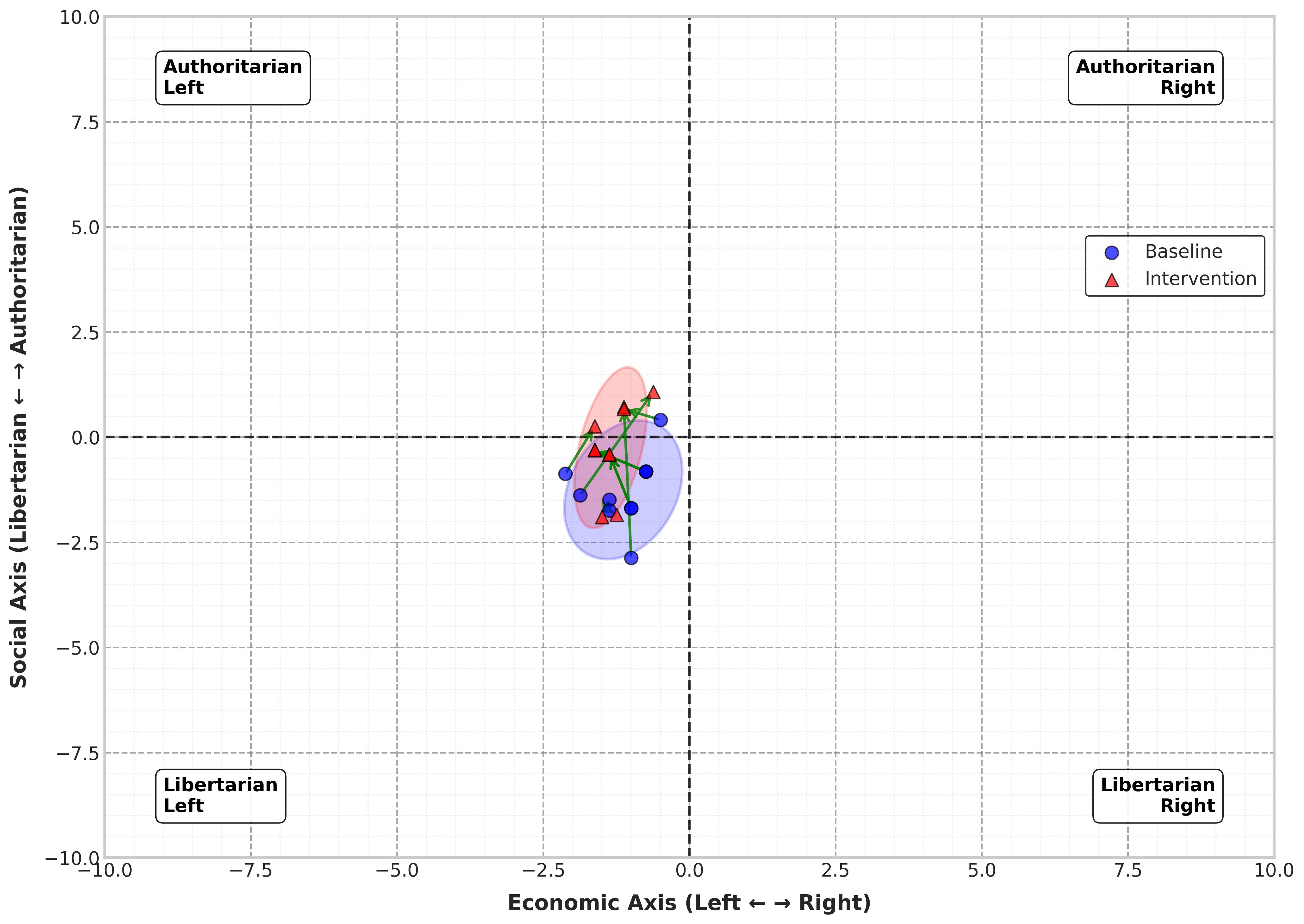}
    \caption{\texttt{Intervention Strength of 25}}
    \label{fig:inter_256_25_1_fr}
  \end{subfigure}

  \vspace{0.5em}

  \begin{subfigure}[t]{0.48\textwidth}
    \centering
    \includegraphics[width=\linewidth]{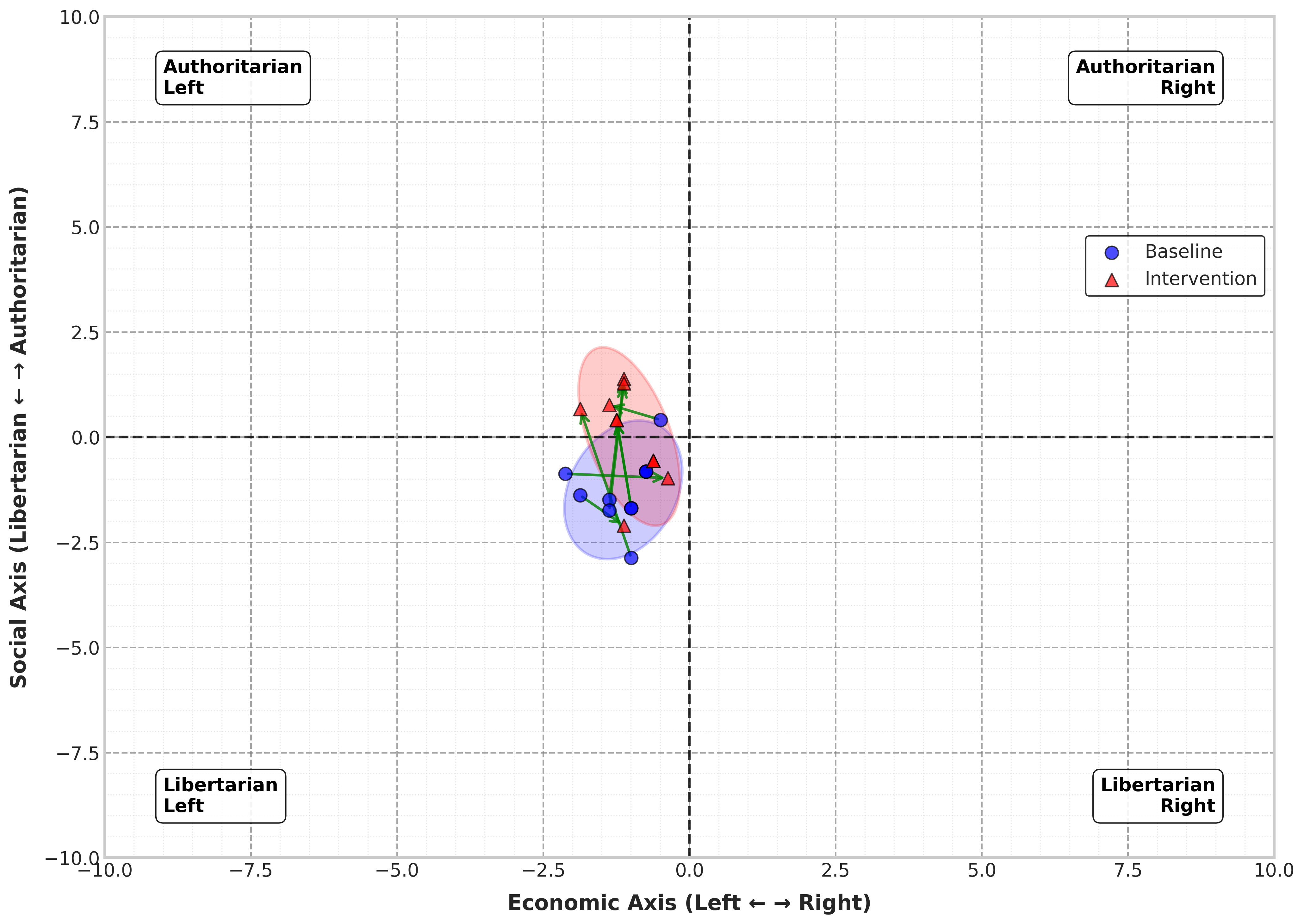}
    \caption{\texttt{Intervention Strength of 30}}
    \label{fig:inter_256_30_0_fr}
  \end{subfigure}
  \hspace{0.02\textwidth}
  \begin{subfigure}[t]{0.48\textwidth}
    \centering
    \includegraphics[width=\linewidth]{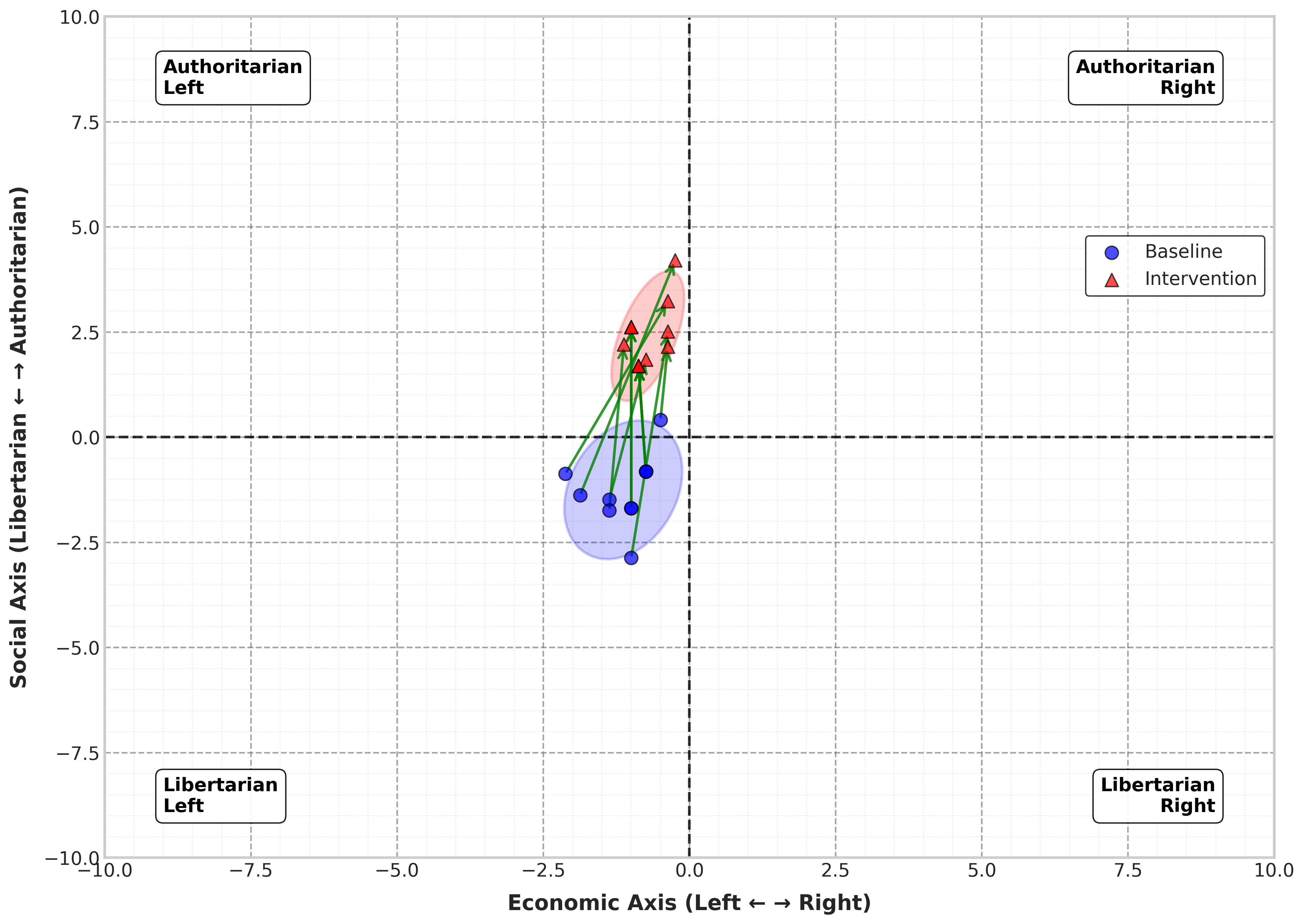}
    \caption{\texttt{Intervention Strength of 30}}
    \label{fig:inter_256_30_1_fr}
  \end{subfigure}

  \caption{Political compass intervention results on 256 heads for two different intervention strengths for both directions on the PCT test in \textbf{French}. The plots on the right demonstrate steering towards politically right responses, and the plots on the left--towards politically left responses.}
  \label{fig:inter_results_fr}
\end{figure*}
\clearpage
\newpage

\begin{figure}[h!]
\section{Detailed Response Choice Analysis}
\label{app:response_counts}
    \centering
    \includegraphics[width=0.9\linewidth]{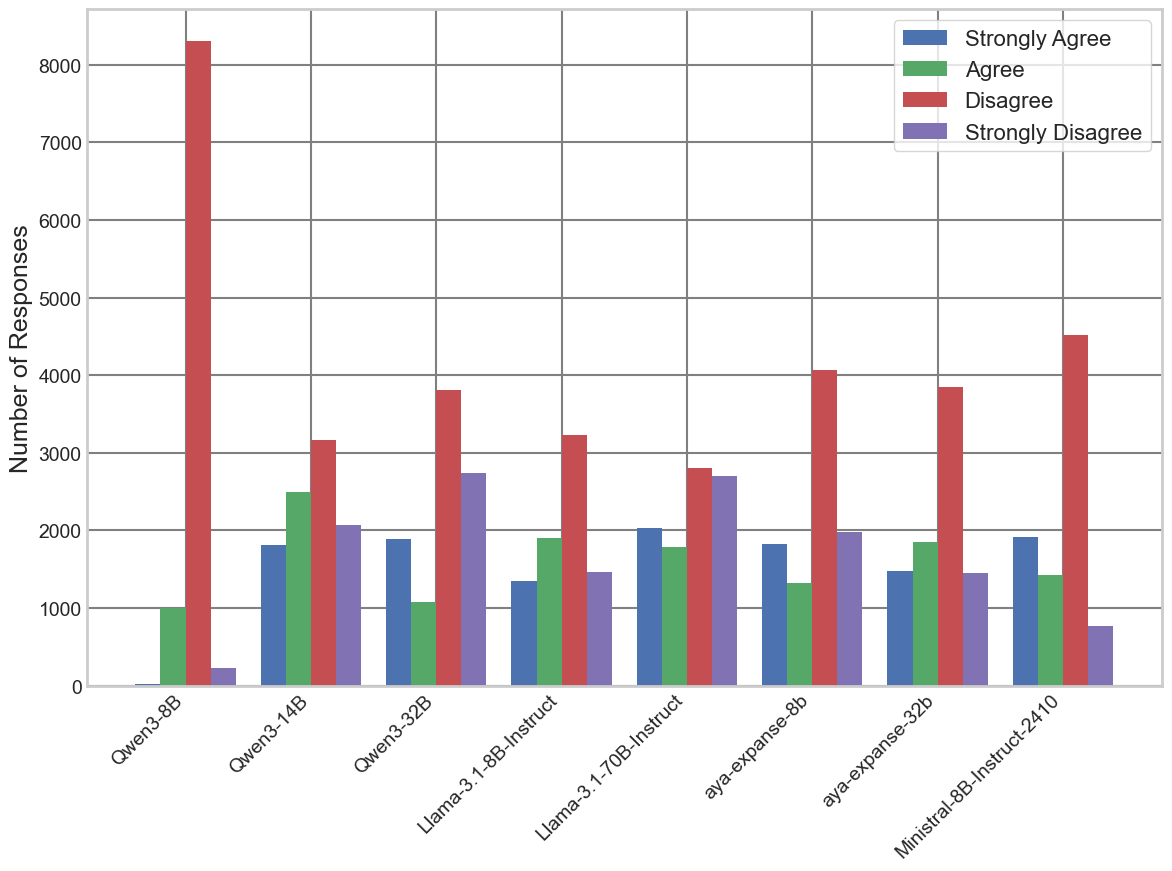}
    \caption{Selected choices across all languages and paraphrases for all tested models.}
    \label{fig:choice-count}
\end{figure}
\clearpage

\end{document}